\documentclass[default,iicol,pdflatex]{sn-jnl}% Math and Physical Sciences Reference Style
\usepackage{caption}
\usepackage{amsmath}

\usepackage{qtree}
\usepackage{bbm}
\usepackage{multirow}
\usepackage{adjustbox}
\usepackage{amsmath}

\DeclareMathOperator*{\argmax}{arg\,max} % Define argmax
\jyear{2021}%

\theoremstyle{thmstyleone}%
\theoremstyle{thmstyletwo}%

\theoremstyle{thmstylethree}%
\newtheorem{definition}{Definition}%
\begin{document}

\title[Article Title]{Counterfactual Learning on Graphs: A Survey}

%%=============================================================%%
%% Prefix   -> \pfx{Dr}
%% GivenName    -> \fnm{Joergen W.}
%% Particle -> \spfx{van der} -> surname prefix
%% FamilyName   -> \sur{Ploeg}
%% Suffix   -> \sfx{IV}
%% NatureName   -> \tanm{Poet Laureate} -> Title after name
%% Degrees  -> \dgr{MSc, PhD}
%% \author*[1,2]{\pfx{Dr} \fnm{Joergen W.} \spfx{van der} \sur{Ploeg} \sfx{IV} \tanm{Poet Laureate}
%%                 \dgr{MSc, PhD}}\email{iauthor@gmail.com}
%%=============================================================%%

\author[1]{\fnm{Zhimeng} \sur{Guo}}

\author[1]{\fnm{Teng} \sur{Xiao}}
%\equalcont{These authors contributed equally to this work.}

\author[1]{\fnm{Zongyu} \sur{Wu}}
%\equalcont{These authors contributed equally to this work.}

\author[2]{\fnm{Charu} \sur{Aggarwal}}

\author[3]{\fnm{Hui} \sur{Liu}}
%\equalcont{These authors contributed equally to this work.}

\author[1]{\fnm{Suhang} \sur{Wang}}
%\equalcont{These authors contributed equally to this work.}

\affil[1]{\orgdiv{College of Information Sciences and Technology}, \orgname{The Pennsylvania State University}, \orgaddress{\city{State College} \postcode{16802},  \country{USA}}}

\affil[2]{\orgdiv{}\orgname{IBM T.J. Watson Research Center}, \orgaddress{\city{New York} \postcode{10598},  \country{USA}}}

\affil[3]{\orgdiv{College of Engineering}, \orgname{Michigan State University}, \orgaddress{\city{East Lansing} \postcode{48824},  \country{USA}}}

%%==================================%%
%% sample for unstructured abstract %%
%%==================================%%

\abstract{Graph-structured data are pervasive in the real-world such as social networks, molecular graphs and transaction networks. Graph neural networks (GNNs) have achieved great success in representation learning on graphs, facilitating various downstream tasks. However, GNNs have several drawbacks such as lacking interpretability, can easily inherit the bias of data and cannot model casual relations. Recently, counterfactual learning on graphs has shown promising results in alleviating these drawbacks. Various approaches have been proposed for counterfactual fairness, explainability, link prediction and other applications on graphs. To facilitate the development of this promising direction, in this survey, we categorize and comprehensively review papers on graph counterfactual learning. We divide existing methods into four categories based on problems studied. For each category, we provide background and motivating examples, a general framework summarizing existing works and a detailed review of these works. We point out promising future research directions at the intersection of graph-structured data, counterfactual learning, and real-world applications. To offer a comprehensive view of resources for future studies, we compile a collection of open-source implementations, public datasets, and commonly-used evaluation metrics. This survey aims to serve as a ``one-stop-shop'' for building a unified understanding of graph counterfactual learning categories and current resources.}

%%================================%%
%% Sample for structured abstract %%
%%================================%%

\keywords{Counterfactual Learning, Graph Neural Networks, Fairness, Explainability}

%%\pacs[JEL Classification]{D8, H51}

%%\pacs[MSC Classification]{35A01, 65L10, 65L12, 65L20, 65L70}

\maketitle

\section{Introduction}\label{intro}

Graphs are a ubiquitous data structure and a universal language for representing objects and complex interactions~\cite{hamilton2020graph}. They can model a wide range of real-world systems, such as social networks~\cite{tabassum2018social}, chemical compounds~\cite{coley2019a}, knowledge graphs~\cite{zhang2020relational}, and recommendation systems~\cite{wu2022graph}. For instance, in social networks~\cite{traud2012social}, nodes represent people, and edges between nodes denote social connections between them. In molecular graphs, nodes correspond to atoms, and edges represent the chemical bonds between them, providing a structural representation of chemical compounds that can be used for tasks like drug discovery or material design~\cite{numeroso2021meg, wang2022molecular}.
The pervasiveness of graph-structured data has raised the broad attention of researchers on graph analytics and mining and various methods have been proposed~\cite{wu2020a}.

Network representation~\cite{tang2015line, grover2016nodevec}, which aims to learn low-dimensional vector representations of nodes or graphs that capture the intrinsic feature and structure information of nodes or graphs, is one essential task of graph mining. The learned representation can facilitate various downstream tasks such as node classification~\cite{kipf2017semisupervised}, link prediction~\cite{zhang2018link}, community detection~\cite{shchur2019overlapping} and graph classification~\cite{sui2022causal}. Neural networks have shown great power in representation learning for many domains such as computer vision~\cite{he2016deep}, natural language processing~\cite{devlin2019bert} and etc.  Neural network-based methods have also inspired the emergence and flourishing of graph neural networks (GNNs)~\cite{kipf2017semisupervised, liu2020graph, xu2019how}. Since graph convolutional network~\cite{kipf2017semisupervised} was proposed, there have been various variants of GNNs~\cite{chen2020simple, klicpera2019predict, kim2021how}. They greatly boost the development of graph learning methods and have achieved state-of-the-art performance on many graph mining tasks, e.g., node classification~\cite{kim2021how}, link prediction~\cite{pan2022neural} and graph classification~\cite{papp2021dropgnn}. 
Due to the great power of graph learning methods, they have been successfully applied in many high-stakes decision scenarios, such as drug discovery~\cite{xiong2019pushing}, fake news detection~\cite{shu2017fake} and financial analysis~\cite{wu2021fedgnn}. 

\begin{figure*}[t]
    \centering
    \includegraphics[width=0.9\linewidth]{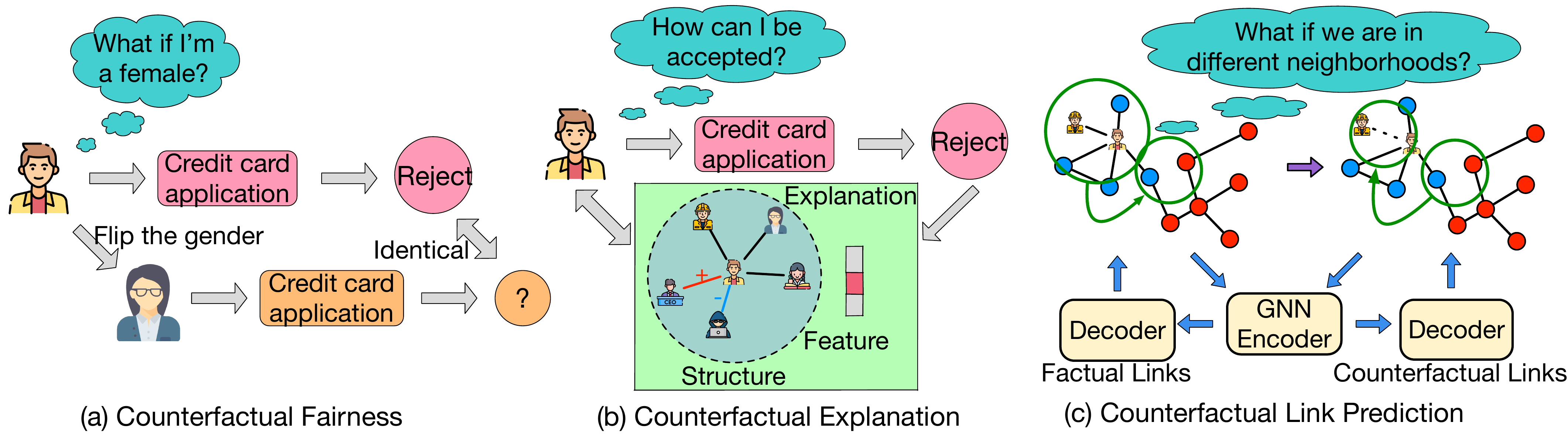}
    \captionsetup{justification=centering}
    \caption{An illustration of counterfactual learning on graphs.}
    \label{fig:total}
\end{figure*}

Despite the great potential of real-world benefits, recent studies show that existing graph learning methods tend to inherit the bias pattern from the biased dataset~\cite{dai2021say}, lack the interpretability~\cite{dai2021towards} and cannot exploit the rich information stored in graph data~\cite{zhao2022learning}. For example, with the biased dataset, GNNs are easy to learn an unfair classifier, e.g., give applicants different decisions based on their races or other sensitive information~\cite{dong2022edits, kose2021fairnessaware}. These issues severely hinder trust in the model and limit the real-world application of graph learning methods~\cite{dai2022a}. Counterfactual learning gives the chance to alleviate the intrinsic bias~\cite{kusner2017counterfactual}, making models interpretable~\cite{verma2020counterfactual} and exploiting the information stored in data well~\cite{pitis2020counterfactual}. The notation of counterfactual comes from the research community of causal inference~\cite{yao2021a}. Counterfactual aims to answer ``what would have happened, given the knowledge of what in fact happened''. The ability to learn with counterfactuals and generalize to unseen environments is considered a significant component of general AI. The topic of learning causality has been well studied in many areas, such as economics~\cite{imbens2004nonparametric}, education~\cite{dehejia1999causal}, and medical science~\cite{consortium2013identification}. To know the causal effect of an action, we need to know the factual outcome with the observed action and the counterfactual outcome with unobserved action. A straightforward approach is to conduct randomized controlled trials to get counterfactual outcomes. However, in the real-world setting, we only have access to the observational factual data, i.e., the observed action and its corresponding factual outcome, which is a key challenge to learn causality~\cite{guo2020a}. Fortunately, the development of information technology gives abundant data sources that we can take advantage of to find the implied information in the data~\cite{kaddour2022causal}. Hence, the core question is how to get the counterfactuals from the observational data~\cite{glymour2016causal}, and how to use the counterfactuals to aid machine learning models~\cite{kaddour2022causal}.

Counterfactual learning on graphs is an emerging direction and only has a very short history~\cite{kaddour2022causal}. However, recent works on graph counterfactual learning have shown great potential to overcome the aforementioned challenges on fairness~\cite{mehrabi2021a}, explanation~\cite{carvalho2019machine} and etc. In Figure~\ref{fig:total}, we show some motivation examples for graph counterfactual learning. Concretely, equipped with counterfactual learning, we can go beyond the fairness definition at the group level and achieve fairness for each individual as in the factual world and the counterfactual world~\cite{kusner2017counterfactual}, where the individual belongs to a different demographic group. As depicted in Figure~\ref{fig:total}~(a), the goal of counterfactual fairness is to ensure that an applicant and his counterfactual counterpart (with a different gender) receive the same credit card application outcome. For counterfactual explanation on graphs, in addition to finding a compact subgraph which is highly correlated to the prediction~\cite{ying2019gnnexplainer}, it aims at finding a reasonable change to have different result~\cite{lucic2022cf}, which can be used to not only answer why the model gives such prediction but also give suggestions on what to do in order to achieve another desired results. As illustrated in Figure~\ref{fig:total}~(b), in a credit card application scenario\cite{verma2020counterfactual}, when an applicant is rejected, a conventional explanation might state that their ``credit score was too low.'' In contrast, a counterfactual explanation could provide actionable recommendations on what minimal changes (e.g., in transaction relationships) the customer could make to alter the decision and ultimately gain approval. Besides the aid on fairness and interpretability, the research community also utilizes counterfactual learning to provide additional information from the counterfactual world, e.g., using both factual links and counterfactual links to help build more powerful GNNs~\cite{zhao2022learning}. As shown in Figure~\ref{fig:total}~(c), two friends live in the same neighborhood. By placing them in different neighborhoods, GNNs can infer the counterfactual link between them. This enables GNNs to gain a deeper understanding of the causal factors that shape their relationships while mitigating the impact of neighborhood factors. Considering the increasing trend of graph counterfactual learning and the diversity of related pretext tasks, there is an urgent need to have a systematic taxonomy to summarize the methodologies and applications of graph counterfactual learning.

\begin{figure*}[t]
    \centering
    \includegraphics[width=0.9\linewidth]{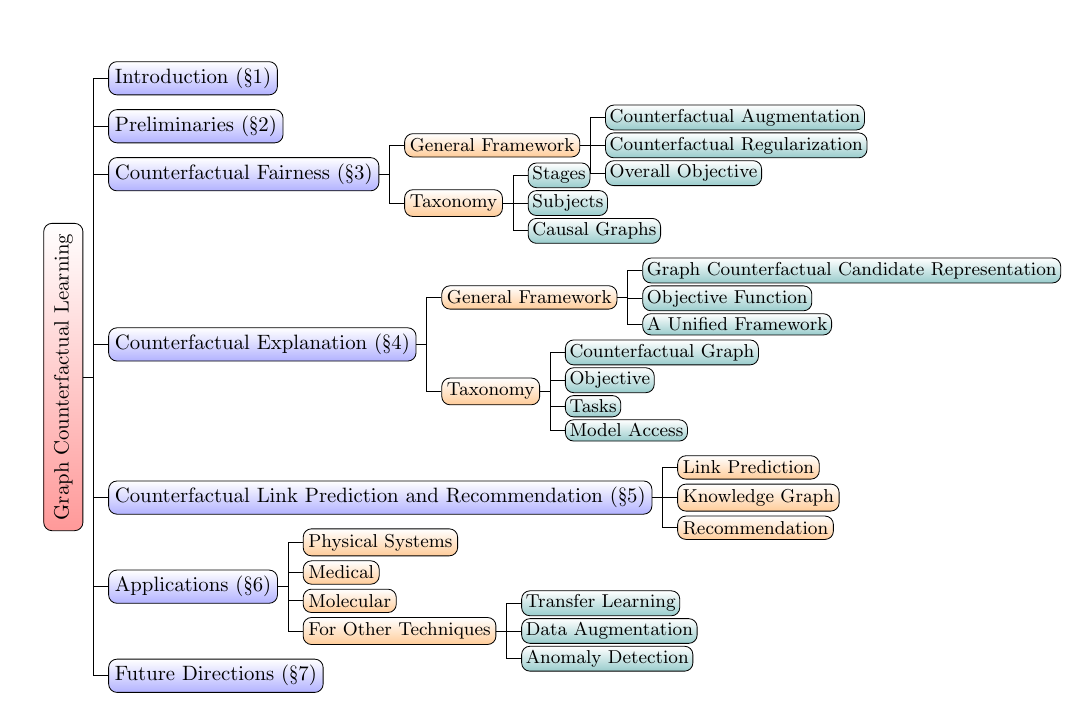}
\caption{Overview of graph counterfactual learning.}
\label{fig:overview}

\end{figure*}

To fill the gap, this survey paper conducts the first comprehensive and up-to-date overview of the booming area of graph counterfactual learning, provides some insights and potential future directions, and creates a ``one-step-stop'' that collects a set of open-source implementations, public datasets and commonly-used evaluation metrics together. The intended audiences for this article are general machine learning researchers who would like to know graph counterfactual learning, graph learning researchers who want to keep track of the most recent advances in graph neural networks, and domain experts who would like to generalize graph counterfactual learning to new applications or other fields. The core contributions of this survey are:
\begin{itemize}[]
    \item \textbf{The first survey of counterfactual learning on graphs}. To the best of our knowledge, our survey is the first to review counterfactual learning techniques for graphs. The most relevant surveys are about causal inference~\cite{guo2020a, yao2021a} and causal machine learning~\cite{kaddour2022causal}. Until now, there has been no dedicated and comprehensive survey about causal learning in graph domain.
    \item \textbf{A comprehensive and up-to-date review.} We review the most up-to-date graph counterfactual learning techniques published in influential international conferences and journals of deep learning, data mining, computer vision, natural language processing and artificial intelligence, including ACM TOIS, ICLR, NeurIPS, ICML, SIGKDD, WSDM, CIKM, WWW, ICDM, NAACL, IJCAI, AAAI, and others. We also include papers in other domains like chemical science.
    \item \textbf{Systematic taxonomy and unified frameworks.} We systematically categorize existing works into counterfactual fairness, counterfactual explanation, counterfactual link prediction and recommendation, and applications. For most of the categories, we provide unified frameworks that mathematically formalize graph counterfactual learning approaches in each category. An overview of the taxonomy is shown in Figure~\ref{fig:overview}.
    \item \textbf{Future directions and ``one-step-stop'' for resources.} From the survey results, we point out promising and important future directions. We also provide a collection of open-source implementations, public datasets, and commonly-used evaluation metrics to facilitate the community. We maintain a repository containing papers in graph counterfactual learning and we will keep updating these papers in the repository: \url{https://github.com/TimeLovercc/Awesome-Graph-Causal-Learning}.
\end{itemize}

\begin{table*}[t]
\centering 
\footnotesize
\renewcommand\arraystretch{1}
\caption{A comparison between existing surveys on causal inference and graph learning and ours. \textbf{Unified} denotes unified framework.} \label{table:comparison}
\resizebox{0.96\textwidth}{!}{
\begin{tabular}{l|c|c|c|c|c|c|c|c|c|c}
\toprule
\multirow{2}{*}{\textbf{Surveys}} & \multirow{2}{*}{\textbf{Graph}} & \multirow{2}{*}{\textbf{Unified}} 
& \multirow{2}{*}{\textbf{Causal}} & \multicolumn{3}{c|}{\textbf{Counterfactual}} & \multirow{2}{*}{\textbf{Applications}} & \multirow{2}{*}{\textbf{Source Code}} & \multicolumn{2}{c}{\textbf{Dataset}}  \\ \cline{5-7}  \cline{10-11}
& & & &\textbf{Fairness}  &\textbf{Explanation}  & \textbf{Data Augmentation} &  & &\textbf{Real-world}  &\textbf{Synthetic} \\ 
\midrule

Our Survey & \includegraphics[scale=0.15]{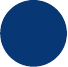} & \includegraphics[scale=0.15]{pics/4.pdf}& \includegraphics[scale=0.15]{pics/4.pdf}& \includegraphics[scale=0.15]{pics/4.pdf}& \includegraphics[scale=0.15]{pics/4.pdf}& \includegraphics[scale=0.15]{pics/4.pdf}& \includegraphics[scale=0.15]{pics/4.pdf}& \includegraphics[scale=0.15]{pics/4.pdf}& \includegraphics[scale=0.15]{pics/4.pdf}& \includegraphics[scale=0.15]{pics/4.pdf} \\

\midrule

% fairness
\citeauthor{dai2021towards}~\cite{dai2021towards} & \includegraphics[scale=0.15]{pics/4.pdf} & \includegraphics[scale=0.15]{pics/4.pdf} & -  & \includegraphics[scale=0.15]{pics/4.pdf} & - & - & \includegraphics[scale=0.15]{pics/4.pdf} & - & \includegraphics[scale=0.15]{pics/4.pdf} & \includegraphics[scale=0.15]{pics/4.pdf}\\
\citeauthor{mehrabi2021a}~\cite{mehrabi2021a} & - & - & \includegraphics[scale=0.15]{pics/4.pdf} & \includegraphics[scale=0.15]{pics/4.pdf} & - & - & \includegraphics[scale=0.15]{pics/4.pdf} & - & \includegraphics[scale=0.15]{pics/4.pdf} & - \\
\citeauthor{makhlouf2020survey}~\cite{makhlouf2020survey} & - & \includegraphics[scale=0.15]{pics/4.pdf} & \includegraphics[scale=0.15]{pics/4.pdf} & \includegraphics[scale=0.15]{pics/4.pdf}  & - & -  & - & \includegraphics[scale=0.15]{pics/4.pdf} & - \\

\midrule

% causal
\citeauthor{yao2021survey}~\cite{yao2021survey} & - & \includegraphics[scale=0.15]{pics/4.pdf} & \includegraphics[scale=0.15]{pics/4.pdf} & - & - & \includegraphics[scale=0.15]{pics/4.pdf}  & \includegraphics[scale=0.15]{pics/4.pdf} & \includegraphics[scale=0.15]{pics/4.pdf} & \includegraphics[scale=0.15]{pics/4.pdf} &
\includegraphics[scale=0.15]{pics/4.pdf}\\
\citeauthor{kaddour2022causal}~\cite{kaddour2022causal} & \includegraphics[scale=0.15]{pics/4.pdf} & - & \includegraphics[scale=0.15]{pics/4.pdf} & \includegraphics[scale=0.15]{pics/4.pdf} & \includegraphics[scale=0.15]{pics/4.pdf} & \includegraphics[scale=0.15]{pics/4.pdf} & \includegraphics[scale=0.15]{pics/4.pdf} & - & - & - \\
\citeauthor{guo2020survey}~\cite{guo2020survey} & - & \includegraphics[scale=0.15]{pics/4.pdf} & \includegraphics[scale=0.15]{pics/4.pdf} & - & - & \includegraphics[scale=0.15]{pics/4.pdf} & \includegraphics[scale=0.15]{pics/4.pdf} & - & \includegraphics[scale=0.15]{pics/4.pdf} & \includegraphics[scale=0.15]{pics/4.pdf} \\

\midrule

% explanation
\citeauthor{verma2020counterfactual}~\cite{verma2020counterfactual} & - & \includegraphics[scale=0.15]{pics/4.pdf} & \includegraphics[scale=0.15]{pics/4.pdf} &  - & \includegraphics[scale=0.15]{pics/4.pdf} & - & - & - & \includegraphics[scale=0.15]{pics/4.pdf} & - \\
\citeauthor{9321372}~\cite{9321372} & - & - & \includegraphics[scale=0.15]{pics/4.pdf} & - & \includegraphics[scale=0.15]{pics/4.pdf} & - & - & - & - & - \\
\citeauthor{artelt2019computation}~\cite{artelt2019computation} & - & \includegraphics[scale=0.15]{pics/4.pdf} & - & - & \includegraphics[scale=0.15]{pics/4.pdf} & - & - &  & - & - \\
\citeauthor{oneto2022towards}~\cite{oneto2022towards} & \includegraphics[scale=0.15]{pics/4.pdf} & - & - & \includegraphics[scale=0.15]{pics/4.pdf} & \includegraphics[scale=0.15]{pics/4.pdf} & - & - & - & - & - \\
\citeauthor{Prado2022Survey}~\cite{Prado2022Survey} & \includegraphics[scale=0.15]{pics/4.pdf} & \includegraphics[scale=0.15]{pics/4.pdf} & \includegraphics[scale=0.15]{pics/4.pdf} & - & \includegraphics[scale=0.15]{pics/4.pdf} & - & \includegraphics[scale=0.15]{pics/4.pdf} & - & \includegraphics[scale=0.15]{pics/4.pdf} & \includegraphics[scale=0.15]{pics/4.pdf} \\

\midrule

% application
\citeauthor{feder2021causal}~\cite{feder2021causal} & - & \includegraphics[scale=0.15]{pics/4.pdf} & \includegraphics[scale=0.15]{pics/4.pdf} & \includegraphics[scale=0.15]{pics/4.pdf} & \includegraphics[scale=0.15]{pics/4.pdf} & - & - & - & - & \includegraphics[scale=0.15]{pics/4.pdf} \\
\citeauthor{cheng2021causal}~\cite{cheng2021causal} & - & - & \includegraphics[scale=0.15]{pics/4.pdf} & \includegraphics[scale=0.15]{pics/4.pdf} & \includegraphics[scale=0.15]{pics/4.pdf} & \includegraphics[scale=0.15]{pics/4.pdf} & - & - & - & - \\
\citeauthor{liucausal}~\cite{liucausal} & - & - & \includegraphics[scale=0.15]{pics/4.pdf} & - & - & \includegraphics[scale=0.15]{pics/4.pdf} & \includegraphics[scale=0.15]{pics/4.pdf} & - & \includegraphics[scale=0.15]{pics/4.pdf} & \includegraphics[scale=0.15]{pics/4.pdf} \\
\citeauthor{sanchez2022causal}~\cite{sanchez2022causal} & - & - & \includegraphics[scale=0.15]{pics/4.pdf} & - & - & \includegraphics[scale=0.15]{pics/4.pdf} & \includegraphics[scale=0.15]{pics/4.pdf} & - & - & -\\
\citeauthor{vlontzos2022review}~\cite{vlontzos2022review} & - & - & \includegraphics[scale=0.15]{pics/4.pdf} & \includegraphics[scale=0.15]{pics/4.pdf} & \includegraphics[scale=0.15]{pics/4.pdf} & - & - & - & - & - \\
\bottomrule
\end{tabular}
}

\end{table*}

\textit{Comparison with related survey articles.} Table~\ref{table:comparison} highlights the differences between our survey and related survey papers. Most existing surveys primarily focus on general causal inference~\cite{yao2021a, guo2020a}, counterfactual fairness~\cite{kusner2017counterfactual}, and counterfactual explanation~\cite{verma2020counterfactual}, seldom discussing research progress on graph data. While other graph domain surveys address fairness~\cite{dai2022a} and interpretability~\cite{dai2022a, fan2021interpretability}, they rarely summarize existing work from causal or counterfactual learning perspectives~\cite{Prado2022Survey}. Our survey provides the first comprehensive overview of graph counterfactual learning, offering causal learning background, reviewing graph counterfactual learning techniques for fairness-aware models, explainable models, link prediction, recommender systems, real-world applications, and promising research directions. Hence, our survey is distinct from existing surveys and can support the growth of this important and emerging domain.

The overview of this survey is shown in Figure~\ref{fig:overview}. Section~\ref{pre} defines the related concepts and gives notations which will be used in the following sections. Section~\ref{fairness} and Section~\ref{explanation} describe the unified framework of counterfactual fairness and counterfactual explanation on graph data, respectively. We also summarize useful resources including evaluation metrics and datasets.  Section~\ref{link} reviews the categories of counterfactual link prediction and counterfactual recommendation. Section~\ref{application} surveys the real-world applications of graph counterfactual learning in various domains. Section~\ref{future} points out the challenges unsolved and promising future directions. Section~\ref{con} concludes this survey.

\section{Preliminary}\label{pre}

In this section, we give the notations and definitions frequently used in this survey. We also provide some background knowledge for the following sections.

\subsection{Notations and Definitions}

\begin{table}
\centering
  \caption{Notations and explanations}
  \resizebox{1\columnwidth}{!}{%
    \begin{tabular}{ll}
    \toprule
    \textbf{Notations} & \textbf{Explainations} \\
    \midrule
	$d$						& Dimension of vectors \\
	$N$						& Number of nodes \\
    $\mathbf{A}$		& Adjacency matrix \\
    $\mathbf{x}_{i}$ or $\mathbf{x}_{i}^{\text{F}}$		& Factual attributes of node $v_i$ \\
    $\mathbf{x}_{i}^{\text{CF}}$		& Counterfactual attributes of node $v_i$ \\
	$\mathbf{h}_{i}$ or $\mathbf{h}_{i}^{\text{F}}$		& Factual embeddings of node $v_i$ \\
    $\mathbf{h}_{i}^{\text{CF}}$		& Counterfactual ebmeddings of node $v_i$ \\
    $\mathbf{Y}$            & Labels \\
 	$\mathcal{N}_{i}$	& Neighbors of node $v_i$ \\
 	$\sigma$				& Sigmoid function \\
 	$\odot$					& Hadamard product \\
    $S$ and $s$ & Sensitive variable and sensitive attribute \\
    \bottomrule
    \end{tabular}%
    }
  \label{notations}
\end{table}

Throughout the paper, variables are denoted by capital letters. Lowercase letters denote specific values of variables. Matrices are written in boldface capital letters and vectors are denoted in boldface lowercase letters, respectively. Capital letters in calligraphic math font are used to denote sets.
such as $ \mathcal{P} $ are used to denote sets and $\lvert \mathcal{P} \rvert$ denotes the cardinality of $\mathcal{P}$. Let $\mathcal{G}=\{\mathcal{V},\mathcal{E},\mathbf{X}\}$ represent a graph, where $\mathcal{V}=\{v_1, \dots,v_N\}$ denotes the set of $N$ nodes, $\mathcal{E}$ represents the set of edges, and $\mathbf{X} \in \mathbb{R}^{N \times d_x}$ signifies the node features. The $i$-th row of $\mathbf{X}$, $\mathbf{x}_i$, is a $d_x$-dimensional feature vector for node $v_i$. The graph's adjacency matrix is denoted as $\mathbf{A} \in \{0,1\}^{n \times n}$, where $\mathbf{A}_{ij}=1$ if there is an edge from $v_i$ to $v_j$; otherwise, $\mathbf{A}_{ij}=0$. The labels of graphs or nodes, depending on the task, are represented by $\mathbf{Y}=\{y_1, \dots, y_N\}$, where $y_i \in \{1,2,\dots,C\}$ is the label of the $i$-th data sample and $C$ denotes the number of classes. Table~\ref{notations} summarizes the most frequently used notations and definitions. Next, we provide basic definitions for causal inference and counterfactual learning within the potential outcome framework~\cite{rubin1974estimating,rubin2005causal, yao2021a}.

\begin{definition}[Unit~\cite{yao2021a}]
A unit is an atomic research object when studying the treatment effect.
\end{definition}

\begin{definition}[Treatment]
Treatment refers to the action that applies (exposes, or subjects) to a unit. We consider binary treatment (0/1) here. We use $W$ and $w$ to denote the random variable of treatment and the instance of treatment, respectively. For the observation treatment $W=w$, $w^\prime$ stands for the counterfactual treatment, i.e., $w^\prime = 1 - w$. 
\end{definition}

\begin{definition}[Potential Outcome]
For each individual, there  exist two potential outcomes denoted as  $Y(0)$ and $ Y(1) \in \mathcal{C} $, where $Y(0)$ is a potential outcome associated with treatment $W=0$ and $Y(1)$ is the potential outcome associated with $W=1$. Note that each individual receives only one treatment and reveals the factual outcome value for the received treatment. 
\end{definition}  

\begin{definition}[Observed Outcome (Factual Outcome) and Counterfactual Outcome~\cite{yao2021a}]
The observed outcome $Y(W=w)$ is the outcome  when applying the actual treatment $w$. Counterfactual outcome $Y(W=w^\prime)$ is the outcome if the unit had taken treatment $w^\prime$ given that the unit actually took treatment $w$.
\end{definition}

% \begin{definition}[Counterfactual Outcome~\cite{yao2021a}]
% Counterfactual outcome $Y(W=w^\prime)$ is the outcome if the unit had taken treatment $w^\prime$ given that the unit actually took treatment $w$.
% \end{definition}

\subsection{Graph Learning}

Graphs are a common data structure used in various fields like social networks~\cite{tabassum2018social}, brain connectomes~\cite{bullmore2011brain}, trade networks~\cite{hamilton2020graph}, and recommender systems~\cite{ying2018graph}. Graph learning tasks are divided into graph-level, node-level, and edge-level tasks~\cite{zhang2018an, kipf2017semisupervised, pan2022neural}. Both node features and geometric structures are crucial for these tasks~\cite{hamilton2020graph,kipf2016semi}, so effectively encoding graph structure and node attributes into vector representations is essential. Simple approaches extract basic graph properties~\cite{bhagat2011node}, while others use graph statistics like kernel functions~\cite{vishwanathan2010graph}, local neighborhood structure~\cite{liben2007link}, and graphlet orbits~\cite{prvzulj2007biological}. However, these methods may be limited in flexibility, expressiveness, and computational efficiency due to their reliance on hand-engineered features~\cite{hamilton2020graph}.

Another line of research focuses on network representation learning, aiming to learn low-dimensional vector representations of nodes or graphs~\cite{qiu2018network, tang2015line, grover2016nodevec, kipf2017semisupervised, velickovic2018graph}. Among these, Graph Neural Networks (GNNs)\cite{kipf2016semi, chen2020simple, xu2018powerful, xiao2021learning, velickovic2018graph, xu2022hpgmn} have demonstrated a strong ability to learn node and graph representations, facilitating various downstream tasks\cite{kipf2016semi, chen2020simple, xu2018powerful, velickovic2018graph, dai2021nrgnn, zhao2021graphsmote}. GNNs generally adopt a message-passing mechanism that learns a node representation by iteratively aggregating the node's neighborhood information~\cite{kipf2016semi}. The learned representations capture both node attributes and local neighborhood information. Concretely, we use $\mathbf{h}_i^{(l)}$ to denote the representation of node $v_i$ in the $l$-th layer of the GNN with  $\mathbf{h}_i^{(0)} = \mathbf{x}_i$. The overall representation of all the nodes in $l$-th layer is denoted as $\mathbf{H}^{(l)}$. Let $\mathcal{N}_i=\{v_j \in \mathcal{V} \mid(v_i, v_j) \in \mathcal{E}\}$ denote the neighbor set of node $v_i$. The message-passing mechanism can be written into a unified framework, i.e., in the $l$-th layer, each node $v_i \in \mathcal{V}$ aggregates information from its neighbors as:
\begin{equation}
% \footnotesize
\fontsize{8pt}{10pt}\selectfont
\begin{aligned}
\mathbf{h}_i^{(l)} &=\operatorname{UP}^{(l-1)}\Big(\mathbf{h}_i^{(l-1)}, \text { AGG }^{(l-1)}\big(\{\mathbf{h}_j^{(l-1)}, \forall v_j \in \mathcal{N}_i\}\big)\Big) \\
&=\operatorname{UP}^{(l-1)}\big(\mathbf{h}_i^{(l-1)}, \mathbf{m}_{\mathcal{N}_i}^{(l-1)}\big),
\end{aligned}
\end{equation}

where \text{AGG} is an aggregation function and \text{UP} is an update function. 
For example, GCN~\cite{kipf2017semisupervised} takes symmetric normalization as $\mathbf{h}_i^{(k)}=\sigma\big(\mathbf{W}^{(k)} \sum_{v_j \in \mathcal{N}_i \cup\{v_i\}} \frac{\mathbf{h}_j}{\sqrt{\lvert\mathcal{N}_i\rvert \lvert\mathcal{N}_j\rvert}}\big)$, where $\operatorname{AGG}(\cdot) = \sum_{v_j \in \mathcal{N}_i} \frac{\mathbf{h}_j}{\sqrt{\lvert\mathcal{N}_i\rvert \lvert\mathcal{N}_j\rvert}}$ and $\operatorname{UP}(\cdot)=\sigma\big(\mathbf{W}\frac{\mathbf{h}_i}{\mathcal{N}_i} + \mathbf{W}\mathbf{m}_{\mathcal{N}_i}\big)$.

Given the graph data $\mathcal{G}=\{\mathcal{V}, \mathcal{E}, \mathbf{X}\}$, we can utilize the aforementioned message-passing framework to get node representations.  A general approach is that we adopt a GNN encoder $f$ to learn node representation as $\mathbf{H} = f(\mathbf{A}, \mathbf{X})$, which can be used for downstream tasks. For a node-level task, we aim at learning an encoder $f: \mathbb{R}^{\lvert\mathcal{V}\rvert \times\lvert\mathcal{V}\rvert} \times \mathbb{R}^{\lvert\mathcal{V}\rvert \times d} \rightarrow \mathbb{R}^{\lvert\mathcal{V}\rvert \times q}$, which computes the representation $\mathbf{H} \in \mathbb{R}^{\lvert\mathcal{V}\rvert \times q}$ for all the nodes. Here $d$ denotes the input feature dimension and $q$ is the node representation dimension. For graph-level task, we hope to learn a graph-level encoder $f:\mathbb{R}^{\lvert\mathcal{V}\rvert \times\lvert\mathcal{V}\rvert} \times \mathbb{R}^{\lvert\mathcal{V}\rvert \times d} \rightarrow \mathbb{R}^q$ as the representation of the whole graph. A commonly-used setting is constructing the graph-level encoder with a node-level encoder and a readout function to get graph representations~\cite{zhang2018an}. An example of the readout function can be computing the sum of all node representations as the graph representation, i.e., $\mathbf{h}_\mathcal{G} = \sum_{v \in \mathcal{V}} \mathbf{h}_i$, where $\mathbf{h}_\mathcal{G}$ is the graph representation and $\mathbf{h}_i$ indicates the node representation for each node $v \in \mathcal{V}$. In this survey, we mainly focus on counterfactual learning on graphs with graph neural networks. 

\subsection{Causal Inference in Machine Learning}

Typically, machine learning algorithms heavily rely on the independent and identically distributed (i.i.d.) assumption~\cite{bishop2007pattern} and are trained using standard empirical risk minimization. Recent works~\cite{liu2021just,sagawa2019distributionally} have shown that such machine learning algorithms can be negatively affected by spurious correlations or dependencies between observed features and class labels that only hold for certain data groups. For instance, in image classification tasks, images labeled as camels often have sand as the background, and images labeled as fish usually have water as the background. Consequently, the empirical risk minimization classifier may rely on the spurious background pattern for predictions rather than the objects in the images. However, this spurious correlation does not indicate causality~\cite{guo2020a} and may severely limit the robustness, explainability, and generalization ability of machine learning models~\cite{arjovsky2019invariant}. For example, such a model may misclassify cows on sand as camels or birds flying over water as fish during testing. Hence, there is a growing trend to incorporate causal inference in machine learning methods to learn causal features and improve generalization across different testing distributions. For instance, in an animal classification task, a classifier that focuses on the camel pattern rather than the sand pattern for classification can perform well for camels with various backgrounds~\cite{mitrovic2021representation}.

\begin{table}
\centering
\caption{A simple taxonomy on causal models.}
\resizebox{1\columnwidth}{!}{
\begin{tabular}{lccc}
\toprule
\textbf{ Model } & \textbf{ i.i.d. } & \textbf{ Intervention } & \textbf{ Counterfactual } \\
\midrule  Physical  &  yes  &  yes  &  yes  \\
  Structural causal  &  yes  &  yes  &  yes  \\
  Causal graphical  &  yes  &  yes  &  no  \\
  Statistical  &  yes  &  no  &  no  \\
\toprule
\end{tabular}
}
\label{table: models}
\end{table}

Compared with traditional statistical models, casual models have better generalization ability in modeling real-world systems~\cite{arjovsky2019invariant}. According to Pearl's Causal Hierarchy~\cite{pearl2009causal} and the description about levels of causal modeling in~\cite{scholkopf2021towards}, there are four levels of models to represent the real-world systems as shown in Table~\ref{table: models}. The \textit{physical model} of a set of coupled differential equations is a quite comprehensive description of a system. Differential equations allow us to predict the future behavior of a physical system and give us physical insights of the functioning~\cite{scholkopf2021towards}. Compared with the physical models, a \textit{statistical model} is quite superficial since it can only model the correlation between variables and when the input distribution changes the model performance may degrade dramatically~\cite{wang2021generalizing}. \textit{Causal models} are between these two levels of models. When we have a graphical model with all the edges as causal edges endowed with the notion of direct causal effect, the model would be a causal graphical model and allows us to perform interventions on variables. Learning from interventions can help models learn invariant mechanisms against distribution shifts~\cite{scholkopf2022from}. Go beyond the causal graphical model, a \textit{structural causal model} (SCM) is composed of a set of causal variables and a set of structural equations with noise distributions~\cite{shanmugam2018elements}. With SCMs, we can not only get the interventional distribution but also perform counterfactual reasoning~\cite{pearl2009causal}. As expressed in~\cite{pearl2009causal}, we can replace the parents-child relationship $p\left(X_i \mid \mathbf{P A}_i\right)$ with its functional counterpart $X_i=f_i\left(\mathbf{P A}_i, U_i\right)$ in Bayesian networks to get the structural causal models, where $\mathbf{P A}_i$ denotes the parent variables of $X_i$. The definition of the structural causal model is given as: 
\begin{definition}[Structural Causal Model (SCM)~\cite{scholkopf2022from}] 
To describe the causal relationship between a set of $n$ random variables $\mathcal{X} = \{X_1, \ldots, X_n\}$, a SCM $\mathcal{M}=\left(\mathcal{F}, p_{\mathcal{U}}\right)$ consists of $(i)$ a set $\mathcal{F}$ of $n$ assignments (the structural equations):
\begin{equation}
\fontsize{9pt}{10pt}\selectfont
\mathcal{F}=\left\{X_i:=f_i\left(\mathbf{P} \mathbf{A}_i, U_i\right)\right\}_{i=1}^n,    
\end{equation}
where $f_i$ is a deterministic function which is used for computing each variable $X_i$ from its causal parents $\mathbf{P} \mathbf{A}_i \subseteq \mathcal{X} \backslash\left\{X_i\right\}$ and an exogenous noise variable $U_i$. The assignment symbol ``:='' stands for defining, which is used to indicate the asymmetry of the causal relationship; and (\textit{ii}) a joint distribution $p_\mathcal{U}\left(U_1, \ldots, U_n\right)$ over the exogenous noise variables. 
\end{definition} 
The definition of interventional distribution in SCM framework is quite natural. To model an intervention, we can replace the corresponding structural equation and consider the resulting entailed distribution~\cite{scholkopf2021towards}. 
Counterfactuals are quite different from observations and interventions. Observations are used to describe what is passively seen or measured and interventions are used to describe active external manipulation or experimentation; while counterfactuals indicate what would or could have been, given that something else was in fact observed~\cite{scholkopf2022from}. The formal definition of counterfactuals is given as:
\begin{definition}[Counterfactuals in SCMs~\cite{scholkopf2022from}]
Given evidence $\mathbf{X}=\mathbf{x}$ observed from a SCM $\mathcal{M}=\left(\bar{\mathcal{F}}, p_\mathcal{U}\right)$, the counterfactual SCM $\mathcal{M}^{\mathbf{X}=\mathbf{x}}$ is obtained by updating $p_\mathcal{U}$ with its posterior: $\mathcal{M}^{\mathbf{X}=\mathbf{x}}=\left(\mathcal{F}, p_{\mathcal{U} \mid \mathbf{X}=\mathbf{x}}\right)$. Counterfactuals are then computed by performing interventions in the counterfactual SCM $\mathcal{M}^{\mathbf{X}=\mathbf{x}}$. The intervention $d o\left(X_i:=x_i\right)$ is modeled by replacing the $i^{\text {th }}$ structural equation in $\mathcal{F}$ by $X_i:=x_i$, yielding the intervened SCM $\mathcal{M}^{do\left(X_i:=x_i\right), \mathbf{X=\mathbf{x}}}=\left(\mathcal{F}^{\prime}, p_{\mathcal{U} \mid \mathbf{X}=\mathbf{x}}\right)$. The interventional distribution $p\left(\mathbf{X}_{-i} \mid d o\left(X_i=x_i\right)\right)$, where $\mathbf{X}_{-i}=\mathbf{X} \backslash\left\{X_i\right\}$, and intervention graph $\mathcal{G}^{\prime}$ are those induced by $\mathcal{M}^{do\left(X_i:=x_i\right), \mathbf{X=\mathbf{x}}}.$
\end{definition} 
Here $do(X_i:=x_i)$ is to emphasize the difference between passive observation and active intervention, denoting an intervention by which variable $X_i$ is set to value $x_i$~\cite{pearl2009causal}.

Counterfactual learning is a learning scheme with the aid of counterfactuals. Causal inference community focuses on estimating the counterfactuals~\cite{pearl2009causal}. Machine learning community leverages counterfactual estimation as one module of the model~\cite{ma2022learning, lucic2022cf}, and designs different modules besides the counterfactual estimation module to achieve the goal of fairness~\cite{ma2022learning}, interpretation~\cite{lucic2022cf} or link prediction and recommender systems~\cite{zhao2022learning}. 

\section{Counterfactual Fairness on Graphs}\label{fairness}
In many real-world applications, training data may contain demographic biases due to societal or historical factors. Machine learning models trained on such data can inherit these biases and produce unfair predictions, as seen in credit card decision-making tasks where models can exhibit gender or race biases~\cite{mehrabi2021a,chen2018fair}. Biased predictions can result in systemic discrimination and undermine public trust in machine learning models, which has led to growing interest in fairness-aware machine learning~\cite{mehrabi2021a}. \citeauthor{dai2022a}~\cite{dai2022a} highlights that fairness issues are more severe in graphs, where both node features and graph structure can contribute to bias. As many real-world systems rely on graph neural networks, unfair predictions can have serious consequences. Researchers have proposed methods to ensure fairness in graph learning~\cite{dai2021say, dong2022fairness, ma2022learning, agarwal2021towards, dai2022learning}. However, existing fairness notions are predominantly correlation-based and may not detect statistical anomalies such as Simpson's paradox~\cite{makhlouf2020survey, 10.1145/3458509, 10.1145/3547333, 10.1145/3564284}. To address this issue, counterfactual learning has emerged as a promising approach for building fair machine learning models that achieve counterfactual fairness~\cite{kusner2017counterfactual, makhlouf2020survey}. Counterfactual fairness is based on the idea that a prediction for an individual is fair if it remains the same in a counterfactual world where the individual belongs to a different demographic group~\cite{kusner2017counterfactual}. Counterfactual fairness on graphs is attracting increasing attention~\cite{ma2022learning, zhang2021multi, agarwal2021towards, kose2021fairnessaware}. In this section, we introduce the background of fairness issues in graph learning and counterfactual fairness, followed by a general framework of graph counterfactual fairness models and their details. We conclude with widely used fairness evaluation metrics and datasets.

\subsection{Background of Graph Counterfactual Fairness}

\subsubsection{Sources of Biases}
Various biases are prevalent in real-world data sources, and machine learning models trained on biased datasets can easily produce unfair predictions. Bias in machine learning can occur at different stages, such as data, algorithm, and user interaction~\cite{mehrabi2021a, zhu2022learning}. We can categorize commonly observed bias in i.i.d. data into four categories: historical bias, representation bias, temporal bias, and attribute bias~\cite{dai2022a}. Specifically, \textit{historical bias} refers to a bias in data due to historical reasons, such as gender or race bias. \textit{Representation bias} denotes the discrepancy between collected data points and real-world data points, which arises from the under-representation of the collected data. \textit{Temporal bias} results from changes over time. These biases are widely present in image datasets~\cite{jalal2021fairness}, text datasets~\cite{garg2019counterfactual}, and video datasets~\cite{nathan2019endtoend}. \textit{Similar to i.i.d data, graphs also exhibit the aforementioned biases in their node attributes}. 

In addition to the biases in node attributes, graph-structured data also has distinctive biases due to topology structures such as the linking bias and the structural bias~\cite{dai2022a}. \textit{Linking bias} happens since nodes have intrinsic preferences to connect with others, e.g., nodes of similar ages are more likely to connect to each other. \textit{Structural bias} is caused due to the information propagation between nodes on graph-structure data. The information propagation can amplify the sensitive information on each node~\cite{mehrabi2021a}. In real-world networks, nodes with the same sensitive attribute (e.g., ages) are more likely to get connected. For instance, young people tend to make friends with people of similar ages on social networks. The message-passing of GNNs will further magnify this bias due to the smoothing nature of GNN, i.e., representations of nodes of the same sensitive attribute will be more similar while representations of nodes of different sensitive attributes will be more different after the aggregation. It will make the predictions for downstream tasks highly correlated with sensitive attributes because the learned node representations are highly dependent on the sensitive attributes of nodes~\cite{dai2022a}. GNNs are commonly used on graph-structure data and the message-passing framework of GNNs can mix the biases stored in node attributes and graph structures together into the node representations~\cite{dong2022fairness}, which makes the bias issue more severe.

\subsubsection{Definition of Graph Counterfactual Fairness}
In machine learning area, biases are measured from different perspectives. Group fairness and individual fairness are two commonly used algorithmic bias metrics~\cite{dong2022edits}. Group fairness aims to ensure that any demographic group should not get discriminatory treatment. Here a group means a set of people that have the same sensitive attribute, e.g., people of the same gender. Instead of focusing on demographic statistics, individual fairness aims to ensure that two similar individuals should receive similar algorithmic treatment~\cite{dai2022a}. For example, two people who have similar credit records should get a similar credit loan.

\citeauthor{kusner2017counterfactual}~\cite{kusner2017counterfactual} find that group fairness and individual fairness have some issues of incompatibility and they only focus on subpopulations or individuals in the actual world. These fairness notions rely on the statistical correlation among variables, which may not describe the intrinsic causal structure~\cite{bickel1975sex}. They also fail to capture the discrepancy when an individual belonging to the actual world and a counterfactual world. For instance, in real-world, people tend to come up with counterfactual questions, like ``What if I am a girl, will I be admitted to xxx university?''. A system that cannot answer counterfactual questions and fulfill counterfactual fairness is not reliable and can severely damage users' experience~\cite{makhlouf2020survey}. Inspired by the development of counterfactual learning, counterfactual fairness is proposed as a notion to measure fairness from the perspective of causal inference~\cite{kusner2017counterfactual}.  Counterfactual fairness makes the outputs of a machine learning model for an individual in the actual world remain the same when we flip the sensitive attribute of the same individual to the counterfactual world. We give an example to illustrate the notion of counterfactual fairness. Many companies use machine learning and graph neural networks to assign credit loans to applicants. However, the model may give unfair predictions due to the bias in the training data. For example, a model may predict that a white person should get a large loan based on his/her features; while when we flip the person's ethnicity to black, the model gives the decision that the person can only get a small loan, which is counterfactually unfair. In this example, we do not compare the individual's outcome with other groups of people or any other individuals but just the same individual in a counterfactual world. Following~\cite{ma2022learning}, we extend the formal definition of graph counterfactual fairness as:
\begin{definition}[Graph Counterfactual Fairness~\cite{ma2022learning}]
An encoder $\Phi(\cdot)$ or classifier $\phi(\cdot)$ satisfies graph counterfactual fairness if for any node $v_i$: 
\begin{multline}
\fontsize{9pt}{10pt}\selectfont
P((\mathbf{z}_i)_{S \leftarrow s}\vert\mathbf{X}, \mathbf{A}) = 
P((\mathbf{z}_i)_{S \leftarrow s^\prime} \vert \mathbf{X}, \mathbf{A})\quad \text{or}\\
 \quad P((\hat{y}_i)_{S \leftarrow s}\vert\mathbf{X}, \mathbf{A}) = 
P((\hat{y}_i)_{S \leftarrow s^\prime} \vert \mathbf{X}, \mathbf{A}) \quad \forall s \neq s^\prime,
\end{multline}

where $s, s^\prime \in \{0,1\}^n$ are two arbitrary sensitive attribute values of all nodes. $\mathbf{z}_i=(\Phi(\mathbf{X}, \mathbf{A}))_i$ denotes the node representations for node $v_i$ and $\hat{y}_i =\phi(\mathbf{z}_i)$ denotes the predicted label for node $v_i$. In other words, given a graph $\mathcal{G}=\{\mathbf{A}, \mathbf{X}\}, \Phi(\cdot)$ should be optimized by minimizing the distribution discrepancy between the representations $(\Phi(\mathbf{X}_{S \leftarrow s}, \mathbf{A}_{S \leftarrow s}))_i$ and $(\Phi(\mathbf{X}_{S \leftarrow s^{\prime }}, \mathbf{A}_{S \leftarrow s^{\prime}}))_i$. $\phi(\cdot)$  should be optimized by minimizing the distribution discrepancy between the predictions $(\phi(\mathbf{X}_{S \leftarrow s}, \mathbf{A}_{S \leftarrow s}))_i$ and $(\phi(\mathbf{X}_{S \leftarrow s^{ \prime}}, \mathbf{A}_{S \leftarrow s^{ \prime}}))_i$.
\label{def:fairness}
\end{definition} 
Here we use $\mathbf{X}_{S \leftarrow s^{\prime}}$ to denote a specific value of the counterfactual, i.e., ``what would the node feature have been if the sensitive attribute of the nodes had been set as $s^\prime$, given the original node features $\mathbf{X}$ and graph structure $\mathbf{A}$.'' $\mathbf{A}_{S \leftarrow s^{\prime}}$ is also got in a similar manner~\cite{ma2022learning}.

\subsection{Methods of Graph Counterfactual Fairness}

To eliminate the bias in graph-structure data, extensive efforts have been put on building fairness-aware GNNs~\cite{dai2022a}. Generally, existing debiasing methods can be roughly categorized into three categories, i.e., adversarial debiasing methods~\cite{masrour2020bursting, bose2019compositional}, fairness constrain methods~\cite{li2021on, wang2022unbiased, dai2021say}, counterfactual-based methods~\cite{kose2021fairnessaware} and other methods, such as random walk-based method~\cite{khajehnejad2022crosswalk, rahman2019fairwalk} and so on. Specifically, \textit{adversarial debiasing methods}~\cite{masrour2020bursting, bose2019compositional} utilize adversarial learning to learn node representations that are independent of the sensitive attribute. \textit{Fairness constrain methods}~\cite{li2021on, wang2022unbiased, dai2021say} directly add carefully designed fairness constraints such as the covariance between sensitive attributes and predicted labels to the objective function to achieve fairness. For a comprehensive overview of these debiasing techniques, please refer to~\cite{dai2022a}. In this survey, we focus on \textit{counterfactual-based methods}, which leverage counterfactual reasoning to ensure GNN fairness.  Counterfactual fairness on graphs overcomes the inherent limitations of correlation-based fairness notions such as group fairness. However, counterfactual fairness introduces new challenges, including constructing causal relationships among variables to obtain counterfactuals and achieving counterfactual fairness with observed data and estimated counterfactuals. Additionally, counterfactual fairness in graph-structured data presents unique technical challenges. The nodes in graphs are interconnected, leading to a non-independent and non-identically distributed nature in data generation. In this section, to provide insight into counterfactual learning on graphs, we first introduce a general framework for achieving counterfactual fairness on graphs. Following that, we will discuss representative and state-of-the-art works in detail.

\begin{figure*}
    \centering
    \includegraphics[width=0.8\linewidth]{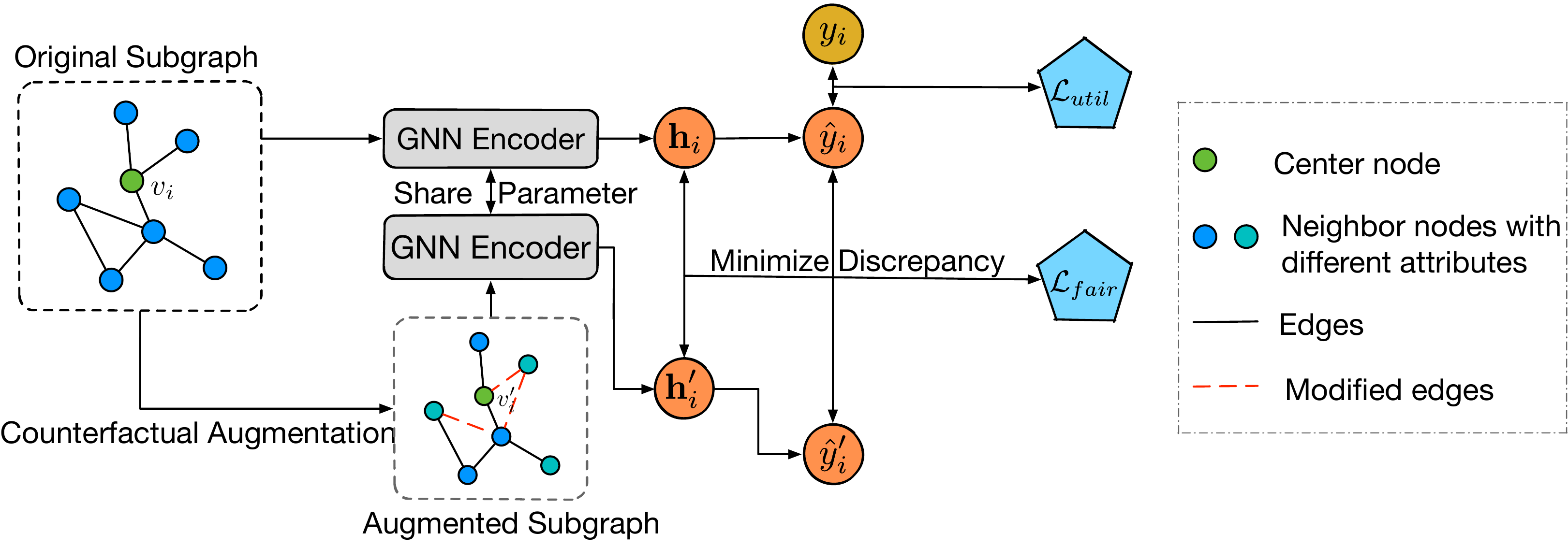}
    \caption{Overall framework of graph counterfactual fairness.}
    \label{fig:fair_frame}
    \vskip -1em
\end{figure*}
\subsubsection{General Framework of Counterfactual Fairness}

Inspired by the advances of counterfactual learning, several methods have been proposed to achieve counterfactual fairness on graph-structure data~\cite{dong2022edits,ma2022learning, zhang2021multi, agarwal2021towards, kose2021fairnessaware,Guo0X0W23}. 
In general, most of these methods can be unified into a two-step framework designed to address the aforementioned challenges. Figure~\ref{fig:fair_frame} illustrates the general framework, using node $v_i$ as an example.
\textit{First}, counterfactual augmentations are generated by flipping the sensitive attributes and related information of observed data.
\textit{Second}, both observed factual and generated counterfactual samples are fed into graph neural networks to obtain factual and counterfactual representations while minimizing the discrepancy between them to achieve counterfactual fairness.

\textbf{Counterfactual Augmentation}
In graph counterfactual fairness, we want to mitigate the bias caused by the sensitive information, which means the counterfactuals should have the opposite sensitive information. Thus, a straightforward way is to flip the sensitive attributes of observed samples to get the counterfactuals. However, this method only works when we have the assumption that the sensitive attributes do not have any causal effect on the generation of other attributes and graph structures, which is not valid in most situations. To solve this issue, various augmentation methods~\cite{anonymous2023learning, ding2022data, jiang2022fmp, agarwal2021towards} to model the dependency between graph structures and sensitive attributes are proposed. Here, we summarize these augmentation methods into a unified framework. 

Let $\mathbf{s}=\{s_{i}\}_{i=1}^{n}$ represent the vector of sensitive attributes, where $s_i \in \{0,1\}$ is the sensitive attribute of node $v_i$. 
Let $\mathbf{X}_{S \leftarrow s^\prime}$ denote the the counterfactual features when the sensitive attributes are flipped to $s^\prime$. 
This corresponds to ``what would the node features have been if the sensitive attributes of the nodes had been set as $s^\prime$, given the original node features $\mathbf{X}$ and graph structure $\mathbf{A}$?''.
Thus, we can represent the actual counterfactual graph as $(\mathbf{A}_{\mathbf{S} \leftarrow s^\prime},\mathbf{X}_{\mathbf{S} \leftarrow s^\prime})$, where $s^\prime$ represents the flipped sensitive attributes of $s$. Existing counterfactual fairness methods use augmentations to approximate the counterfactual graph. Considering the counterfactual augmentation as $\mathcal{T}(\cdot)$, which can modify the graph structure or feature matrix, we define the approximated counterfactual graph as $\mathcal{G}^\prime=(\mathbf{A}^\prime, \mathbf{X}^\prime)$. For simplicity, in the following sections, we use the term ``counterfactual graph'' to denote the approximated counterfactual graph. The counterfactual graph is obtained by:
\begin{equation}
\fontsize{9pt}{10pt}\selectfont
    \mathbf{A}^\prime,\mathbf{X}^\prime=\mathcal{T}(\mathbf{A},\mathbf{X}),
\label{eq:4},
\end{equation}
where $\mathcal{T}(\cdot)$ denotes the counterfactual augmentation that flips sensitive attributes and sensitive-related information, e.g., $\mathcal{T}(\cdot)$ can be just flipping the sensitive attribute of a node or flipping the sensitive attributes of neighbor nodes in each ego-graph.

\textbf{Counterfactual Regularization}
For graph counterfactual fairness, we want to minimize the discrepancy of the node representations for factual graph $\mathcal{G}$ and counterfactual graph $\mathcal{G}^\prime$, which makes the representation independent of sensitive attributes. 
Many works~\cite{ma2022learning, agarwal2021towards} adopt the Siamese neural networks to learn fair node representations as:
\begin{equation}
\fontsize{9pt}{10pt}\selectfont
    \mathbf{H}=f_\theta (\mathbf{A}, \mathbf{X}), \quad \mathbf{H}^\prime=f_\theta (\mathbf{A}^\prime, \mathbf{X}^\prime), 
    \label{eq:5},
\end{equation}
where $f_\theta$ denotes the graph neural network with parameter $\theta$. $\mathbf{H}$ is the factual node representation and $\mathbf{H}^\prime$ stands for the counterfactual node representation. To minimize the discrepancy between factual and counterfactual node representations, the objective function can be written as:
\begin{equation}
\fontsize{9pt}{10pt}\selectfont
    \mathcal{L}_{\text{fairness}} = \operatorname{dist}(\mathbf{H}, \mathbf{H}^\prime),
\end{equation}
where $dist(\cdot,\cdot)$ is the distance measure such as the negative cosine similarity and L2 distance~\cite{agarwal2021towards,ma2022learning}.

\textbf{Overall Objective}
For graph counterfactual fairness models, the overall objective function can be formalized as:
\begin{equation}
\fontsize{9pt}{10pt}\selectfont
    \min _{\theta} \mathcal{L}_{\text {utility }}+\beta \mathcal{L}_{\text {fairness }},
\end{equation}
where $\beta$ is the hyper-parameter to control the trade-off between utility and fairness. $\mathcal{L}_{\text {utility}}$ is the loss function for the utility of the model such as node classification loss and graph classification loss. $\mathcal{L}_{\text{fairness}}$ is the graph counterfactual fairness constrain $\mathcal{L}_{\text{fairness}}=\operatorname{dist}(f_\theta (\mathbf{A}, \mathbf{X}), f_\theta (\mathcal{T}(\mathbf{A}, \mathbf{X})))$. 
 $\theta$ is the set of model parameters to be learned. 

\subsubsection{Methodologies} \label{fair_method}

\begin{table*}
    \scriptsize
    \centering
    \caption{Summary of graph counterfactual fairness papers.}     \label{tab:debias}
    \begin{adjustbox}{width=\textwidth}
    \begin{tabular}{lcccccc}
    \toprule
    \textbf{Methods}  & \textbf{Edge Aug} & \textbf{Feature Aug} & \textbf{Learnable Mask}  & \textbf{Stage} & \textbf{Subjects} & \textbf{Causal Graph}   \\
    \midrule
    GEAR~\cite{ma2022learning}& \checkmark & \checkmark & \checkmark &  In-processing & Subgraph & Yes \\
    \midrule
    MCCNIFTY~\cite{zhang2021multi} & \checkmark & \checkmark & \checkmark & In-processing & Subgraph & No \\
    \midrule
    NIFTY~\cite{agarwal2021towards} & \checkmark & \checkmark & $\times$ & In-processing & Whole graph & No \\
    \midrule
    Fairness-Aware~\cite{kose2021fairnessaware} & \checkmark & \checkmark & \checkmark & In-processing & Whole graph & No \\
    \midrule
    CAF~\cite{Guo0X0W23} & $\times$ & $\times$ & $\times$ & In-processing & Subgraph & Yes \\
    
    \bottomrule
    \end{tabular}
    \end{adjustbox}
\end{table*}

The aforementioned general framework can be easily extended to different scenarios. For instance, in the augmentation step, 
different kinds of augmentation techniques can be applied, such as random perturbation~\cite{agarwal2021towards}, parameterized augmentation~\cite{ma2022learning}, heuristic-based augmentation~\cite{agarwal2021towards, anonymous2023learning, ding2022data} and data-based~\cite{Guo0X0W23}. 
For the counterfactual regularization, various distance measures on distributions can be used, such as triplet loss function~\cite{ma2022learning}. The utility function can be the classification or regression losses, depending on downstream tasks~\cite{agarwal2021towards, agarwal2022probing}. Next, we will introduce details of representative graph counterfactual fairness methods. Table~\ref{tab:debias} summarizes existing graph counterfactual fairness papers.

\textbf{NIFTY~\cite{agarwal2021towards}}
To build trustworthy models on graph data, NIFTY~\cite{agarwal2021towards} tries to establish a key connection between fairness and stability and learn fair and stable node representations at the same time. It defines three kinds of perturbations on node attributes, sensitive attributes and graph structure. In NIFTY, the counterfactual augmentation is obtained as:
\begin{equation}
\fontsize{9pt}{10pt}\selectfont
\mathbf{A}^\prime,\mathbf{X}^\prime=\mathcal{T}(\mathbf{A}, \mathbf{X})=\mathbf{A}, \mathbf{X}_{\text {flip} (\mathbf{s})} ,
\end{equation}
where $\mathcal{T}(\cdot)$ denotes the counterfactual augmentation. NIFTY just flips the sensitive attributes $\mathbf{s}$ in the feature matrix $\mathbf{X}$. Then, together with the randomized perturbation on node attributes and graph structure, different views are fed into GNN-based encoder to get node representations. Then, NIFTY adopts a triplet-based objective to maximize the similarity between the original representation and its augmented (i.e., counterfactual and noisy) representations to ensure that the representations are independent of sensitive attributes. The loss function can be written as:
{\small %
\begin{multline}
% \fontsize{8pt}{10pt}\selectfont
    \mathcal{L}_{\text{total}} = \mathcal{L}_{\text{utility}} + \mathcal{L}_{\text{fairness}} = 
    \frac{1}{\lvert\mathcal{V}_L\rvert} \sum_{v_i \in \mathcal{V}_L} l(f_{\Phi} (\mathbf{h}_i), y_i) \\
    + \frac{1}{2\lvert\mathcal{V}\rvert} \sum_{v_i \in \mathcal{V}} 
    \big( \operatorname{dist}(t_\phi(\mathbf{h}_i), \operatorname{sg}(\mathbf{h}_i^\prime)) + \operatorname{dist}(t_\phi(\mathbf{h}_i^\prime), \operatorname{sg}(\mathbf{h}_i)) \big) \nonumber
\end{multline}

}where $f_{\Phi}(\cdot)$ is the prediction head, i.e., $\hat{y}_i=f_{\Phi}(\mathbf{h}_i)$. $t_\phi(\cdot)$ denotes an additional projection head. $\operatorname{sg}(\cdot)$ stands for the stop-gradient operation. Here $\operatorname{dist}(\cdot,\cdot)$ is cosine distance.

\textbf{GEAR~\cite{ma2022learning}} 
Many existing works neglect the causal interaction between a node and its neighbors~\cite{agarwal2021towards, zhang2021a}. Thus, \citeauthor{ma2022learning}~\cite{ma2022learning} propose a framework GEAR to generate counterfactual data with the causal interaction between nodes and their neighborhood into consideration. Concretely, they utilize self-perturbation and neighbor-perturbation to flip the sensitive information in each node's ego-graph to generate counterfactuals.
Self-perturbation just flips the sensitive attribute of the center node. Neighbor-perturbation modifies the sensitive attribute of neighbor nodes. Aligned with our unified framework, the augmentation process can be modeled as: 
\begin{equation}
\fontsize{9pt}{10pt}\selectfont
\mathbf{A}^\prime,\mathbf{X}^\prime=\mathcal{T}_{\text {VGAE}}(\mathcal{T}_{\text {perturb}}(\mathbf{A}, \mathbf{X}) ),
\end{equation}
where $\mathcal{T}_{\text {perturb}}$ represents self-perturbation $\mathcal{T}_{\text {self}}$ and neighbor-perturbation $\mathcal{T}_{\text {neighbor}}$. These perturbations can help to find counterfactuals which have a different sensitive attribute in itself or neighbors.
 $\mathcal{T}_{\text {VGAE}}$ is a variational graph auto-encoder which help to model the dependencies in the latent representation level so that the model can flip the causally-related information in ego-graphs. Concretely, the two kinds of counterfactuals can be generated by:
\begin{equation}
\fontsize{9pt}{10pt}\selectfont
\begin{aligned}
\mathbf{X}_{S_i=s^\prime}&= \mathcal{T}_{\text {self}}(\mathbf{X}), \\
\quad (\mathbf{A}^\prime,\mathbf{X}^\prime) &= \mathcal{T}_{\text {VGAE}}(\mathbf{A}, \mathbf{X}_{S_i=s^\prime}),  \\
\mathbf{X}_{\mathcal{N}_i=s^\prime}&= \mathcal{T}_{\text {neighbor}}(\mathbf{X}), 
\\
    (\mathbf{A}^{\prime\prime},\mathbf{X}^{\prime\prime}) &= \mathcal{T}_{\text {VGAE}}(\mathbf{A}, \mathbf{X}_{\mathcal{N}_i=s^\prime}).
\end{aligned}
\end{equation}
With the sensitive-information-flipped counterfactuals, GEAR minimizes the discrepancy between factual representation and counterfactual representation so that the model can find the shared information for observed data and counterfactuals. The fairness regularization can be written as:
\begin{equation}
\fontsize{8pt}{10pt}\selectfont
\begin{aligned}
    \mathcal{L}_{\text{fairness}} = \frac{1}{N} {\sum}_{i = 1}^N \left(\left(1-\lambda_s\right) d\left(\mathbf{h}_i, {\mathbf{h}}_i^\prime\right)+\lambda_s d\left(\mathbf{h}_i, {\mathbf{h}}_i^{\prime \prime}\right)\right),
\end{aligned}
\end{equation}
where $\mathbf{H}=f_\theta (\mathbf{A}, \mathbf{X}), \mathbf{H}^\prime=f_\theta (\mathbf{A}^\prime,\mathbf{X}^\prime), \mathbf{H}^{\prime \prime}=f_\theta (\mathbf{A}^{\prime \prime}, \mathbf{X}^{\prime \prime})$. The obtained representation can be used for various kinds of downstream tasks. For node classification, the utility loss function is:
\begin{equation}
\fontsize{9pt}{10pt}\selectfont
    \mathcal{L}_{\text{utility}} = \frac{1}{\lvert\mathcal{V}_L\rvert} \sum_{v_i \in \mathcal{V}_L} l(f_{\Theta} (\mathbf{h}_i), y_i),
\end{equation}
where $\mathcal{V}_L \in \mathcal{V}$ is the set of labeled nodes. $f_{\Theta}(\cdot)$ is parameterized with $\theta^\prime$ and make predictions for node classification task, i.e., $\hat{y}_i=f_{\Theta}(\mathbf{h}_i)$. $l(\cdot, \cdot)$ is the cross entropy loss function.

\textbf{MCCNIFTY}~\cite{zhang2021multi}
Building upon NIFTY~\cite{agarwal2021towards}, MCCNIFTY~\cite{zhang2021multi} employs the same perturbation techniques but with a focus on subgraph-level processing to improve computational efficiency. Specifically, MCCNIFTY first utilizes the personalized PageRank algorithm (PPP) to extract informative subgraphs for each node, and then applies perturbations to these subgraphs. Using these subgraphs and their counterfactual counterparts, MCCNIFTY designs a multi-view uncertainty-aware node embedding learning module inspired by evidential theory. This approach has the advantage of estimating uncertainty scores that help control the information propagation process of GNN, allowing the model to adaptively adjust edge attention and defend against adversarial attacks~\cite{agarwal2021towards}. With the uncertainty-aware GNN encoder, MCCNIFTY follows NIFTY's pipeline and designs a loss function to maximize the similarity of multi-view node embeddings. The primary distinctions between MCCNIFTY and NIFTY are: 1) MCCNIFTY focuses on subgraph-level processing for enhanced efficiency, and 2) MCCNIFTY adopts an uncertainty quantification module to help defend against adversarial attacks.

\textbf{Fairness-Aware}~\cite{kose2021fairnessaware}
As most existing counterfactual augmentation approaches simply adopt the randomized perturbation on feature matrix and adjacency matrix, ~\citeauthor{kose2021fairnessaware}~\cite{kose2021fairnessaware} propose to leverage the knowledge from the data to adaptively augment the features and graph structures. 
As GNNs can propagate and amplify biases in graph-structured data, adaptive augmentation under suitable supervision can help to reduce possible bias inherited in a graph structure and node features. 
Concretely, the adaptive counterfactual augmentation can be modeled as:
\begin{equation}
\fontsize{9pt}{10pt}\selectfont
\begin{aligned}
\mathbf{A}^\prime,\mathbf{X}^\prime &= \mathcal{T}(\mathbf{A}, \mathbf{X}) \\
&= \text{DropEdge}(\mathbf{A}) , \text{MaskFeat}(\mathbf{X})
\end{aligned}
\end{equation}

where $\text{MaskFeat}(\mathbf{X}) = \left\{\mathbf{m} \circ \mathbf{x}_1 , \ldots , \mathbf{m} \circ \mathbf{x}_N\right\}$ and $\mathbf{m}\in\{0,1\}^{F}$ is the mask vector. Each entry $m_i$ of $\mathbf{m}$ is drawn independently from a Bernoulli distribution determined by the correlation score between the nodal features and the sensitive attributes of the node. 
The intuition is that features with a high correlation with the sensitive attribute will tend to induce an unfair prediction. 
$\text{DropEdge}(\mathbf{A})$ assigns different edge-dropping probabilities for edges to get different views. For an edge $e_{ij}$, the adaptive edge deletion probabilities are designed as:
\begin{equation}
\fontsize{9pt}{10pt}\selectfont
\begin{aligned}
p\left(e_{i j}^{(1)}\right) &= 
\begin{cases}
p_1, & \text{ if } s_i=s_j \\
p_2, & \text{ if } s_i \neq s_j
\end{cases} \\
p\left(e_{i j}^{(2)}\right) &= 
\begin{cases}
p_3, & \text{ if } s_i \neq s_j \\
p_4, & \text{ if } s_i = s_j
\end{cases}
\end{aligned}
\end{equation}

where the superscript $(1)$ and $(2)$ are used to denote augmented view 1 and view 2, respectively. $p\left(e_{i j}^{(1)}\right)$ and $p\left(e_{i j}^{(2)}\right)$ are the deletion probabilities for the edge $e_{ij}$ in two views.  The probabilities should follow $p_1 > p_2$ and $p_3 < p_4$ so that the edges connecting two nodes of the same sensitive attribute have a larger probability to be deleted than edges that connect two nodes of different sensitive attributes, which can alleviate the structural bias of the graph.
With two different views of less bias, it performs contrastive learning  to learn node representation as:
% \begin{equation}
%     \ell\left(\mathbf{h}_i^1, \mathbf{h}_i^2\right)=-\log \frac{e^{s\left(\mathbf{h}_i^1, \mathbf{h}_i^2\right) / \tau}}{e^{s\left(\mathbf{h}_i^1, \mathbf{h}_i^2\right) / \tau}+\sum_{k=1}^N \mathbbm{1}_{[k \neq i]} e^{s\left(\mathbf{h}_i^1, \mathbf{h}_k^2\right) / \tau}+\sum_{k=1}^N \mathbbm{1}_{[k \neq i]} e^{s\left(\mathbf{h}_i^1, \mathbf{h}_k^1\right) / \tau}},
%     \label{eq:cont}
% \end{equation}
\begin{equation}
\fontsize{9pt}{10pt}\selectfont
    \ell\left(\mathbf{h}_i^1, \mathbf{h}_i^2\right) = -\log \frac{e^{s\left(\mathbf{h}_i^1, \mathbf{h}_i^2\right) / \tau}}{Z},
    \label{eq:cont}
\end{equation}

where $Z = \sum_{k=1}^N \mathbbm{1}_{[k \neq i]} e^{s\left(\mathbf{h}_i^1, \mathbf{h}_k^2\right) / \tau} +e^{s\left(\mathbf{h}_i^1, \mathbf{h}_i^2\right) / \tau}+ \sum_{k=1}^N \mathbbm{1}_{[k \neq i]} e^{s\left(\mathbf{h}_i^1, \mathbf{h}_k^1\right) / \tau}$. $s(\cdot, \cdot)$ is a score function, i.e., cosine similarity. $\tau$ is the temperature parameter and $\mathbbm{1}_{[k \neq i]} \in \{0,1\}$ is the indicator function and takes the value 1 if and only if $k \neq i$. The final fairness regularization can be written as $\mathcal{L}_{\text{fairness}}=\frac{1}{2 N} \sum_{i=1}^N\left[\ell\left(\mathbf{h}_i^1, \mathbf{h}_i^2\right)+\ell\left(\mathbf{h}_i^2, \mathbf{h}_i^1\right)\right]$ to be in a symmetric form, where $\ell(\cdot, \cdot)$ is the contrastive loss from Equation~\ref{eq:cont}.

\textbf{CAF}~\cite{Guo0X0W23} CAF is a little different from our framework. CAF claims that random perturbation or generative approach have their own drawbacks. It proposes to obtain counterfactuals within the training data using sensitive attributes and labels as guidance rather than random perturbation or generation which means that they do not follow~\ref{eq:4} and~\ref{eq:5}. They first use one GNN to learn latent representation $C$ and $E$.
\begin{equation}
\fontsize{9pt}{10pt}\selectfont
    [C,E] = H = f_\theta(A,X),
    \label{eq:CAF1}
\end{equation}
To make content representation $C$ informative to the downstream tasks, they introduce a classifier $f_\theta$.  The utility loss function is:
\begin{equation}
\fontsize{9pt}{10pt}\selectfont
    \mathcal{L}_{\text{utility}} = \frac{1}{\lvert\mathcal{V}_L\rvert} \sum_{v_i \in \mathcal{V}_L} l(f_{\Theta} (\mathbf{c}_i), y_i),
\end{equation}
In order for the information in the representation to be sufficient to reconstruct the observed graphs, CAF uses a sufficiency constraint:
\begin{equation}
\fontsize{7pt}{10pt}\selectfont
    \mathcal{L}_{\text{surf}} = \frac{1}{\lvert\mathcal{E}\rvert+\lvert\mathcal{E}^-\rvert}\sum_{(v_i,v_j)\in \mathcal{E}\cup\mathcal{E}^-} -e_{ij}logp_{ij}-(1-e_{ij})logp_{ij},
\end{equation}
where $\mathcal{E}^-$ is the set of sampled negative edges and $p_{ij}$ is the link existence probability between node $i$ and node $j$. CAF also uses orthogonal constraint to ensure that $c_i$ and $e_i$ are disentangled, i.e. $c_i^{T}e_i=0$.

To make environment representation $e_i$ be invariant to the content information existing in $c_i$, invariance constraint is defined as:
\begin{equation}
\fontsize{8.5pt}{10pt}\selectfont
\begin{aligned}
\mathcal{L}_{\text{inv}} = & \frac{1}{\lvert\mathcal{V}\rvert \cdot K}\sum_{v_i\in \mathcal{V}}\sum_{k=1}^{K}\lvert[dis(c_i,c_i^{e_k})+dis(e_i,e_i^{c_k})\rvert] \\
& + \gamma K \cdot \vert cos(c_i,e_i)\vert,
\end{aligned}
\end{equation}

The fairness regularization can be written as:
\begin{equation}
\fontsize{9pt}{10pt}\selectfont
\begin{aligned}
    \mathcal{L}_{\text{fairness}} =  \alpha\mathcal{L}_{\text{surf}} + \beta\mathcal{L}_{\text{inv}},
\end{aligned}
\end{equation}

\subsection{Taxonomy of Graph Counterfactual Fairness Methods}

In this subsection, we provide a detailed categorization of graph counterfactual fairness models. First, we categorize models based on the stage at which fairness is addressed. Then, we classify existing works according to research subjects and whether they model causal graphs or not. Table~\ref{tab:debias} summarizes this categorization.

\begin{itemize}[]
    \item{\bf Stages.} Based on the stage at which fairness is addressed, existing graph fairness models can be classified into pre-processing, in-processing, and post-processing methods~\cite{dai2022a}. It should be noted that, currently, no works fall into the pre-processing and post-processing methods categories. Most existing works are in-processing methods~\cite{ma2022learning, zhang2021multi, agarwal2021towards, kose2021fairnessaware,Guo0X0W23}. These methods utilize carefully designed GNNs to obtain fair node representations, which help minimize the discrepancy between factual and counterfactual node representations. Researchers have developed various loss functions to improve performance on different downstream tasks. 
    % For example, \citeauthor{agarwal2021towards}~\cite{agarwal2021towards} employ a loss function for node classification along with a loss function for graph counterfactual fairness, achieving strong performance in both node classification and counterfactual fairness simultaneously.
    
    \item{\bf Subjects.} Based on the augmentation subject, existing works can be categorized into subgraph-based methods and entire graph-based methods.
    \textit{Subgraph-based} methods~\cite{ma2022learning, zhang2021multi,Guo0X0W23} generate counterfactuals using target nodes' subgraphs. Due to the high cost of modeling causal relations on large graphs, GEAR~\cite{ma2022learning} samples ego-centric subgraphs for individual nodes, considering only causal relationships within subgraphs. It employs a personalized PageRank algorithm to compute node importance scores, selecting the most influential neighbors for each central node. MCCNIFTY~\cite{zhang2021multi} similarly uses personalized PageRank scores to extract informative regional substructures for each node, but its aim is to make node operations more computationally friendly and reduce noise from the global structure. CAF~\cite{Guo0X0W23} also samples ego subgraphs for each nodes.
    \textit{Whole graph-based} methods~\cite{dong2022edits, agarwal2021towards, kose2021fairnessaware} treat the entire graph as a whole and flip all nodes' sensitive attributes. This approach reduces computation costs, but fails to model the intrinsic causal structures of the observational data.
    
    \item{\bf Causal Graphs.} We categorize existing graph counterfactual fairness works into \textit{causal-graph-aware} methods and \textit{causal-graph-unaware} methods. Most of the existing works~\cite{zhang2021multi, dong2022edits, agarwal2021towards, kose2021fairnessaware} just perturb the sensitive attributes to get the potential outcomes. These works don't give an explicit causal graph or structural equations. Hence, we categorize them as causal-unaware methods. In contrast, GEAR~\cite{ma2022learning} constructs an explicit causal graph to model the intrinsic causal dependency of the variables in the ego-centric subgraphs for each node. Then they use Variational Graph Auto-encoder to get rid of the causal effects of the sensitive attributes. CAF~\cite{Guo0X0W23} builds a Structural Causal Model to present the causal relationships among the following five variables: sensitive attribute, ground-truth label, environment feature, content feature, and ego-graph for each node.
\end{itemize}

\subsection{Evaluation Metrics for Graph Counterfactual Fairness}

To evaluate models for graph counterfactual fairness, we consider both model utility and bias mitigation. Model utility evaluation metrics vary depending on the downstream tasks, with accuracy~\cite{ma2022learning}, AUC~\cite{agarwal2021towards}, and F1 score~\cite{agarwal2021towards} commonly used for node classification tasks. Models achieving graph counterfactual fairness should also satisfy other fairness constraints, so group fairness metrics such as statistical parity~\cite{dwork2012fairness} and equalized odds~\cite{hardt2016equality} can be adopted. However, these measures do not directly assess counterfactual fairness. Therefore, a counterfactual fairness metric~\cite{ma2022learning} has been proposed, measuring performance discrepancy between factual and counterfactual individuals with different sensitive information. We now provide details on these fairness measures.

\begin{definition}[Statistical Parity~\cite{dwork2012fairness}]
Statistical parity requires the prediction $\hat{y}$ to be independent with the sensitive attribute $s$, i.e., $\hat{y} \perp s$. 
Most works are based on a binary attribute and binary classification setting, i.e., $y \in\{0,1\} $ and $ s \in\{0,1\}$. In this setting, statistical parity is defined as:
\begin{equation}
\fontsize{9pt}{10pt}\selectfont
    P(\hat{y}=1 \mid s=0)=P(\hat{y}=1 \mid s=1).
\end{equation}
\end{definition} 
Based on the definition of statistical parity, we can measure statistical parity as:
\begin{equation}
\fontsize{9pt}{10pt}\selectfont
    \Delta_{S P}=\vert P(\hat{y}=1 \mid s=0)-P(\hat{y}=1 \mid s=1)\vert,
\end{equation}
where a smaller $\Delta_{S P}$ indicates a fairer classifier.  
% We can extend statistical parity to multi-class and multi-category case by ensuring that $\hat{y} \perp s$~\cite{locatello2019on}, which can be written as
% \begin{equation}
%     \Delta_{S P}=\frac{1}{k} {\sum}_{i=1}^{k} \max _{y_{j}}\left|P\left(\hat{y}=y_{j}\right)-P\left(\hat{y}=y_{j} \mid s=s_{i}\right)\right|.
% \end{equation}
% where $y \in\left\{y_{1}, \ldots, y_{c}\right\}$ and $s \in\left\{s_{1}, \ldots, s_{k}\right\}$ denotes the multi-class label and multi-category sensitive attribute, respectively.

% Statistical parity is a well-known fairness metric and are widely adopted in the literature\cite{dwork2012fairness}. However, \citeauthor{hardt2016equality}~\cite{hardt2016equality} pointed out that statistical parity may hurt the utility of the model. Thus, they further propose equal opportunity to alleviate the issue, which is defined as:
\begin{definition} [Equal Opportunity~\cite{hardt2016equality}]
The equal opportunity is defined for different subgroups. For different subgroups, it demands that the probability of an instance being assigned to a positive outcome should be equal, i.e., 
\begin{equation}
\fontsize{8pt}{10pt}\selectfont
    P(\hat{y}=1 \mid y=1, s=0)=P(\hat{y}=1 \mid y=1, s=1).
\end{equation}
\end{definition}
Based on the above definition, equal opportunity can be measured by:
\begin{equation}
\fontsize{8pt}{10pt}\selectfont
    \Delta_{E O}=\lvert P(\hat{y}=1 \mid y=1, s=0)-P(\hat{y}=1 \mid y=1, s=1)\rvert.
\end{equation}
% Similar to statistical parity, the extension of equal opportunity to multi-class and multi-category sensitive attributes setting can be done by changing the range of sensitive attributes and labels.

Ideally, one also wants to evaluate the quality of the counterfactual fairness directly. However, we lack the ground-truth counterfactual. According to  Definition~\ref{def:fairness}, we would expect a model to give similar predictions for factual input and counterfactual input. Thus, \citeauthor{ma2022learning}~\cite{ma2022learning} introduced a metric $\delta_{C F}$ for graph counterfactual fairness based on the prediction difference of factual and counterfactual, which is defined as:
\begin{definition}[Counterfactual Fairness Metric on Graphs~\cite{ma2022learning}] The metric is defined by the discrepancy between the factual prediction and the counterfactual prediction. With a binary sensitive attribute $S$, we set the sensitive attribute with factual value $s$ and counterfactual value $s^\prime$ and get both predictions. Then the difference would be measured as: 
\begin{equation}
\fontsize{9pt}{10pt}\selectfont
    \delta_{C F}=\left\vert P\left(\hat{y}_{S = s} \mid \mathbf{X}, \mathbf{A}\right)-P\left(\hat{y}_{S = s^{\prime }} \mid \mathbf{X}, \mathbf{A}\right)\right\vert,
\end{equation}
where $s, s^{ \prime} \in\{0,1\}^n$ are arbitrary values of sensitive attribute of all nodes. Note that we use $S=s$ and $S=s^\prime$ instead of $S \leftarrow s$ and  $S \leftarrow s$ since the desired counterfactual is varied with different counterfactual generation models. $S=s^\prime$ may not be the real counterfactual but it can be generated based on the augmentation method described in Section~\ref{fair_method}.
\end{definition}

\subsection{Datasets for Graph Counterfactual Fairness}

To evaluate the model performance in terms of fairness, we need graph datasets with both node labels and node sensitive attributes available. In this section, we will list widely-used synthetic dataset and real-world datasets.

\subsubsection{Synthetic Dataset}

\citeauthor{ma2022learning}~\cite{ma2022learning} propose a synthetic dataset for fair node classification. The sensitive attributes $s_i$, node feature $\mathbf{x}_i$, graph structure $\mathbf{A}$ and labels $y_i$ are generated by:
\begin{equation}
\fontsize{9pt}{10pt}\selectfont
\begin{aligned}
    & s_i \sim \operatorname{Bernoulli}(p), \quad \mathbf{z}_i \sim \mathcal{N}(0, \mathrm{I}), \\
    &\mathbf{x}_i=\mathcal{S}\left(\mathbf{z}_i\right)+s_i \mathbf{v}, \\
    & P\left(\mathbf{A}_{ij}=1\right)=\sigma\big(\cos (\mathbf{z}_i, \mathbf{z}_j)+a \cdot \mathbbm{1}_{[s_i=s_j]}\big)\\
    & y_i=\mathcal{B}\Big(\mathbf{w}^\top \mathbf{z}_i+w_s \frac{\sum_{j \in \mathcal{N}_i} s_j}{\lvert\mathcal{N}_i\rvert}\Big),
\end{aligned}
\end{equation}
where the sensitive attribute $s_i$ is sampled from Bernoulli distribution with $p$ as the probability of $s_i=1$. The latent representation $\mathbf{z} \in \mathbb{R}^{d}$ is sampled from a Gaussian distribution. $s_i$ and $\mathbf{z}_i$ work together to generate the node features $\mathbf{X}$, the adjacency matrix $\mathbf{A}$ and the labels $\mathbf{Y}$.
$\mathcal{S}(\cdot)$ is a sampling operation which selects $d$ dimensions out of the latent representation and $\mathbf{v} \in \mathbb{R}^d, \mathbf{v} \sim \mathcal{N}(0, \mathrm{I})$ controls influence of sensitive attributes. The sampled latent representation $\mathbf{z}_i$ and $s_i \mathbf{v}$ generate the node feature $\mathbf{x}_i$. $\mathbf{A}$ is generated with the cosine similarity of latent representations and the sensitive attributes. $\mathbbm{1}_{[\cdot, \cdot]}$ is the indicator function. It outputs 1 if $s_i=s_j$; otherwise outputs 0.
$a$ is a parameter to control the influence of sensitive attributes. Then the cosine similarity and the output of the indicator function will be fed into a Sigmoid function $\sigma(\cdot)$ to compute the probability of each edge. $\mathbf{w} \in \mathbb{R}^{d}$ is parameters sampled from a Normal distribution and $w_s$ is a scalar to control the influence of sensitive attributes on labels. The weighted representation $\mathbf{w}^T\mathbf{z}_i$ and the weighted average of each node's neighbors' sensitive attribute are used to generate binary value labels $\mathbf{Y}$ with $\mathcal{B}(\cdot)$, where $\mathcal{B}(\cdot)$ maps value into a binary value. %Specifically, they compute the mean value over all nodes, and set $y_i=1$ if it is larger than the mean value, otherwise $y_i=0$.

\subsubsection{Real-World Datasets}
There are several real-world datasets widely used to evaluate the performance of fair GNNs.
\begin{itemize}[]
\item{\textbf{Pokec-z \& Pokec-n}~\cite{dai2021say}}:
These two datasets are sampled from a social network dataset Pokec~\cite{takac2012data}. Pokec is a social network platform in Slovakia and the network has millions of users. Pokec-z and Pokec-n are sampled according to the regions of the users~\cite{dai2021say}. In these two datasets the region information is the sensitive attribute and the working field of the users is the label for node classification. Pokec-z has 67,797 users and Pokec-n has 66,569 users. The feature dimension for each node in the datasets is 59. Note that the sampling approaches of the datasets vary in different works. Therefore there may be different statistics in different papers.

\item{\textbf{UCSD34}~\cite{traud2012social}}:
This is a Facebook friendship network of the University of California San Diego. Each node denotes a user, and edges represent the friendship relations between nodes. In this dataset, ``gender'' is the sensitive attribute, and the task is to predict the major of each user. UCSD34 has 4,132 nodes. The feature dimension for each node is 7.

\item{\textbf{Credit Defaulter}~\cite{yeh2009the}}:
This dataset contains people's default payment information~\cite{ma2022learning}. Each node represents an individual and an edge between a pair of nodes represents the similarity of two individuals' spending and payment patterns. In this dataset, age is the sensitive attribute and the task is to predict the users' default ways of payment, i.e., credit card or not. This dataset has 30000 nodes and the feature dimension is 13.

\item{\textbf{German Credit}~\cite{asuncion2007uci}}:
It is a client network in a German bank. Each node is a client of the bank and an edge between a pair of nodes denotes the similarity of two clients' credit accounts.  ``gender'' is the sensitive attribute and the task is to classify the credit risk of the clients as high or low. This dataset has 1,000 nodes and the feature dimension is 27.

\item{\textbf{Recidivism (Bail)}~\cite{jordan2015the}}:
It has 18,876 nodes denoting defendants who got released on bail at the U.S. state courts during 1990-2009~\cite{ma2022learning}. Defendants are connected based on the similarity of past criminal records and demographics. In this dataset the defendants' race is the sensitive attribute. The goal is to classify defendants into bail (i.e., unlikely to commit a violent crime if leased) and no bail. This dataset has 18,876 nodes and the feature dimension is 18.
\end{itemize}

\section{Counterfactual Explanation on Graphs} \label{explanation}

Deep neural networks (DNNs) have achieved remarkable success in various domains, such as molecular biology~\cite{noe2020machine}, social networks~\cite{wu2020a}, and financial systems~\cite{carvalho2019machine}. However, the black-box nature of DNNs hinders their widespread adoption~\cite{BARREDOARRIETA202082}. Transparent and interpretable models are essential to ensure that developers understand model behavior and potential biases, and to gain user trust, especially in high-stakes scenarios~\cite{martin2019interpretable}. GNNs also face interpretability challenges, which are further exacerbated by complex and discrete graph structures. Thus, improving GNN interpretability is crucial for user trust and further maximizing GNN potential~\cite{dai2022a,zhang2022trustworthy, zhao2023faithful}. For instance, in disease diagnosis, GNNs might use patient data, genetics, and social connections to predict illness likelihood. Clear explanations can foster trust and understanding among patients and doctors. In drug discovery, GNN interpretations can aid in discovering effective molecular structures~\cite{cai2022fpgnn}.

\begin{figure*}[t]
    \centering
    \includegraphics[width=0.8\linewidth]{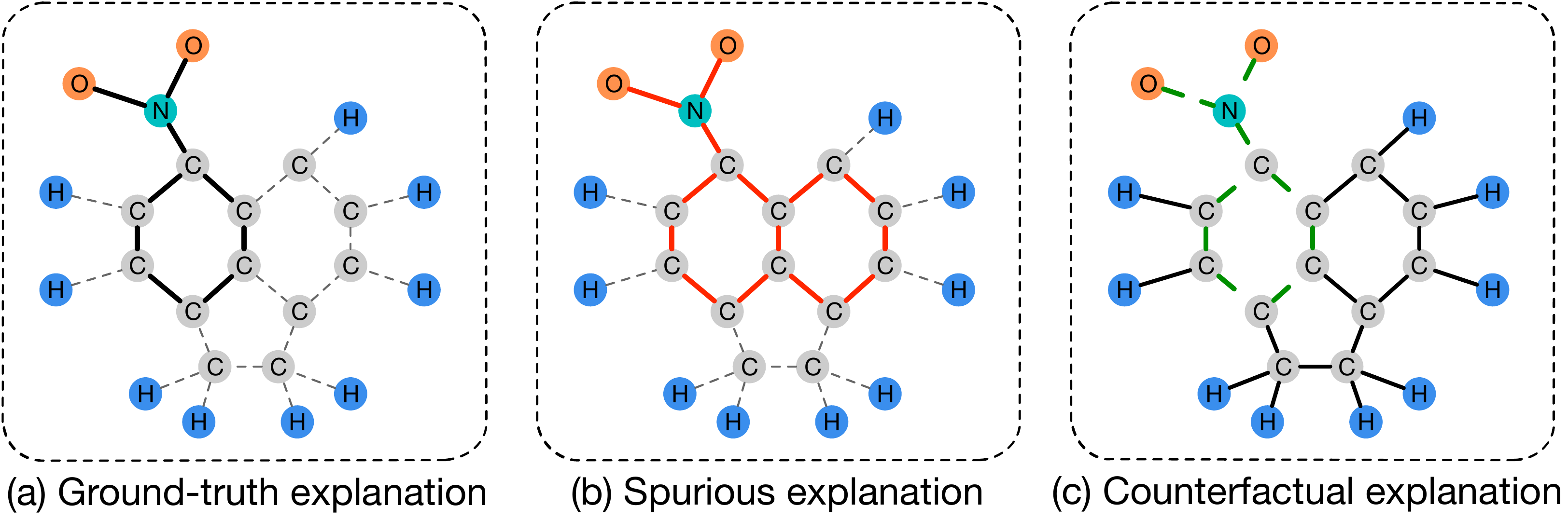}
    \caption{Illustration of counterfactual explanation on graphs for mutagenic prediction. (a) is an example of a 5-Nitroacenaphthene molecular structure (factual explanation). (b) is spurious explanation. (c) is the counterfactual explanation of molecular prediction~\cite{Prado2022Survey}.}
    \label{fig:exp_ill}
    \vskip -1em
\end{figure*}

Parallel to the development of explanation techniques on other DNN models, various interpretability methods for GNNs have been developed~\cite{yuan2022explainability}. Many existing works~\cite{ying2019gnnexplainer, luo2020parameterized, yuan2022explainability} aim to identify a subgraph that is highly correlated with the prediction result. However, due to the complex graph structure, such methods are likely to get spurious explanations that have a high correlation with the prediction result, instead of the key substructure that causes the label~\cite{zhao2023towards}.  To avoid the spurious explanation and find the causal explanation which contributes significantly to the prediction, researchers have built various models to get counterfactual explanations on graphs~\cite{bajaj2021robust, tan2022learning, lucic2022cf, sun2021preserve, numeroso2021meg,abrate2021counterfactual}. 
Instead of simply finding a subgraph that is highly correlated with the prediction result, counterfactual explanation on graphs aims to identify the necessary changes to the input graph that can alter the prediction outcome, which can help to filter out spurious explanations. Figure~\ref{fig:exp_ill} shows different explanations for a mutagenic prediction result~\cite{tan2022learning, Prado2022Survey}. In the mutagenic prediction task, the Nitrobenzene structure highlighted in black in Figure~\ref{fig:exp_ill}~(a) is the primary cause of mutagenicity, which is the ground-truth factual explanation. The edges highlighted in red in Figure~\ref{fig:exp_ill}~(b) show the explanation obtained by a factual explanation method. Explanation models tend to contain some undesired edges outside the primary cause to give more confident predictions. This is because, in the dataset, edges in red but outside the Nitrobenzene structure having high co-occurrence with the Nitrobenzene structure. Consequently, the models tend to consider these edges as highly correlated with mutagenicity, leading to potentially misleading explanations (spurious information). The green dashed edges in Figure~\ref{fig:exp_ill}~(c) serve as counterfactual explanations for the mutagenicity prediction. The intuition is that removing the edges in the Nitrobenzene structure would likely cause the disappearance of mutagenicity. Thus, counterfactual explanations can help identify the most critical edges for model prediction, aligning well with the ground-truth Nitrobenzene structure~\cite{Prado2022Survey}.
Thus, by focusing on identifying the necessary changes to the input graph that can alter the prediction outcome, counterfactual explanation methods mitigate the impact of spurious explanations and better align with the ground-truth causal factors. Hence, counterfactual explanations on graphs are promising to improve the interpretability and trustworthiness of GNNs~\cite{dai2022a} and many efforts have been taken.

Next, we first introduce background and definition of graph counterfactual explanation. Then we summarize existing works into a general framework of graph counterfactual explanation followed by a detailed review of existing methods~\cite{bajaj2021robust, tan2022learning, lucic2022cf, sun2021preserve, abrate2021counterfactual, ma2022clear, numeroso2021meg, huang2023global, liu2021multiobjective, cai2022on, ohly2022flowbased, Prado2022Ensemble}. Finally, we review widely used metrics and datasets.

\subsection{Background of Graph Counterfactual Explanations}

Various efforts have been taken for the explainability of GNNs, which can be generally categorized into instance-level post-hoc explanations~\cite{ying2019gnnexplainer, luo2020parameterized}, model-level post-hoc explanations~\cite{yuan2020xgnn} and self-explainable methods~\cite{dai2021towards}. Concretely, for each node or graph, \textit{instance-level post-hoc} explanations aim at finding an important subgraph that leads to the prediction of the target GNN~\cite{ying2019gnnexplainer}. Many strategies have been proposed to identify key subgraphs, which can be categorized into three groups, i.e., attribution methods~\cite{ying2019gnnexplainer}, decomposition methods~\cite{baldassarre2019explainability} and surrogate methods~\cite{huang2022graphlime,dai2022a}. For example, GNNExplainer~\cite{ying2019gnnexplainer} is the  first work on explaining GNN models. It identifies important subgraph substructure and node attributes that can preserve the prediction of the raw graph. The identified subgraph is treated as an explanation for the input instance. %Instead of focusing on finding a representative substructure of each instance, 
\textit{Model-level post-hoc} explanations aim to give global-level explanations which are independent of the inputs of the target model~\cite{yuan2020xgnn}. For example, XGNN~\cite{yuan2020xgnn} trains a graph generator to generate graph patterns that maximize a certain prediction of the target model for a model-level explanation. Unlike post-hoc explanations that need another model or strategy to explain a target GNN model, self-explainable approaches aim to learn a GNN model that can simultaneously give predictions and explain predictions~\cite{dai2021towards,miao2022interpretable,dai2022towards}. For example, SE-GNN~\cite{dai2021towards} identifies $K$-nearest labeled nodes for each unlabeled node to give explainable node classification. 

Although the aforementioned methods~\cite{ying2019gnnexplainer, luo2020parameterized, dai2021towards, yuan2020xgnn} can help to understand and explain GNNs, they can be easily stuck at spurious explanations~\cite{yao2021a}. Specifically, they might only identify the substructures which have a high correlation with the model predictions but cannot distinguish causal effects and spurious effects~\cite{lin2021generative, zhao2023towards}. As shown in Figure~\ref{fig:exp_ill}, counterfactual learning can help to find the most critical edges and filter out spurious edges. 
%Therefore, the generated explanations are usually not counterfactual since removing a high-correlation substructure can not necessarily change the prediction result. 
Generally, for an input instance and a trained GNN model, graph counterfactual explanation aims to explain the prediction by finding a minimal change of the input features and graph structure that would cause the target model to classify the modified input to a desired different class~\cite{kaddour2022causal}. Hence, most graph counterfactual explanations fall into the instance-level post-hoc category. Following~\cite{kaddour2022causal}, we give the formal definition of instance-level post-hoc graph counterfactual explanation as follows: 

\begin{definition}[Graph Counterfactual Explanation]
\label{def:explanation}
\end{definition} 
Given a Graph Neural Network classifier $\Phi(\cdot)$ which takes graph (or subgraph) $\mathcal{G}^\mathrm{F}=\{\mathcal{V}, \mathcal{E}, \mathbf{X}\}$ as input and gives predictions on graph classification task (or node classification task), a counterfactual explanation $\mathcal{G}^{\text{CF}}$ for the input $\mathcal{G}^\mathrm{F}$ is given by solving the following optimization problem:
\begin{equation} \label{eq:general_counterfactual_explanation}
\fontsize{9pt}{10pt}\selectfont
\begin{aligned}
    \mathcal{G}^{\text{CF}} &= \arg \min _{\mathcal{G} } \operatorname{dist}(\mathcal{G}, \mathcal{G}^{\mathrm{F}})\\
    \text { s.t. } \Phi(\mathcal{G}) &\neq \Phi(\mathcal{G}^{\mathrm{F}}) \\ \text{ and } \; \Phi(\mathcal{G}) &= y_t, \mathcal{G} \in \mathcal{P},
    \end{aligned}
\end{equation}

where $y_t$ is a target label for class $t$ so that there will be a counterfactual graph for each class except the original predicted class $\Phi\left(\mathcal{G}^{\mathrm{F}}\right)$. $\operatorname{dist}(\cdot, \cdot)$ measures the distance between $\mathcal{G}$ and $\mathcal{G}^{\text{F}}$, and $\mathcal{P}$ is an set of domain-specific constraints of the obtained counterfactual explanations~\cite{joshi2019towards}.

This definition says that the graph counterfactual explanation tries to find the counterfactual graph that gives different predictions and also emphasizes that an effective counterfactual graph should be similar to the original factual graph.
% i.e., add some edges, remove some existing edges, or change the feature values~\cite{lucic2022cf}.
Note that counterfactual explanation is different from classical explanation as graph counterfactual explanation can add/delete edges or change feature values instead of simply selecting a substructure highly correlated with the prediction~\cite{yuan2022explainability}. 
% Many existing explanation approaches in finding subgraph that preserve the prediction can be considered as a special case of the general counterfactual explanation framework in Equation~\eqref{eq:general_counterfactual_explanation} by removing the constraint $\Phi(\mathcal{G}) = y_t$ and replace $\Phi(\mathcal{G}) \ne \Phi\left(\mathcal{G}^{\mathrm{F}}\right)$ by $\Phi(\mathcal{G}) = \Phi\left(\mathcal{G}^{\mathrm{F}}\right)$.

\subsection{Methods of Graph Counterfactual Explanations}
Different from other explanation methods~\cite{ying2019gnnexplainer, luo2020parameterized} that only give substructures of high correlation score with the prediction, graph counterfactual explanation aims to get more actionable and useful explanations by understanding how the prediction can be changed in order to achieve
an alternative outcome~\cite{verma2020counterfactual}. For example, considering someone who failed to apply for a credit loan, an explanation of why the model makes the decision of rejection is acceptable. However, it is more pragmatic to tell him/her how to adjust a few changeable features or make a few transactions to get approved~\cite{prado2022gretel}, which is what counterfactual explanations are capable of. 

\subsubsection{General Framework of Graph Counterfactual Explanation}
%Taking all the aforementioned requirements into account and facing the unique challenges on graphs, 
Many graph counterfactual explanations methods have been proposed~\cite{bajaj2021robust, tan2022learning, lucic2022cf, sun2021preserve, numeroso2021meg,abrate2021counterfactual,chhablani2024game}.
In this subsection, we unify existing methods into a general framework composed of two steps, i.e., graph counterfactual candidate representation step and optimization step. An illustration of the overall framework is shown in Figure~\ref{fig:fair_frame}, where we use $\bar{y}_i$ to denote the desired label and use $\hat{y}_i$ to denote original predicted label. As shown in Figure~\ref{fig:fair_frame}, for a given GNN and the desired label $\bar{y}_i$, we need to generate appropriate masks in input space. Then the perturbed input will be fed into the GNN and give the desired $\bar{y}_i$ instead of the original $\hat{y}_i$. This goal can be achieved by several rounds of optimizations with regularization of input space and predictions (output space). Given the input instance $\mathcal{G}$, the graph counterfactual candidate representation step aims to mathematically represent the actionable space to perturb graph $\mathcal{G}$ for getting counterfactual explanations.  %we need to find a way to represent the manipulation in the input space. 
We can achieve this goal by considering the decision boundary~\cite{bajaj2021robust}, manipulating the input with masks~\cite{lucic2022cf} or any other manipulations~\cite{abrate2021counterfactual}. With the counterfactual candidate representation, the optimization step aims to design proper supervision to realize the desiderata and find reasonable graph counterfactual explanations~\cite{verma2020counterfactual}. Next, we give the details.

\textbf{Graph Counterfactual Candidate Representation}
Here we focus on post-hoc explanation setting. Given a trained GNN classifier $\Phi(\cdot)$ and a graph/subgraph $\mathcal{G}$, to provide counterfactual explanations on the prediction of $\mathcal{G}$, we need to find counterfactual graph $\mathcal{G}^{\text{CF}}$ that can explain why $\mathcal{G}$ is predicted as label $y$ and what changes we need on $\mathcal{G}$ to result in other predictions. However, in the observational study, we only have access to the observed factual graph $\mathcal{G}$. Thus, the main challenge is how to utilize the observational graph to find a reasonable counterfactual graph $\mathcal{G}^{\text{CF}}$, which has minimal change to $\mathcal{G}$ but a different prediction label from $\mathcal{G}$. A straightforward method is to try all possible masks on the input $\mathbf{A}$ and $\mathbf{X}$ and try to find masks that give the desired prediction.  However, there is a combinatory number of candidate masks for an input instance. We need to keep querying the target model to get a prediction for each candidate. Thus the computation cost is unacceptable.  Moreover, this straightforward method cannot fulfill the desiderata we mentioned above, such as validity, actionability, sparsity, data manifold closeness and causality~\cite{verma2020counterfactual}.

To address these challenges, many graph counterfactual candidate representation approaches have been proposed~\cite{bajaj2021robust, tan2022learning, lucic2022cf, sun2021preserve, numeroso2021meg,abrate2021counterfactual}. It should be noted that for other graph explanation methods~\cite{ying2019gnnexplainer, luo2020parameterized}, we only remove edges to find the critical subgraph as the explanation for the current prediction. However, in counterfactual explanation, we want to find counterfactual graphs by \textit{adding} and \textit{deleting} minimal amount of edges that can result in the counterfactual prediction, i.e., the prediction that's different from $\Phi(\mathcal{G})$ or predictions with a desired label. Therefore, we also need to consider masks on the zeros of the adjacency matrix, which makes the problem more challenging. To mathematically represent the counterfactual graph, we let the supplement of binary adjacency matrix $\mathbf{A}$ as $\overline{\mathbf{A}}=\mathbf{J}-\mathbf{I}-\mathbf{A}$, where $\mathbf{J}$ is an all-one matrix and $\mathbf{I}$ is an identity matrix. Then $\mathbf{C}=\overline{\mathbf{A}}-\mathbf{A}$ is all possible edge perturbation candidates. $\mathbf{C}$ can be decomposed as:
% \begin{equation}
%     \mathbf{C}=\mathbf{C}^{-}+\mathbf{C}^{+}, ~ \text{where}~ \mathbf{C}^{-} \in \{-1,0\}^{N \times N}, \mathbf{C}^{+} \in \{1,0\}^{N \times N}.
% \end{equation}
\begin{equation}
\fontsize{9pt}{10pt}\selectfont
\begin{aligned}
\mathbf{C} &= \mathbf{C}^{-} + \mathbf{C}^{+}, \\
\text{where} \quad  \mathbf{C}^{-} &\in \{-1,0\}^{N \times N}, \\
\mathbf{C}^{+} &\in \{1,0\}^{N \times N}.
\end{aligned}
\end{equation}

%Here $\mathbf{C}^{-} \in \{-1,0\}^{N \times N}$ 
A negative element of $\mathbf{C}^{-}$, say ${C}_{ij}^{-}=-1$, denotes that the edge $e_{ij}$ exists in $\mathbf{A}$ and can be potentially removed from $\mathbf{A}$. Similarly,  a positive element of $\mathbf{C}^{+}$, say $C_{ij}^{+}=1$, denotes that the edge $e_{ij}$ does not exist in $\mathbf{A}$ and can be potentially added~\cite{sun2021preserve}. $\mathbf{C} \in \{-1,0,1\}^{N \times N}$ and only the diagonal entries of $\mathbf{C}$ are zeros, denoting that we do not add self-loops to nodes. With $\mathbf{C}$, we can select edges from $\mathbf{C}$ with the aid of masks to form counterfactual graph candidates, which can be written as:%. Specifically, the topology mask can be modeled as:
\begin{equation}
\fontsize{9pt}{10pt}\selectfont
    \mathbf{A}^\prime = \text{Mask}(\mathbf{A})=\mathbf{A}+\mathbf{C}^{-} \odot \mathbf{M}^{-}+\mathbf{C}^{+} \odot \mathbf{M}^{+},
\end{equation}
where $\odot$ denotes element-wise product. $\mathbf{M}^{-} \in \{0,1\}^{N \times N}$ works with $\mathbf{C}^{-}$ to select the edges which need to be removed. $\mathbf{M}^{-}$ maintains zero entries of $\mathbf{C}^{-}$ and $\mathbf{M}^{-}_{ij}=1$ denotes that the edge $e_{ij}$ should be removed from $\mathbf{A}$. %$\mathbf{M}^{+} \in \{0,1\}^{N \times N}$ is similar to $\mathbf{M}^{-}$ but 
$\mathbf{M}^{+} \in \{0,1\}^{N \times N}$. Similarly, $\mathbf{M}^{+}$ stands for adding edges from the possible candidates. $\mathbf{M}^{+}$ maintains zero entries of $\mathbf{C}^{+}$ and $\mathbf{M}^{+}_{ij}=1$ denotes that the edge $e_{ij}$ should be added to $\mathbf{A}$. $\mathbf{M}=\left[\mathbf{M}^{-}, \mathbf{M}^{+}\right]$ summarizes all the adding and deleting operations. Similarly, for node features,
a mask $\mathbf{F} \in \{0,1\}^{N \times d}$ is adopted on the feature matrix $\mathbf{X} \in \mathbb{R}^{N \times d}$ to select important features as:
\begin{equation}
\fontsize{9pt}{10pt}\selectfont
    \mathbf{X}^\prime = \text{Mask}(\mathbf{X})=\mathbf{X} \odot \mathbf{F}.
\end{equation}

For graph counterfactual candidate representation, we can apply the actionability constraint on the masks, which demands additional domain knowledge. For example, we can not change one's race to change the college application result. This can be achieved by constraining perturbations in an actionable subset of all potentials, i.e., $\mathbf{F} \in \mathcal{A}_F \in \{0,1\}^{N \times d}$ and $\mathbf{M} \in  \mathcal{A}_M \in  \{0,1\}^{N \times N}$, where $\mathcal{A}_F$ and $\mathcal{A}_M$ are the actionable sets for feature mask and structure mask.

\begin{figure*}[h]
    \centering
    \includegraphics[width=0.7\linewidth]{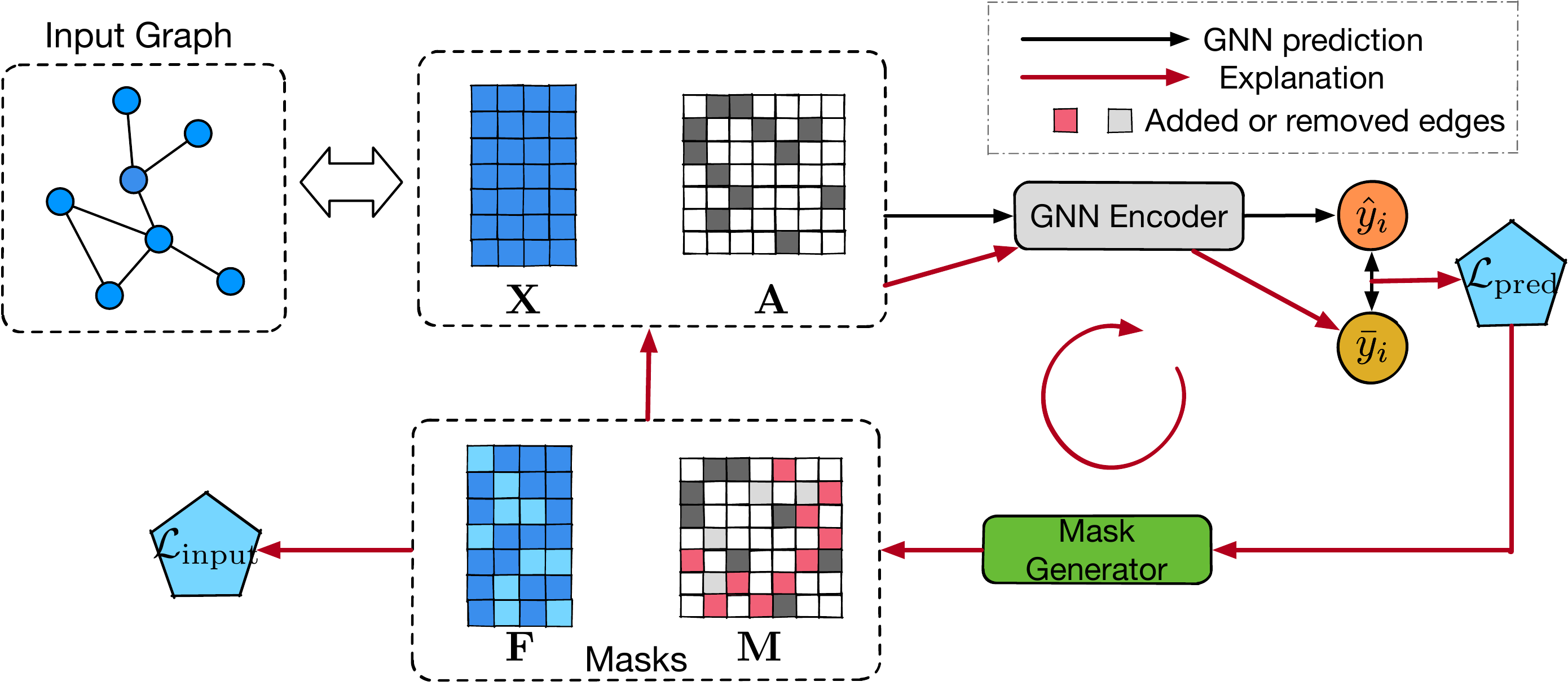}
    \caption{Overall framework of graph counterfactual explanation.}
    \label{fig:expla_frame}
    \vskip -1em
\end{figure*}

\textbf{Objective Function}
With the adjacency matrix mask $\mathbf{M}$ and the feature matrix mask $\mathbf{F}$, we have converted the graph counterfactual explanation problem into an optimization problem of finding $\mathbf{M}$ and $\mathbf{F}$ with the constraints in the input space and output space. Specifically, given the target GNN model $\Phi(\cdot)$, the instance $\mathcal{G}=\{\mathbf{A}, \mathbf{X}\}$ and the prediction $\hat{\mathbf{Y}}=\Phi(\mathbf{A}, \mathbf{X})$, we try to find the counterfactual explanation $\mathcal{G}^{CF} = \{\mathbf{A}^{'}, \mathbf{X}^{'}\}$ with $\mathbf{A}^{'} = \text{Mask}(\mathbf{A})$ and $\mathbf{F}^{'} =\text{Mask}(\mathbf{M})$. % which is represented by the masks on the full set of the potential edges and the features. 
Based on Definition~\ref{def:explanation}, 
the objective function of the optimization problem can be written as:
% \begin{equation}
% \begin{split}
%     \mathcal{L}_{\text{explain}}(\mathbf{M},\mathbf{F}) =  \mathcal{L}_{\text{pred}}(\Phi(\mathbf{A}^\prime, \mathbf{X}^\prime), \Phi(\mathbf{A}, \mathbf{X})) + \lambda \mathcal{L}_{\text{input}}((\mathbf{A}^\prime, \mathbf{X}^\prime),(\mathbf{A}, \mathbf{X})),
% \end{split}
% \end{equation}
\begin{equation}
\fontsize{9pt}{10pt}\selectfont
\begin{aligned}
\mathcal{L}_{\text{explain}}(\mathbf{M},\mathbf{F}) = & \mathcal{L}_{\text{pred}}(\Phi(\mathbf{A}', \mathbf{X}'), \Phi(\mathbf{A}, \mathbf{X})) \\
&+ \lambda \mathcal{L}_{\text{input}}((\mathbf{A}', \mathbf{X}'),(\mathbf{A}, \mathbf{X})),
\end{aligned}
\end{equation}

where $\lambda$ is the hyper-parameter to balance
the contributions of two losses. The above loss functions come from two levels, i.e., the output level and the input level. $\mathcal{L}_{\text{pred}}$ stands for the validity constraint applied on the model prediction, i.e., output level~\cite{verma2020counterfactual}. We utilize $\mathcal{L}_{\text{pred}}$ to make sure $\Phi(\mathbf{A}^\prime, \mathbf{X}^\prime) \neq \Phi(\mathbf{A}, \mathbf{X})$. An intuitive choice of $\mathcal{L}_{\text{pred}}$ is to maximize the discrepancy between the factual output and counterfactual output. $\mathcal{L}_{\text{input}}$ is used to impose constraints in the input space, which can be a set of loss terms to achieve the goal of sparsity, data manifold closeness and causality. 
To achieve sparsity, one way is to add a constraint to make factual input and counterfactual input similar, thus the changes from factual to counterfactual are minimized. For example, CF-GNNExplainer~\cite{lucic2022cf} adopts a sparsity constraint to minimize the element-wise different between factual input and counterfactual input.
For data manifold closeness, a regularization is usually added between counterfactual and factual data distribution to make sure that counterfactual distribution should not violate the factual data distribution by a large margin. As for causality, we need to know the causal relationship of the observed data and make sure the input perturbation follows the causal relationship.

\textbf{A Unified Framework}
Considering the two steps together, the unified framework of graph counterfactual explanation can be written as:
% \begin{equation}
%     \min _{\mathbf{M},\mathbf{F}} \mathcal{L}_{\text {explain }} =  \mathcal{L}_{\text {pred }} + \lambda \mathcal{L}_{\text{input}} \quad \text{ s.t. }  \quad \mathbf{M} \in \mathcal{A}_C,  \quad \mathbf{F} \in \mathcal{A}_F.
% \end{equation}
\begin{equation}
\fontsize{9pt}{10pt}\selectfont
\begin{aligned}
    \min_{\mathbf{M},\mathbf{F}} \mathcal{L}_{\text{explain}} &= \mathcal{L}_{\text{pred}} + \lambda \mathcal{L}_{\text{input}} \\
    \text{s.t.} \quad \mathbf{M} &\in \mathcal{A}_C, \\
    \mathbf{F} &\in \mathcal{A}_F.
\end{aligned}
\end{equation}

The unified framework aims to optimize from both the input level and output level. Based on different choices of graph counterfactual candidate representation approaches and optimization goals, the framework can generalize to different kinds of extensions of counterfactual explanation models. And we can utilize the constraint to achieve various specific goals of counterfactual explanation~\cite{verma2020counterfactual}. Note that the counterfactual explanation should be the necessary change to get different predictions, which corresponds to $\Delta_\mathbf{A}=\mathbf{A}^\prime-\mathbf{A}$ and $\Delta_\mathbf{X}=\mathbf{X}^\prime-\mathbf{X}$.

\subsubsection{Methodologies} \label{exp_method}
Researchers have come up with a series of carefully-designed models to get counterfactual explanations on graphs~\cite{bajaj2021robust, tan2022learning, lucic2022cf, sun2021preserve, numeroso2021meg,abrate2021counterfactual}. They usually differ in terms of $\mathcal{L}_{\text{pred}}$, $\mathcal{L}_{\text{input}}$ and optimization methods. We summarize existing works in Table~\ref{tab:explanation} and show the significant differences between these methods. Next, we will give the details of representative graph counterfactual explanation methods.

\begin{table*}[t]
\tiny
\centering
\caption{Summary of graph counterfactual explanation papers. ``Model Access'' is to distinguish between black-box models or not. ``Counterfactual'' denotes the method to obtain counterfactuals. ``Candidate'' indicates the modification to obtain counterfactuals, e.g., ``M(-, +), F, V'' stands for removing and adding edges, feature modification and node modification, respectively. ``Factual Loss'' stands for the classical explanation term, i.e., getting the part that contributes the most to the model prediction. ``CF Loss'' is to ensure the model prediction is changed. For ``Task'', we use G to denote graph classification and use N to denote the node classification.} \label{tab:explanation}
\begin{tabular}{lccccccc}
\toprule
\textbf{Method} & \textbf{Model Access} & \textbf{Counterfactual}  & \textbf{Candidate} & \textbf{Factual Loss} & \textbf{CF Loss} & \textbf{Sparse Loss} & \textbf{Task}  \\ \midrule
RCExplainer~\cite{bajaj2021robust} & \checkmark & GNN & M(-) & \checkmark & \checkmark & \checkmark & G, N  \\ \hline
CF$^2$\cite{tan2022learning} & \checkmark & Perturbation & M(-), F & \checkmark & \checkmark & \checkmark & G,N  \\ \midrule
CF-GNNExplainer\cite{lucic2022cf} & \checkmark & Perturbation & M(-)  & $\times$ & \checkmark & \checkmark & N  \\ \midrule
GNNViz~\cite{sun2021preserve} & \checkmark & Perturbation & M(-,+) & $\times$ & \checkmark & \checkmark & G  \\ \midrule
\citeauthor{abrate2021counterfactual}\cite{abrate2021counterfactual} & $\times$ & Search & M(-,+) & $\times$ & \checkmark & \checkmark & G  \\
\midrule
CLEAR\cite{ma2022clear} & $\times$ & VAE & M(-, +), F, V & $\times$ & \checkmark & \checkmark & G \\
\midrule
MEG~\cite{numeroso2021meg} & \checkmark & Perturbation & M(-,+), F, V & $\times$ & \checkmark & \checkmark & G  \\ \midrule
GCFExplainer~\cite{huang2023global} & $\times$ & Random Walk & M(-, +), V & $\times$ & \checkmark & \checkmark & G \\
\midrule
MOO~\cite{liu2021multiobjective} & $\times$ & Perturbation & M(-, +), V & \checkmark & \checkmark & \checkmark & N \\
\midrule
NSEG~\cite{cai2022on} & \checkmark & Neighbor Matching & M(-), F & \checkmark & \checkmark & \checkmark & N, G \\
\midrule
Flow-based~\cite{ohly2022flowbased} & \checkmark & ECINN~\cite{hvilshoj2021ecinn} & F & $\times$ & \checkmark & $\times$ & N \\
\midrule
Ensemble~\cite{Prado2022Ensemble} & $\times$ & Any & M(-), V & $\times$ & \checkmark & $\times$ & G \\
\midrule
\citeauthor{chhablani2024game} & $\times$ & Thresholded Banzhaf Value & M(-) & $\times$ & $\times$ & $\times$ & N \\

\bottomrule
\end{tabular}
\end{table*}

\textbf{RCExplainer~\cite{bajaj2021robust}}
Targeting at instance-level post-hoc explanation for graph classification task, this work gives a detailed analysis of the decision region of GNNs. The core idea of RCExplainer is to extract the decision region of a target GNN $\Phi(\cdot)$ in the output space of the last convolution layer through linear decision boundary analysis. The decision boundary analysis enables capturing the shared decision logic  across multiple input graphs, mitigating the potential overfitting due to the noise present in individual graph instances. As a result, the model generates more robust counterfactual explanations. With the aforementioned analysis, first they leverage a different GNN to perform graph counterfactual candidate representation by generating the edge masks as:
\begin{equation}
\fontsize{9pt}{10pt}\selectfont
    \mathbf{Z} = g_\phi(\mathbf{A}, \mathbf{X}), \quad \mathbf{M}_{i j}=f_\theta\left(\mathbf{z}_i, \mathbf{z}_j\right),
\end{equation}
where $g_{\phi}(\cdot)$ is the GNN parameterized by $\phi$ and $f_{\theta}(\cdot)$  is a MLP parameterized by $\theta$. Note that RCExplainer only considers removing edges and do not consider adding edges or changing features. With the edge masks, RCExplainer designs the objective function as:
\begin{equation}
\fontsize{8pt}{10pt}\selectfont
    \mathcal{L}_{\text{explain}} = \lambda\mathcal{L}_{\text{same}} + (1-\lambda)\mathcal{L}_{\text{opp}} + \beta\mathcal{R}_{\text{sparse}} + 
    \mu \mathcal{R}_{\text{discrete}},
\end{equation}
% where $\mathcal{R}_{\text{sparse}} = \|\mathbf{M}\|_1$ and $\mathcal{R}_{\text {discrete }}=-\frac{1}{\lvert\mathbf{M}\rvert} \sum_{i, j}\left(\mathbf{M}_{i j} \log \left(\mathbf{M}_{i j}\right)+\left(1-\mathbf{M}_{i j}\right) \log \left(1-\mathbf{M}_{i j}\right)\right)$.
where $\scriptstyle \mathcal{R}_{\text{sparse}}=\|\mathbf{M}\|_1$ and $\scriptstyle \mathcal{R}_{\text{discrete}}=\begingroup\scriptsize -\frac{1}{\lvert\mathbf{M}\rvert} \sum_{i, j}\left(\mathbf{M}_{ij} \log \left(\mathbf{M}_{ij}\right)+\left(1-\mathbf{M}_{ij}\right) \log \left(1-\mathbf{M}_{ij}\right)\right) \endgroup$.

The first two terms work for output level, which corresponds to $\mathcal{L}_{\text{pred}}$ in our unified framework. Specifically, 
$\mathcal{L}_{\text{opp}}$ works for the counterfactual property.
It encourages counterfactual graph embedding and the original graph embedding to lie on the opposite side of the decision boundary. $\mathcal{L}_{\text{same}}$ is to ensure that the removed edges themselves should have the same prediction result as the original graph. The last two terms work on the input level, which corresponds to $\mathcal{L}_{\text{input}}$. They encourage the edge mask $\mathbf{M}$ to be sparse, and discrete, i.e., be close to 0 or 1. \citeauthor{bajaj2021robust}~\cite{bajaj2021robust} also demonstrate the possibility to extend this work on node classification tasks.

\textbf{CF-GNNExplainer~\cite{lucic2022cf}}
Targeting at instance-level post-hoc explanation for node classification task, CF-GNNExplainer also follows the general framework. Let $\hat{y}_i=\argmax_c \Phi(\mathbf{A}, \mathbf{X})_i$ be the prediction given by the target GNN $\Phi(\cdot)$ for node $v_i$,
where $\Phi(\mathbf{A}, \mathbf{X})_i \in [0,1]^C$ is the predicted class probability distribution for $v_i$ and $C$ is number of classes. 
CF-GNNExplainer first defines $\hat{\mathbf{A}}=\mathbf{M} \odot \mathbf{A}$ as the counterfactual example for node $v_i$, where $\mathbf{M}$ is the mask matrix that perturbs $\mathbf{A}$. Then, it uses $\Phi(\cdot)$ to get predictions for counterfactual example as $\hat{y}^\prime_i=\argmax_c \Phi(\mathbf{M}\odot\mathbf{A}, \mathbf{X})_i$. Note that the parameters of $\Phi(\cdot)$ are fixed and we optimize the mask $\mathbf{M}$ during training.  The loss function follows the general framework as $\mathcal{L}_{\text{explain}}=\mathcal{L}_{\text{pred}} +\mathcal{L}_{\text{input}}$. Specifically,  $\mathcal{L}_{\text{pred}}$ is given as:
% \begin{equation}
%     \mathcal{L}_{\text{explain}}=\mathcal{L}_{\text{pred}} +\mathcal{L}_{\text{input}}
% \end{equation}
\begin{equation}
\fontsize{9pt}{10pt}\selectfont
    \mathcal{L}_{\text{pred}}=-\mathbb{I}_{{[}
    \hat{y}_i=\hat{y}^\prime_i
    {]}} \cdot l_{\text{NLL}}(\hat{y}_i, \Phi(\mathbf{M}\odot\mathbf{A}, \mathbf{X})_i),
\end{equation}
where $\mathbb{I}_{{[}\hat{y}_i=\hat{y}^\prime_i{]}}$ is an indicator function which outputs 1 if and only if $\hat{y}_i=\hat{y}^\prime_i$; otherwise it outputs 0. $l_{\text{NLL}}(\cdot,\cdot)$ is the negative log-likelihood loss. Note that this work does not aim at generating a counterfactual with the desired label but just a counterfactual that has a different label from the factual. For the second term $\mathcal{L}_{\text{input}}$, 
% it is computed as:
% \begin{equation}
%     \mathcal{L}_{\text{input}}=\sum_{i,j}(\mathbf{A}_{ij}-\mathbf{M}_{ij}\odot\mathbf{A}_{ij})
% \end{equation}
%where 
the authors take the element-wise distance between the adjacency matrix and its perturbed counterfactual adjacency matrix as the sparsity constraint. The mask $\mathbf{M}$ in CF-GNNExplainer is a binary perturbation matrix and CF-GNNExplainer iteratively optimizes the mask $\mathbf{M}$ with the regularization of $\mathcal{L}_{\text{input}}$ and $\mathcal{L}_{\text{pred}}$. Finally, they retrieve the optimal counterfactual explanation as the difference between the original adjacency matrix $\mathbf{A}$ and the optimal counterfactual adjacency matrix $\mathbf{M}^*\odot\mathbf{A}$.  $\Delta\mathbf{A}^*=\mathbf{A}-\mathbf{M}^*\odot\mathbf{A}$ gives the minimal change needed to change model prediction on the input instance. 

\textbf{CF$^2$~\cite{tan2022learning}}
Most of the graph explanation techniques focus on performing factual reasoning to find a subgraph that is highly correlated with the prediction. And the aforementioned graph counterfactual explanation methods only aim at finding counterfactual explanations which can change the prediction result. CF$^2$ combines these two approaches and aims to get necessary and sufficient explanations. Here \textit{sufficiency} means for factual reasoning the information induced by explanation should be enough to produce the same prediction as the original graph. \textit{Necessity} means that removing the minimal part will result in different prediction results. In this work, same as the graph counterfactual candidate representation step,  CF$^2$ first defines masks on the adjacency matrix and feature matrix as $\mathbf{M}$ and $\mathbf{F}$. It tackles graph classification task, where each graph $\mathcal{G}_k=\{\mathbf{A}_k, \mathbf{X}_k\}$ is associated with a ground-truth class label $y_k \in \mathcal{C}$ with $\mathcal{C}=\{1,2, \cdots, r\}$ as the set of classes. Let $\hat{y}_k={\arg \max }_{c \in \mathcal{C}} P_{\Phi}\left(c \mid \mathbf{A}_k, X_k\right)$ be the predicted label made by model $\Phi$ for $\mathcal{G}_k$. 
We set $S_f(\mathbf{M}, \mathbf{F})=P_{\Phi}\left(\hat{y}_k \mid \mathbf{A}_k \odot \mathbf{M}_k, \mathbf{X}_k \odot \mathbf{F}_k\right)$ and $S_c(\mathbf{M}, \mathbf{F})=-P_{\Phi}\left(\hat{y}_k \mid \mathbf{A}_k-\mathbf{A}_k \odot \mathbf{M}_k, \mathbf{X}_k-\mathbf{X}_k \odot \mathbf{F}_k\right)$
Then the condition for factual reasoning and counterfactual reasoning can be given as~\cite{tan2022learning}:
% \begin{equation}
% \fontsize{8pt}{10pt}\selectfont
% \begin{split}
% \mathcal{L}_f &=\operatorname{ReLU}\left(\gamma+P_{\Phi}\left(\hat{y}_{k, s} \mid \mathbf{A}_k \odot \mathbf{M}_k, \mathbf{X}_k \odot \mathbf{F}_k\right)\right.\left.-S_f\left(\mathbf{M}_k, \mathbf{F}_k\right)\right), \\
% \mathcal{L}_c &= \operatorname{ReLU}\left(\gamma-S_c\left(\mathbf{M}_k, \mathbf{F}_k\right)-P_{\Phi}\left(\hat{y}_{k, s} \mid \mathbf{A}_k-\mathbf{A}_k \odot \mathbf{M}_k, \mathbf{X}_k-\mathbf{X}_k \odot \mathbf{F}_k\right)\right),
% \end{split}
% \end{equation}

\begin{equation}
\fontsize{7pt}{8pt}\selectfont
\begin{split}
\mathcal{L}_f = \operatorname{ReLU}\bigg(&\gamma + P_{\Phi}(\hat{y}_{k, s} \mid \mathbf{A}_k \odot \mathbf{M}_k, \mathbf{X}_k \odot \mathbf{F}_k)\\
&- S_f(\mathbf{M}_k, \mathbf{F}_k)\bigg), \\
\mathcal{L}_c = \operatorname{ReLU}\bigg(&\gamma - S_c(\mathbf{M}_k, \mathbf{F}_k)\\
&- P_{\Phi}(\hat{y}_{k, s} \mid \mathbf{A}_k - \mathbf{A}_k \odot \mathbf{M}_k, \mathbf{X}_k - \mathbf{X}_k \odot \mathbf{F}_k)\bigg),
\end{split}
\end{equation}

where $\gamma$ is the parameter to control the margin.  $S_f(\cdot)$ and $S_c(\cdot)$ controls the factual reasoning and counterfactual reasoning, respectively.  $\hat{y}_{k, s}$ denotes  denotes the label other than $\hat{y}_k$ that has the largest probability
score predicted by the GNN model. $\mathcal{L}_f$ aims to ensure that the predicted label $\hat{y}_k$ has the highest probability among all labels when only using the factual explanation sub-graph $\mathbf{A}_{k}\odot \mathbf{M}_k$ and $\mathbf{X}_k \odot \mathbf{F}_k$, which guarantees that the prediction remains the same for factual explanation. On the other hand, $\mathcal{L}_c$ is to ensure that when the factual explanation subgraph is removed, the probability of $\hat{y}_k$ should be lower than the probability of $\hat{y}_{k,s}$ by $\gamma$, leading to a change in the prediction. Therefore, $\mathcal{L}_f$ and $\mathcal{L}_c$ serve to ensure that the sub-graph explanation is both necessary and sufficient for the model's prediction. With these two conditions, the loss function of CF$^2$ is:
\begin{equation}
\fontsize{9pt}{10pt}\selectfont
    \mathcal{L}_{\text{explain}}=\mathcal{L}_{\text{input}}+\lambda(\alpha \mathcal{L}_f+(1-\alpha) \mathcal{L}_c),
\end{equation}
% \begin{equation}
% \mathcal{L}_{\text{input}}=\left\|\mathbf{M}_k\right\|_1+\left\|\mathbf{F}_k\right\|_1
% \end{equation}
where $\mathcal{L}_{\text{input}}=\left\|\mathbf{M}_k\right\|_1+\left\|\mathbf{F}_k\right\|_1$ %$\left\|\mathbf{M}_k\right\|_1+\left\|\mathbf{F}_k\right\|_1$ 
works on the input level to make the masks sparse. %$\mathcal{L}_f$ and $\mathcal{L}_c$ work on the output level. %where $\mathcal{L}_f$ corresponds to the condition for factual reasoning and $\mathcal{L}_c$ corresponds to the condition for counterfactual reasoning. 

\textbf{GNNViz~\cite{sun2021preserve}}
%\citeauthor{sun2021preserve}~\cite{sun2021preserve} proposes GNNViz to get counterfactual explanations for graphs. 
GNNViz only considers the perturbation of adding and removing edges, i.e., the perturbed adjacency matrix is $ \mathbf{A}^\prime = \mathbf{A}+\mathbf{C}^{-} \odot \mathbf{M}^{-}+\mathbf{C}^{+} \odot \mathbf{M}^{+}$, where $\mathbf{M}=\left[\mathbf{M}^{-}, \mathbf{M}^{+}\right]$ is the mask matrix. To generate a counterfactual explanation, GNNViz encourages the target model to have higher confidence in other labels than the origin predicted label, i.e.,
% \begin{equation}
%     \min_{\mathbf{M}} \mathcal{L}_{\text{pred}} +  \lambda \mathcal{L}_{\text{input}} = \min_{\mathbf{M}} -\min \big\{\max _{t \notin \Omega} p\left(t | \mathbf{A}^\prime, \mathbf{X} \right)-   \max _{c \in \Omega} p\left(c | \mathbf{A}^\prime, \mathbf{X} \right), \kappa  \big\} + \lambda (- \lambda_1 \| \mathbf{M}^- \|_1 +  \lambda_2 \| \mathbf{M}^+ \|_1) %\\
%     %\mathcal{L}_{\text{input}}(\mathbf{M}) = - \lambda_1 \| \mathbf{M}^- \|_1 +  \lambda_2 \| \mathbf{M}^+ \|_1,
% \end{equation}
\begin{equation}
%\begin{aligned}
\fontsize{8pt}{10pt}\selectfont
    \mathcal{L}_{\text{pred}}(\mathbf{M}) = -\min \big\{\max _{t \notin \Omega} p\left(t \vert \mathbf{A}^\prime, \mathbf{X} \right)-   \max _{c \in \Omega} p\left(c \vert \mathbf{A}^\prime, \mathbf{X} \right), \kappa  \big\},
    %\mathcal{L}_{\text{input}}(\mathbf{M}) &= - \lambda_1 \| \mathbf{M}^- \|_1 +  \lambda_2 \| \mathbf{M}^+ \|_1,
%\end{aligned}
\end{equation}
where $\Omega$ is the set of original predicted labels, and $\kappa$ is the confidence gap used to encourage the model to give more confidence on changed labels. The final loss function is
\begin{equation}
\fontsize{9pt}{10pt}\selectfont
    \min_{\mathbf{M}} \mathcal{L}_{\text{pred}}(\mathbf{M}) - \lambda_1 \| \mathbf{M}^- \|_1 +  \lambda_2 \| \mathbf{M}^+ \|_1,
\end{equation}
where $\lambda_1$ and $\lambda_2$ are scalars to control the contribution of two regularization terms. %Note that the first regularization term %of %$\mathcal{L}_{\text {input }}(\mathbf{M})$ 
%is 
$- \lambda_1 \| \mathbf{M}^- \|_1 +  \lambda_2 \| \mathbf{M}^+ \|_1$ is used to get a sparse counterfactual adjacency matrix, i.e., by minimizing $- \lambda_1 \| \mathbf{M}^- \|_1$, we encourage removing more edges and by minimizing $\lambda_2 \| \mathbf{M}^+ \|_1$, we prefer adding fewer edges.
%$-\lambda_1\left\|\mathbf{M}^{-}\right\|_1$. 
%The reason is that the model want to get a sparse counterfactual adjacency matrix $\mathbf{A}^\prime$ instead of a sparse mask $\mathbf{M}$.

\textbf{MEG~\cite{numeroso2021meg}}
Given a target model for molecular property prediction (graph classification), MEG aims to get counterfactual explanation for the prediction of each molecule, which can help to find molecules that have the desired properties.  Our unified framework gives a feasible answer to counterfactual explanation questions, i.e., first find a way to represent the potential manipulation space for generating counterfactuals, then design objectives on output level and input level to make the model give different prediction results with reasonable regularization. The framework is flexible and can be easily adopted in different frameworks, e.g., the reinforcement learning framework in MEG. In the reinforcement learning framework of MEG, it defines the action space as:
\begin{equation}
\fontsize{9pt}{10pt}\selectfont
    \mathcal{A}_t=\mathcal{A}_a \cup \mathcal{A}_b^{+} \cup \mathcal{A}_b^{-} \cup\{\perp\},
\end{equation}
where $\mathcal{A}_a$ stands for adding nodes (atoms), $\mathcal{A}_b^{+}$ or $\mathcal{A}_b^{-}$ stands for adding or removing edges (bonds), $\perp$ means no action is taken. The reward function of MEG is given as
\begin{equation}
\fontsize{8.5pt}{10pt}\selectfont
    \begin{aligned}
        \max \mathcal{R}_{\text{explain}} &= \alpha \mathcal{R}_{\text{pred}} + (1-\alpha) \mathcal{R}_{\text{input}}\\  &= -\alpha p(c \vert \mathcal{G}^{\text{CF}}) +(1-\alpha) \mathcal{K}(\mathcal{G}, \mathcal{G}^{\text{CF}}),%, \\
    \end{aligned}
\end{equation}
% \begin{equation}
%     \begin{aligned}
%         \mathcal{L}_{\text{pred}} &= \alpha p(c | \mathcal{G}^{\text{CF}}), \\
%         \mathcal{L}_{\text{input}} &= -(1-\alpha) \mathcal{K}\left[\mathcal{G}, \mathcal{G}^{\text{CF}}\right],
%     \end{aligned}
% \end{equation}
 where $c$ is the predicted label of the original molecule graph $\mathcal{G}$. $\mathcal{G}^{\text{CF}}$ is the generated counterfactual graph. $\mathcal{K}(\cdot,\cdot)$ is defined as the similarity score of the representations of $\mathcal{G}$ and $\mathcal{G}^{\text{CF}}$. $\alpha$ controls the contribution of $\mathcal{L}_{\text{pred}}$ and $\mathcal{L}_{\text{input}}$. The loss function encourages the model to give less confidence to the original predicted label and have minimal changes to generate counterfactual graph $\mathcal{G}^{\text{CF}}$.

\textbf{CLEAR~\cite{ma2022clear}}
CLEAR is another method that aligns well with our unified framework, with a focus on graph classification. Unlike other methods that use masks to perturb the graph structure, CLEAR uses a variational auto-encoder (VAE) to generate the counterfactual graph. The advantage of using VAE is that, by properly injecting auxiliary information, VAE can preserve the latent causal structure of the graph. The counterfactual graph is generated as:
\begin{equation}
\fontsize{9pt}{10pt}\selectfont
    \mathcal{G}^{\text{CF}} = \{\mathbf{A}^{\text{CF}}, \mathbf{X}^{\text{CF}} \} = \operatorname{VAE}(\mathcal{G}, y_t, s),
\end{equation}
where $y_t$ is the desired label for $\mathcal{G}^{\text{CF}}$ and $s$ is an auxiliary observed attribute, such as an index for each sample or class label. CLEAR formulates the loss function as:
\begin{equation}
\fontsize{9pt}{10pt}\selectfont
    \begin{aligned}
        \mathcal{L}_{\text{pred}} &= \alpha \cdot l(\Phi(\mathcal{G}^{\text{CF}}), y_t) + \mathcal{L}_{\text{VI}},  \\
        \mathcal{L}_{\text{input}} &= \|\mathbf{A}^{\text{CF}} - \mathbf{A}\|_F^2 + \|\mathbf{X}^{\text{CF}} - \mathbf{X}\|_F^2,\\
    \end{aligned}
\end{equation}

where $\Phi(\cdot)$ is the target GNN, $y_t$ is the desired label for $\mathcal{G}^{\text{CF}}$, $\mathcal{L}_{\text{VI}}$ is defined as the variational distribution regularization for VAE and $l(\cdot, \cdot)$ is the cross-entropy prediction loss. The regularization term $\|\mathbf{A}^{\text{CF}} - \mathbf{A}\|_F^2 + \|\mathbf{X}^{\text{CF}} - \mathbf{X}\|_F^2$ is applied to both the adjacency matrix and feature matrix to ensure that the model makes minimal changes to the graph structure, i.e., adding or removing fewer edges and modifying fewer features. This regularization penalizes large changes in the graph, promoting explanations that are as close as possible to the original graph while still achieving the desired counterfactual effect, i.e., making $\Phi(\mathcal{G}^{\text{CF}})$ to predict the label $y_t$.

\textbf{Brain~\cite{abrate2021counterfactual}}
In the context of brain network classification, \citeauthor{abrate2021counterfactual}~\cite{abrate2021counterfactual} proposes a search-based counterfactual explanation approach to produce local post-hoc explanations of any black-box graph classifier. %The counterfactual explanation problem can be formulated as changing the links of a given graph to create a new graph that is very similar to the original but classified differently by the same black-box classifier. 
The approach is composed of two steps. In the first step, starting with the original graph, the authors iteratively query the target model by adding or removing several edges to the original graph. The process continues until the prediction changes. This step allows the authors to identify an appropriate number of edge changes required to alter the prediction result. In the second step, the authors iteratively add or remove edges to create a counterfactual graph that has a smaller edit distance from the original input. The authors' two-step framework aligns with our unified framework where the search space is used to obtain graph counterfactual candidates, and the first and second steps play similar roles to the objective function, i.e., altering the prediction and making smaller changes to the inputs.

\textbf{Other Works} 
In this part, we will explore a variety of studies that, while demonstrating strong alignment with the unified framework outlined in the previous section, exhibit distinct emphases on their respective goals or methodologies. For each work, we do not introduce too many technical details but aim to highlight the diversity of these works. These works are also shown in Table~\ref{tab:explanation}.

\begin{itemize}[]

\item{\bf Model-level Graph Counterfactual Explanation.}
The aforementioned works focus on instance-level explanations, i.e., explaining the predictions made for specific instances. In contrast, GCFExplainer~\cite{huang2023global} aims to provide model-level counterfactual explanations for GNNs. In graph classification tasks, given the target GNN classifier $\Phi(\cdot)$, the global counterfactual explanation aims to find a global rule $r_c(\cdot)$ for each class $c$ such that for each graph $\mathcal{G}_i$ with $\Phi(\mathcal{G}_i) \neq c$, $r_c(\cdot)$ can modify $\mathcal{G}_i$ to counterfactual graph $r(\mathcal{G}_i)$ with prediction $\Phi(r(\mathcal{G}_i)) = c$. In GCFExplainer, \citeauthor{huang2023global}~\cite{huang2023global} simplify this problem of finding $r(\cdot)$ into finding a set of \textit{representative} counterfactual graphs for each class, where counterfactual graph means a modified from the original graph but with a different label with the original graph. Based on the language of our framework, in the graph counterfactual candidate representation step, they structure the search space for counterfactuals. The loss function incorporates regularization at both the input level and output level, where the regularization encourages the model prediction in the desired class at the output level and regularization ensures the manipulation should be sparse at the input level. Those \textit{representative} counterfactual graphs are treated as model-level counterfactual explanations.

\item{\bf Factual and Counterfactual Explanation.}
For graph classification task, \textit{factual explanation} aims to identify the key subgraph of $\mathcal{G}_i$ whose information is sufficient to preserve the label information of $\mathcal{G}_i$. \textit{Counterfactual explanation} aims to find minimal manipulation of $\mathcal{G}_i$ that is \textit{necessary} to result in different prediction. CF$^2$~\cite{tan2022learning} is an early effort to combine these two directions to get a sufficient and necessary explanation for target GNN models as discussed in Section~\ref{exp_method}.  MOO~\cite{liu2021multiobjective} is another work combining these two directions and trying to find optimal explanations that are well-balanced in sufficiency and necessity. MOO also aligns well with our framework, where the perturbation is employed to get graph counterfactual candidates and both regularizations of sufficiency and necessity are applied as the objective. However, \citeauthor{cai2022on}~\cite{cai2022on}argue that the trade-off of necessity and sufficiency in these approaches is heuristically determined. Hence, the obtained explanation might not be an optimal balance for both desiderata.   NSEG~\cite{cai2022on} is proposed to alleviate the issue. NSEG adopts a technique named Probability of Necessity and Sufficiency (PNS) to quantify the necessity and sufficiency and aims to maximize the lower bound of PNS to find the optimal explanation. Specifically, NSEG hinges on a nearest neighbor matching strategy to obtain graph counterfactual candidates and then uses counterfactuals to maximize the lower bound of NSEG.

\item{\bf Generative Model-Based Explanation.}
Another intriguing development in the realm of graph counterfactual explanation is the generative model-based counterfactual explanations~\cite{ohly2022flowbased, ma2022clear}. These works also align with our unified framework, where they use generative models to obtain graph counterfactual candidates and adopt regularizations on output and input level to supervise the learning process. For example, CLEAR~\cite{ma2022clear} uses VAE to generate the counterfactual graphs.  \citeauthor{ohly2022flowbased}~\cite{ohly2022flowbased} designs a flow-based model to generate counterfactuals. Specifically, \citeauthor{ohly2022flowbased} adopts ECINN~\cite{hvilshoj2021ecinn}, a normalizing-flow-based generative model that can generate counterfactual candidates, to alter the node features. A classifier is adopted to make sure that the counterfactual examples get changed prediction results compared with origin predictions.

\item{\bf Ensemble-Based Explanation.}
Instead of sticking to one single model, \citeauthor{Prado2022Ensemble}~\cite{Prado2022Ensemble} introduces an ensemble-based approach, which leverages the combined power of multiple models to improve the overall interpretability and performance. In their model, multiple graph counterfactual explanation models are used to generate a set of counterfactual explanations on graphs. Then the shared substructure of these explanations is treated as the final counterfactual explanation. 

\item{\bf Semivalue-Based Explanation.}
Different from previous methods that often necessitate the training of additional graphs, ~\citeauthor{chhablani2024game} introduce a non-learning-based approach grounded in semivalues~\cite{dubey1979mathematical}. Specifically, they design a function to measure the importance of edges from the utility functions in cooperative game theory based on Banzhaf values.
\end{itemize}

\subsection{Taxonomy of Graph Counterfactual Explanation Methods}
In this subsection, we categorize existing graph counterfactual explanation methods from four perspectives: counterfactual graph, objective, performed task and model access. A summary of the categorization can be found in Table~\ref{tab:explanation}. Next, we will introduce each perspective in detail.

\begin{itemize}[]
\item{\bf Counterfactual Graph.}
From the perspective of counterfactual graph design, existing methods can be classified into \textit{perturbation-based methods}~\cite{tan2022learning, lucic2022cf, sun2021preserve, numeroso2021meg, liu2021multiobjective}, \textit{search-based methods}~\cite{abrate2021counterfactual}, \textit{neural network-based methods}~\cite{bajaj2021robust, ma2022clear, ohly2022flowbased}, and \textit{other methods}~\cite{huang2023global, cai2022on,chhablani2024game}.
\textit{Perturbation-based methods} develop perturbation masks for the adjacency matrix and feature matrix, which convert counterfactual graph generation into mask optimization problem~\cite{tan2022learning, lucic2022cf, sun2021preserve, numeroso2021meg, liu2021multiobjective}. These methods are easy to implement and can be applied to explain a variety of target models.
\textit{Search-based method} employs a straightforward approach, i.e., iteratively removing or adding edges until the desired requirements are met~\cite{abrate2021counterfactual}. It needs to query the target model but does not require any additional information.  
\textit{Neural network-based methods} utilize various neural networks, such as GNN~\cite{kipf2017semisupervised}, VAE~\cite{kingma2013autoencoding}, and ECINN~\cite{hvilshoj2021ecinn}, to learn the counterfactual graph structure. Neural networks possess a powerful representation capacity to capture the latent relationships among variables, providing semantically meaningful counterfactuals.
\textit{Other methods} obtain the counterfactual graphs with different specific concerns. For example, random walk is used in~\cite{huang2023global} to help identify important representative nodes for graphs in a class, where the shared nodes are thought to be representative for the class. Then these nodes are used to form counterfactual graphs, which work as the explanation of the mentioned class. \cite{Prado2022Ensemble} adopts ensemble learning to combine explanations from multiple graph counterfactual models. \cite{chhablani2024game} utilizes a function to measure the importance of edges to decide what edges to delete.

\item{\bf Objective.} 
The Graph Counterfactual Candidate Representation Step formulates the search space for counterfactual graphs and generates a selection of counterfactual graph candidates. Following this, objectives are applied to assess the model's performance concerning various factors. Within the context of graph counterfactual explanation, objectives can take on different forms, including reward functions, loss functions, and termination conditions for the search process. Regardless of the form, we can categorize the objectives' desiderata into several types, such as \textit{factual regularization}, \textit{counterfactual regularization}, and \textit{sparsity regularization}. 
Most of the graph counterfactual explanation methods incorporate \textit{counterfactual regularization}, which requires changes in model predictions~\cite{bajaj2021robust, tan2022learning, lucic2022cf, sun2021preserve, abrate2021counterfactual, ma2022clear, numeroso2021meg, huang2023global, liu2021multiobjective, cai2022on, ohly2022flowbased, Prado2022Ensemble}. The regularization usually involves decreasing the confidence score of the original prediction or ensuring the counterfactual prediction lies on the opposite side of the decision boundary. With this regularization, the model's prediction is modified, allowing for the examination of the desired change. Some methods also employ \textit{factual regularization}, which assists in identifying sufficient information to complement counterfactual regularization~\cite{bajaj2021robust, tan2022learning, liu2021multiobjective, cai2022on}. These methods aim to identify important substructures from two perspectives and they are capable of more precisely capturing the important substructure. By using the substructure, we can maintain the original prediction while removing them resulting in significant changes in the model's prediction. They usually employ a regularization to increase the prediction score when solely using the explanation substructure. For the \textit{sparsity regularization}, it is employed to promote concise and interpretable explanations by limiting the number of edges or feature changes in the counterfactual explanation~\cite{bajaj2021robust, tan2022learning, lucic2022cf, sun2021preserve, abrate2021counterfactual, ma2022clear, numeroso2021meg, huang2023global, liu2021multiobjective, cai2022on}. The regularization minimizes the $l_1$ norm of the masks of counterfactual explanation or the difference between the counterfactual graph and the original graph.

\item{\bf Tasks.}
Regarding the tasks performed, existing works can be generally categorized into graph-level and node-level tasks. In \textit{graph-level tasks}, each graph, considered as an instance, is independent and identically distributed (i.i.d.). The majority of existing methods aim to find reasonable counterfactual explanations for a single prediction result in graph classification tasks~\cite{bajaj2021robust, tan2022learning, sun2021preserve, abrate2021counterfactual, ma2022clear, numeroso2021meg, huang2023global, cai2022on, Prado2022Ensemble}. Among these works, the objective of GCFExplainer~\cite{huang2023global} is slightly different, as it seeks to find several representative counterfactual graphs for model-level explanations. Some studies focus on the \textit{node classification setting}~\cite{bajaj2021robust, tan2022learning, lucic2022cf, liu2021multiobjective, cai2022on, ohly2022flowbased,chhablani2024game}, which entails finding reasonable modifications that can result in changes to the node classification outcomes. These methods aim to identify and understand the impact of specific modifications on the node classification results within the graph structure.

\item{\bf Model Access.}
In the context of explainers, it is essential to discuss the degree of access to the target model, particularly whether it is treated as \textit{black box or not}. \textit{Black-box} setting means internal workings of the target model are hidden or unknown, making it challenging to explain predictions of the target model. Explanation techniques for black-box setting~\cite{abrate2021counterfactual, ma2022clear, huang2023global, liu2021multiobjective, Prado2022Ensemble,chhablani2024game} often involve querying the black box model and collecting its feedback to guide their neural networks~\cite{ma2022clear}, masks learning~\cite{liu2021multiobjective}, or search processes~\cite{abrate2021counterfactual}. For \textit{white-box} setting which allows access to the target model's parameters, the problem is considerably easier. These methods~\cite{bajaj2021robust, tan2022learning, lucic2022cf, sun2021preserve, numeroso2021meg, cai2022on, ohly2022flowbased} can leverage the target model's prediction probabilities~\cite{sun2021preserve}, embeddings~\cite{bajaj2021robust}, or even model parameters~\cite{lucic2022cf} to supervise the explainer. With these additional information, explainers with access to the target model can more effectively generate insightful and accurate explanations of the target model's decision-making process.

\end{itemize}

\subsection{Evaluation Metrics for Graph Counterfactual Explanations}

Many quantifiable proxies have been proposed to measure the performance of counterfactual explanations, such as accuracy, fidelity, sparsity, stability and graph edit distance. GRETEL also discusses the evaluation of graph counterfactual explanation~\cite{prado2022gretel}.

\textbf{Accuracy, F1 and AUC}
When ground-truth rationales are available for graphs, a direct evaluation method can be employed by comparing the identified explanatory components to the ground-truth explanations~\cite{luo2020parameterized}. This allows for a comprehensive assessment of the explanation's quality and relevance. Metrics such as accuracy score, F1 score, and ROC-AUC score can be computed based on this comparison to the ground-truth explanation. Higher scores signify more accurate explanations.
% \suhang{I deleted what you added. An explanation is a subgraph that's composed of many edges.  Then what is a correct explanation? Do we calculate from the edge level or what? Could you please think about the details instead of simply writing down a general equation of number of correct divided by total}
% Higher scores signify more accurate explanations. A potential accuracy metric can be formulated as follows:
% \begin{equation}
%     \text{Accuracy} = \frac{\text{Number of Correct Explanations}}{\text{Total Number of Explanations}}
% \end{equation} 
% By using this metric, we can quantitatively assess the performance of our explanation model.

\textbf{Fidelity}
When ground-truth explanations are unavailable, fidelity-based metrics can be employed to measure the explanation performance. The underlying idea is that explanatory substructures should have a more significant impact on predictions, thus removing them or adding them will have a noticeable change in the GNN prediction score. Fidelity+~\cite{yuan2022explainability} is computed by removing all input elements first, then gradually adding edges with the highest explanation scores. Intuitively, a faster increase in GNN’s prediction indicates stronger fidelity of obtained explanations. On the contrary, Fidelity-~\cite{yuan2022explainability} is computed by sequentially removing edges following assigned importance weight. A faster performance drop represents stronger fidelity of removed explanations.

% Specifically, Fidelity+ is defined as the difference between the original predictions and the predictions obtained after removing the explanatory substructure. On the other hand, Fidelity- is defined as the difference between the original predictions and the predictions generated by retaining the explanatory substructure while removing other information~\cite{yuan2020explainability}. We give the fidelity accuracy~\cite{yuan2022explainability} of graph classification explanation as: 
% \begin{equation}
%     \text{Fidelity+}=\frac{1}{N} \sum_i^N(\mathbbm{1}(\hat{y}_i=y_i)-\mathbbm{1}(\hat{y}^\prime=y_i)), \text{Fidelity-}=\frac{1}{N} \sum_i^N(\mathbbm{1}(\hat{y}_i=y_i)-\mathbbm{1}(\hat{y}^{\prime \prime}=y_i)),
% \end{equation}
% where $y_i$ represents the label of graph $i$, $N$ denotes the number of graphs, $\hat{y}_i$ is the original prediction score, $\hat{y}^\prime$ is the prediction score after removing the edges and features in the explanation, and $\hat{y}^{\prime \prime}$ is the prediction score when only retaining the edges and features in the explanation.

\textbf{Sparsity} 
Sparsity, in the context of graph counterfactual explanation, refers to the minimal number of changes or interventions required to generate a counterfactual instance that would lead to a different outcome~\cite{bajaj2021robust}. A sparse explanation is desirable because it provides a understandable and concise rationale for the change in the outcome. In other words, the fewer changes needed, the easier it is for humans to comprehend and trust the explanation. Graph counterfactual explanations involve generating an alternative graph with a limited number of modifications to original graph, such as adding or removing edges or nodes. These changes should be sufficient to alter the outcome predicted by the model. Sparsity aims to minimize these modifications while maintaining the desired counterfactual effect.
% The sparsity~\cite{prado2022gretel} of features can be defined as $\text{Sparsity} = 1 - \frac{\| \mathbf{X}^{\text{F}}-\mathbf{X}^{\text{CF}} \|_2}{|\mathbf{X}|}$, 
% \begin{equation}
%     \text{Sparsity} = 1 - \frac{\| \mathbf{X}^{\text{F}}-\mathbf{X}^{\text{CF}} \|_2}{|\mathbf{X}|}
% \end{equation}
% where $|\mathbf{X}|$ is the number of attributes and $\|\cdot \|_2$ is the L2 norm. 

\textbf{Stability}
Ideally, good explanations should capture the intrinsic causal connections between input graphs and their labels. This criterion necessitates that the identified explanations remain stable under small perturbations~\cite{tan2022learning}. By making perturbations in the input graph, such as adding new nodes or removing edges, the stability scores can be calculated by comparing changes in explanations before and after the perturbation. 

\textbf{Graph Edit Distance (GED)}
The Graph Edit Distance (GED)~\cite{prado2022gretel, Prado2022Survey} is employed to measure the structural distance between the original graph $\mathcal{G}^{\text{F}}$ and the counterfactual graph $\mathcal{G}^{\text{CF}}$. Let $p_i \in \mathcal{P}(\mathcal{G}^{\text{F}},\mathcal{G}^{\text{CF}})$ represent a path consisting of actions to be performed on $\mathcal{G}^{\text{F}}$ in order to generate $\mathcal{G}^{\text{CF}}$. Each path contains specific actions $r_j \in p_i$ with an associated cost $\gamma(r_j)$. The GED can be defined as $\text{GED} = \min_{p_i \in \mathcal{P}\left(\mathcal{G}^{\text{F}}, \mathcal{G}^{\text{CF}}\right)} \sum{r_j \in p_i} \gamma\left(r_j\right)$. 
% \begin{equation}
% \text{GED} = \min_{p_i \in \mathcal{P}\left(\mathcal{G}^{\text{F}}, \mathcal{G}^{\text{CF}}\right)} \sum{r_j \in p_i} \gamma\left(r_j\right).
% \end{equation}
Ideally, a smaller GED is preferred, indicating that $\mathcal{G}^{\text{F}}$ is close to $\mathcal{G}^{\text{CF}}$. This implies that only minimal changes are required in the input space to alter the prediction result.

\subsection{Datasets for Counterfactual Explanations on Graphs}
% revise this (especially the assumption)
To evaluate the performance of graph counterfactual explanation models on both node classification tasks and graph classification tasks, a set of synthetic datasets and real-world datasets are used by researchers.
% Note that the ground-truth explanations are worked as counterfactual explanation. \suhang{no, groundtruth explanation is not counterfactual explanation. }
In this section, we will first introduce the design of synthetic datasets. Then we will discuss the widely-used real-world datasets. 
% \zhimeng{ground-truth factual explanations, why unique for this}

% \begin{table}[t]
%     \scriptsize
%     \centering
%     \begin{tabular}{lllllll}
%         \hline Tasks & Dataset & Graphs & Avg.Nodes & Avg.Edges & Features & Classes \\
%         \hline & BA-Shapes & 1 & 700 & 4,110 & 10 & 4  \\
%         Bode Classification & BA-Community & 1 & 1,400 & 8,920 & 1 & 8  \\
%         & Tree-Cycles & 1 & 871 & 1,950 & 10 & 2  \\
%         & Tree-Grid & 1 & 1,231 & 3,410 & 10 & 2 \\
%         & Syn-Cora & 1 & 1,895 & 2,769 & 1,433 & 7  \\
%         \hline  
%         Graph Classification& BA-2motifs & 1,000 & 25 & $51.4$ & 10 & 2  \\
%         & Infection & 10 & 1000 & 3996 & 2 & 6  \\
%         & Graph-SST2 & 70,042 & $10.199$ & $9.20$ & 768 & 2  \\
%         & Graph-SST5 & 11,855 & $19.849$ & $18.849$ & 768 & 5  \\
%         & Graph-Twitter & 6,940 & $21.103$ & $21.10$ & 768 & 3  \\
%         & MUTAG & 188 & $19.79$ & $17.93$ & 14 & 2 \\
%         \hline
%         \end{tabular}
%     \caption{Datasets for counterfactual explanation.}
%     \label{tab:explanation_dataset}
% \end{table}

% \suhang{This paper also discussed some datasets: When Comparing to Ground Truth is Wrong: On Evaluating GNN Explanation Methods. Can you take a look and include the relevant datasets and evalaution metrics?}

\subsubsection{Synthetic Datasets}
Different from images and texts, it is typically difficult to get the grountruth explanations for real-world graphs due to the complex graph structure. To address this problem, we can develop synthetic graphs with groundtruth explanations. Next, we introduce some widely used synthetic datasets for explainable node classification and graph classification.
%There \keypoint{Node Classification}
% Node classification: BA-shapes, BA-community, Tree-cycles, Tree-grid
\begin{itemize}[]
    \item{\textbf{BA-Shapes}~\cite{ying2019gnnexplainer}}: It is a single graph dataset which is consisting of a base Barabasi-Albert (BA) graph (300 nodes) and 80 ``house''-structured motifs (each for 5 nodes). The motifs are randomly attached to the base BA graph. Nodes in the base BA graph are labeled as 0 while the nodes in motifs are labeled as 1, 2 and 3 based on their location in the ``house''. The ground truth explanation for each node is the corresponding motif.
    \item{\textbf{Tree-Cycles}~\cite{ying2019gnnexplainer}}: It is a single graph with an 8-layer balanced binary tree as the base graph and 80 cycle motifs (each for 6 nodes) are randomly attached to the base graph. Similar to the BA-Shapes dataset, nodes in the base graph are labeled as 0 and those in the motifs are labeled as 1. The ground truth explanation for each node is the corresponding cycle.
    \item{\textbf{Tree-Grid}~\cite{ying2019gnnexplainer}}: This dataset is the same as Tree-Cycles except that 3-by-3 grid motifs are randomly attached to the base tree graph to replace the cycle motifs.
    \item{\textbf{BA-2motifs}~\cite{luo2020parameterized}}: It is a graph classification dataset which contains 800 graphs. ``house'' motifs are attached to half of the graphs while ``cycle'' motifs are attached to the other half of the graphs. And the motif works as the ground-truth explanation.
\end{itemize}

\subsubsection{Real-world Datasets}
In addition to the synthetic graphs, there are some real-world graphs which are used for evaluating graph counterfactual explanation performance. 
\begin{itemize}[]
\item{\textbf{Mutagenicity}~\cite{debnath1991structureactivity}}: This is a graph classification dataset where each graph corresponds to a molecule.  Nodes represent atoms and edges represent chemical bonds. Molecules are labeled with consideration of their chemical properties, and discriminative chemical groups are identified using prior domain knowledge. 

\item{\textbf{NCI1}~\cite{wale2008comparison}}: This is a graph classification dataset from the cheminformatics domain, where each graph is a chemical compound similar to Mutagenicity dataset. It is relative to anti-cancer screens where the chemicals are assessed as positive or negative for cell lung cancer.

\item{\textbf{TOX21}~\cite{KKMMN2016}}: It is a dataset comprised of 12,060 training samples and 647 test samples that represent chemical compounds. There are 801 ``dense features'' that represent chemical descriptors, such as molecular weight, solubility or surface area, and 272,776 ``sparse features'' that represent chemical substructures. For each sample there are 12 binary labels that represent the outcome (active/inactive) of 12 different toxicological experiments.

\item{\textbf{ESOL}~\cite{KKMMN2016}}: This is a regressive task dataset on the water solubility of chemical compounds, which includes 1,129 compounds. It can be used to estimate solubility directly from chemical structure.

\item{\textbf{Autism Spectrum Disorder (ASD)}~\cite{craddock2013the}}: It is a publicly-available dataset for human-brain. \citeauthor{abrate2021counterfactual}~\cite{abrate2021counterfactual} only focus on the portion of a dataset containing children below 9 years of age, which are 49 individuals in the condition group, labeled as Autism Spectrum Disorder (ASD) and 52 individuals in the control group, labeled as Typically Developed (TD).

\item{\textbf{Attention Deficit Hyperactivity Disorder (ADHD)}~\cite{brown2012the}}: It is also a publicly-available dataset for human-brain. This dataset contains 190 individuals in the condition group, labeled as ADHD and 330 individuals in the control group, labeled as TD.

\item{\textbf{CiteSeer}~\cite{kipf2017semisupervised}}: This is a citation network where nodes denote papers and citations between two papers are represented as edges. The nodes have six classes which stand for different categories.
\end{itemize}

\section{Counterfactual Link Prediction and Recommendation} \label{link}

In the above sections, we mainly focus on counterfactual fairness and counterfactual explanation of GNNs for node classification and graph classification. Link prediction~\cite{kumar2020link}, which aims to predict the missing links in a graph, is another important graph mining task. It has wide adoptions on various applications such as recommender systems~\cite{ying2018graph}, knowledge graphs~\cite{zhang2020relational} and social networks~\cite{traud2012social}. %Go beyond existing link prediction techniques, 
Recently, counterfactual link prediction~\cite{zhao2022learning} was investigated, which aims to explore the root causes of the formation of links, filtering out the spurious factors. Recommender systems, as a special case of link prediction task, can also benefit from removing spurious information and relying on causal information. Hence, counterfactual learning is attracting increasing attention in link prediction task~\cite{zhao2022learning} and recommender systems~\cite{wang2021clicks, chen2022grease, mu2022alleviating, liu2021improving,SongW0L0Y23}. In this section, we will give a comprehensive review of existing works on counterfactual link prediction and counterfactual recommendation with graph learning.

\begin{table}
% \begin{wraptable}[12em]{R}{6cm}
    \centering
    \caption{Summary of counterfactual link prediction and counterfactual recommendation.}
    \fontsize{8pt}{9pt}\selectfont
    \begin{tabular}{lc}
    \toprule
    \textbf{Methods}  & \textbf{Category} \\
    \midrule
    CFLP~\cite{zhao2022learning} & Link Prediction \\
    \midrule
    KGCF~\cite{chang2023knowledge} & Knowledge Graph Completion\\
    \midrule
    CR~\cite{wang2021clicks} & Recommendation  \\
    \midrule
    UKGC~\cite{liu2021improving} & Recommendation \\
    \midrule 
    CGKR~\cite{mu2022alleviating} & Recommendation \\
    \midrule
    COCO-SBRS~\cite{SongW0L0Y23} & Recommendation \\
    \midrule
    GREASE~\cite{chen2022grease} & Recommendation and Explanation \\
    \bottomrule
    \end{tabular}
    \label{tab:link}
\end{table}

\subsection{Counterfactual Link Prediction}

In this subsection, we will first introduce background knowledge about link prediction and then focus on the very recent counterfactual link prediction work.

\textbf{Background}
Link prediction is widely used in applications like social recommendation~\cite{barbieri2014who}, knowledge graph completion~\cite{sun2019rotate}, and chemical interaction prediction~\cite{abbas2021application}. Approaches to link prediction can be categorized into heuristic-based~\cite{katz1953a, nassar2019pairwise} and representation learning-based methods~\cite{kipf2016variational, pan2022neural}.
Heuristic-based approaches rely on statistical properties to estimate link likelihoods~\cite{lv2011link}. For instance, the common-neighbor index scores node pairs based on shared neighbors~\cite{lv2011link}. However, these approaches only utilize topology information, neglect node features, and may make overly strong assumptions, limiting their broader applicability~\cite{pan2022neural}. Representation learning-based approaches predict link probabilities by learning node representations and computing dot products between them. For example, VGAE~\cite{kipf2016variational} learns meaningful low-dimensional representations that can reconstruct the original graph structure and feature information. Then, they use the representations of a pair of nodes $(v_i,v_j)$ to predict the probability of link existence. WalkPool~\cite{pan2022neural} adopts a learnable random-walk-based mechanism to encode both node features and graph topology in the node representations.

\textbf{Motivation of Counterfactual Link Prediction}
Counterfactual link prediction (CLP) lies in the intersection of counterfactual learning and link prediction, which aims to remove spurious factors and investigate the root causes of the link formation~\cite{zhao2022learning}. For example, assume that Alice and Adam are friends because they live in the same community and they share some same interests. If we want to dig out how likely two people with the same interests will become friends, the community information can be a spurious factor and we need to mitigate the influence of communities. CLP aims to know the real causes from observational data instead of performing impossible experiments to have Alice and Adam grow up in different communities and see if they will become friends. CFLP~\cite{zhao2022learning} is the only work on counterfactual link prediction until now, which aims to know the existence of links without the influence of community information, i.e., ``would the link exist if the graph structure became different from observation?''.

\textbf{Methodologies}
%Since there's only one work in CLP, we will briefly introduce the approach in this work, i.e., CFLP~\cite{zhao2022learning} and we leave the discussion about potential directions of counterfactual link prediction in the future directions section
%Based on the aforementioned motivations, CFLP~\cite{zhao2022learning} removes the spurious community factor to make more accurate link predictions. 
Given the observed link distribution, the core idea of CFLP is to model the counterfactual link distribution where the community factor has been changed, which corresponds to having Alice and Adam grow up in different communities. %Then \citeauthor{zhao2022learning} design their model to learn node representations that are informative to both distributions. By comparing the observed link distribution and the counterfactual link distribution, the influence of the community factor can be eliminated. 
Specifically, CFLP is composed of two steps: (i) obtain counterfactual links; (ii) train informative node representations to predict both factual link distribution and counterfactual link distribution. 

%The first step is to model the counterfactual link distribution with different community information. 
Since the community information is unobserved, CFLP first utilizes Louvain~\cite{blondel2008fast} to perform community detection based on the graph structure. With the detected communities, it constructs the treatment matrix $\mathbf{T}$ with $\mathbf{T}_{ij}=1$ if $e_{ij}$ is an intra-community edge and $\mathbf{T}_{ij}=0$ otherwise. %They call matrix $\mathbf{T}$ as treatment matrix since they think same-community and different-community are different treatments for node pairs. 
One challenge is to induce the counterfactual links, i.e., how to know the relationship if Alice and Adam grow up in different communities. CFLP proposes to find another two people who are similar to Alice and Adam but grow up in different communities and treat their link status as the counterfactual link for Alice and Adam. Specifically, for each pair of nodes $(v_i, v_j)$, CFLP finds a pair of nodes $(v_a, v_b)$ that is similar with $(v_i, v_j)$ but has different treatment $\mathbf{T}_{ab}=1-\mathbf{T}_{ij}$ as: %obtains the counterfactual node pair $(v_a, v_b)$ by examining the existence of another pair of nodes which is similar to $(v_i, v_j)$, which can be written as:
\begin{equation}
\fontsize{8pt}{10pt}\selectfont
(v_a, v_b)=\underset{(v_a, v_b) \in \mathcal{V}}{\arg \min }\left\{h\left(\left(v_i, v_j\right),\left(v_a, v_b\right)\right) \mid \mathbf{T}_{ab}=1-\mathbf{T}_{ij} \right\},
\end{equation}
where $\mathcal{V}$ is the set of nodes in the graph, $h(\cdot, \cdot)$ is a metric for measuring the distance between two pairs of node pairs. However, the computation cost of computing the distance with respect to both graph topology and feature information for all pairs of node pairs is intractable. To resolve this issue, CFLP uses the distance of node embeddings to simplify the problem:
% \begin{equation} \label{counterfactual_select}
% \fontsize{8pt}{10pt}\selectfont
% \begin{aligned}
% \left(v_{a}, v_{b}\right)=\underset{v_{a}, v_{b} \in \mathcal{V}}{\arg \min }\left\{d\left(\tilde{\mathbf{x}}_{i}, \tilde{\mathbf{x}}_{a}\right)+d\left(\tilde{\mathbf{x}}_{j}, \tilde{\mathbf{x}}_{b}\right) \mid\right.
% \left.\mathbf{T}_{ab}=1-\mathbf{T}_{ij}, d\left(\tilde{\mathbf{x}}_{i}, \tilde{\mathbf{x}}_{a}\right)+d\left(\tilde{\mathbf{x}}_{j}, \tilde{\mathbf{x}}_{b}\right)<2 \gamma\right\},
% \end{aligned}
% \end{equation}

\begin{equation} \label{counterfactual_select}
\fontsize{8.5pt}{10pt}\selectfont
\begin{aligned}
\left(v_{a}, v_{b}\right) = \underset{v_{a}, v_{b} \in \mathcal{V}}{\arg \min} \Bigg\{ &d\left(\tilde{\mathbf{x}}_{i}, \tilde{\mathbf{x}}_{a}\right) + d\left(\tilde{\mathbf{x}}_{j}, \tilde{\mathbf{x}}_{b}\right) \mid \\
&\mathbf{T}_{ab} = 1 - \mathbf{T}_{ij}, \\
&d\left(\tilde{\mathbf{x}}_{i}, \tilde{\mathbf{x}}_{a}\right) + d\left(\tilde{\mathbf{x}}_{j}, \tilde{\mathbf{x}}_{b}\right) < 2 \gamma \Bigg\}.
\end{aligned}
\end{equation}

where $\tilde{\mathbf{x}_i}$ is the node embedding of node $v_i$ obtained from the encoder MVGRL~\cite{hassani2020contrastive}, $d(\cdot, \cdot)$ is specified as the Euclidean distance in the embedding space, and $\gamma$ is a hyper-parameter that defines the minimal distance that two nodes are considered similar. Here, we utilize $\tilde{\mathbf{x}_i}$ instead of $\mathbf{z}_i$ to denote the node embedding since there is a different encoder to learn the embedding $\mathbf{z}_i$ for downstream link prediction task in CFLP. With this simplification, the model only needs $O(N^2)$ comparisons to find the counterfactual node pairs. Then, the counterfactual treatment matrix $\mathbf{T}^{CF}$ and the counterfactual adjacency matrix $\mathbf{A}^{CF}$ can be obtained as: 
% \begin{equation}
% \mathbf{T}_{ij}^{CF}, \mathbf{A}_{ij}^{C F}=\left\{\begin{array}{cc}1-\mathbf{T}_{ij}, \mathbf{A}_{ab} & , \text { if } \exists\left(v_a, v_b\right) \in \mathcal{V} \times \mathcal{V} \text { satisfies Eq.~\eqref{counterfactual_select} } \\ \mathbf{T}_{ij}, \mathbf{A}_{ij} & , \text { otherwise. \suhang{this is not counterfactual??}}\end{array}\right.
% \end{equation}
\begin{equation}
\fontsize{9pt}{10pt}\selectfont
\begin{aligned}
\mathbf{T}_{ij}^{CF}, \mathbf{A}_{ij}^{C F}&=1-\mathbf{T}_{ij}, \mathbf{A}_{ab},\\ \text { if } \exists\left(v_a, v_b\right) &\in \mathcal{V} \times \mathcal{V} \text { satisfies Eq.~\eqref{counterfactual_select} }.
\end{aligned}
\end{equation}

The second step is to train a representation learning model based on both factual links and counterfactual links. Following the commonly-used link prediction scheme, CFLP utilizes a GCN encoder to get node representations $\mathbf{Z} $ as $\mathbf{Z} = GCN(\mathbf{A}, \mathbf{X})$. Then a MLP decoder is adopted to reconstruct both factual links and counterfactual links as:
\begin{equation}
\fontsize{9pt}{10pt}\selectfont
\begin{aligned}
    \widehat{\mathbf{A}}_{ij}&=\operatorname{MLP}\left([\mathbf{z}_i \odot \mathbf{z}_j, \mathbf{T}_{ij}]\right), \\ \widehat{\mathbf{A}}_{ij}^{C F}&=\operatorname{MLP}\big([\mathbf{z}_i \odot \mathbf{z}_j, \mathbf{T}_{ij}^{C F}]\big),
\end{aligned}
\end{equation}
% \begin{equation}
% \begin{split}
%      \widehat{\mathbf{A}} &= g(\mathbf{Z}, \mathbf{T}), \text { s.t. } \widehat{\mathbf{A}}_{ij}=\operatorname{MLP}\left(\left[\mathbf{z}_i \odot \mathbf{z}_j, \mathbf{T}_{ij}\right]\right), \\
%      \widehat{\mathbf{A}}^{C F} &= g\left(\mathbf{Z}, \mathbf{T}^{C F}\right), \text { s.t. } \widehat{\mathbf{A}}_{ij}^{C F}=\operatorname{MLP}\left(\left[\mathbf{z}_i \odot \mathbf{z}_j, \mathbf{T}_{ij}^{C F}\right]\right)
% \end{split}
% \end{equation}
%where $f(\cdot)$ is a GCN encoder, $g(\cdot)$ is a MLP decoder, $\widehat{\mathbf{A}}_{ij}$ is the predicted adjacency matrix, $\widehat{\mathbf{A}}^{C F}$ is the predicted counterfactual adjacency matrix, 
where $\mathbf{z}_i$ is the node representation of $v_i$ and $[\cdot,\cdot]$ means concatenation for vectors. %\odot$ stands for element-wise product. 
CFLP adopts the binary cross entropy loss to supervise the training process, which can be written as:
\begin{equation}
\fontsize{9pt}{10pt}\selectfont
\begin{aligned}
\mathcal{L}_F&=  \frac{1}{N^2} {\sum}_{i=1}^N {\sum}_{j=1}^N \ell(\mathbf{A}_{ij},\widehat{\mathbf{A}}_{ij}), \\
\mathcal{L}_{C F}&=  \frac{1}{N^2} {\sum}_{i=1}^N {\sum}_{j=1}^N \ell(\mathbf{A}_{ij}^{CF},\widehat{\mathbf{A}}_{ij}^{CF}),
\end{aligned}
\end{equation}

where $\ell(\mathbf{A}_{ij},\widehat{\mathbf{A}}_{ij})$ denotes the cross entropy loss between $\mathbf{A}_{ij}$ and $,\widehat{\mathbf{A}}_{ij}$,   $\mathcal{L}_F$ is the loss for the factual link prediction loss and $\mathcal{L}_{CF}$ is the counterfactual link prediction loss. Let $\hat{P}_f^F$ and $\hat{P}_f^{C F}$ be the node pair representations learned by graph encoder $f$ from factual distribution and counterfactual distribution, i.e., $\left[\mathbf{z}_i \odot \mathbf{z}_j, \mathbf{T}_{ij}\right]$ and $\big[\mathbf{z}_i \odot \mathbf{z}_j, \mathbf{T}_{ij}^{C F}\big]$. Since the test data contains only factual links at the inference stage, the model may suffer from the risk of covariant shift. Thus, CFLP adopts the discrepancy distance to regularize the representation learning, i.e., minimize the discrepancy of learned factual representation distributions $\hat{P}_f^F$ and learned counterfactual representation distribution $\hat{P}_f^{C F}$:
\begin{equation}
\begin{aligned}
\fontsize{9pt}{10pt}\selectfont
    \mathcal{L}_{\text {disc }}&=\operatorname{disc}(\hat{P}_f^F, 
    \hat{P}_f^{C F}),\\ \text { where } \operatorname{disc}(P, Q)&=\|P-Q\|_F.
\end{aligned}
\end{equation}

%Here $\|\cdot \|_F$ denotes the Frobenius norm. 
Combining the above link prediction loss and the discrepancy-minimization regularization, the overall training loss of CFLP is $\mathcal{L}=\mathcal{L}_{F}+\alpha \cdot \mathcal{L}_{C F}+\beta \cdot \mathcal{L}_{d i s c}$, 
% \begin{equation}
% \mathcal{L}=\mathcal{L}_{F}+\alpha \cdot \mathcal{L}_{C F}+\beta \cdot \mathcal{L}_{d i s c}
% \end{equation}
where $\alpha$ and $\beta$ are hyperparameters to control the weights of counterfactual link prediction loss and discrepancy-minimization loss.

\subsection{Counterfactual Knowledge Graph Completion}
Counterfactual knowledge graph completion is a natural extension of counterfactual link prediction. %Link prediction tasks, in most cases, are considered in homogeneous graphs, where the type of nodes and edges are always the same~\cite{kumar2020link, zhang2018link}. 
Knowledge graphs are heterogeneous graphs which have multiple types of nodes and edges~\cite{ji2021a}. For example, a knowledge graph may have entities such as persons, organizations and locations, and relations between them may include ``is located in'', ``work for'', or ``has headquarters in''. Knowledge graph completion (KGC) is an essential task on knowledge graphs, which aims to predict missing relationships in a knowledge graph based on the existing relationships between entities. %Considering the complexity of knowledge graphs, it is infeasible to directly apply most link prediction models to knowledge graph completion tasks (KGC). 
Among existing KGC approaches, embedding-based KGC methods with GNNs~\cite{chen2020knowledge, wang2021dskreg}, which learn representations for entities in the embedding space with specifically designed GNNs, have become a popular choice. For example, R-GCN~\cite{schlichtkrull2018modeling} uses relation-specific transformations and graph convolution to learn entity representations. %In this subsection, since there's only one work, i.e., KGCF, about counterfactual knowledge graph completion, we will briefly introduce the background, motivation and method of KGCF~\cite{} in this subsection.

%\keypoint{KGCF.}
Since there are many different types of relations and the relation types are usually imbalanced on knowledge graphs, GNN-based KGC methods are found to have difficulty in properly modeling the imbalanced distribution of relations~\cite{chang2023knowledge}.  %KGCF~\cite{}, the core idea is to augment existing relations by exploiting the counterfactual space. 
Motivated by the causal relationship among the entities on a
knowledge graph, KGFC~\cite{chang2023knowledge} aims to alleviate this issue by answering a counterfactual question: ``would the relation still exist if the neighborhood of entities
became different from observation?''. Specifically, it generates counterfactual relations by treating the representations of entity pair given relation as context, structural information of relation-aware neighborhood as treatment, and validity of the composed triplet as the outcome. 
With the augmented knowledge graph, the downstream GNN-based model can learn more balanced relations and facilitate pair representation learning for KGC. KGCF extends the idea of CFLP~\cite{zhao2022learning}. It also uses a clustering method to obtain inter-community edges and intra-community edges as different treatments. The main difference between KGCF and CFLP is that the links can have different types. Thus, KGCF makes several changes on CFLP: (a) In Eq.~\eqref{counterfactual_select}, to find counterfactual links, CFLP does not consider the relation type; while KGCF requires that the counterfactual link should have the same relation type as the corresponding factual link; (b) To better model the relations in the representation learning step, KGCF utilizes NBFNet~\cite{zhu2021neural} as the encoder, which is an advanced graph learning framework for formulating different types of relations in knowledge graphs.

\subsection{Counterfactual Recommendation}

Recommendation is a widely-used application of link prediction and a crucial service in information systems~\cite{wu2022graph} and it can also be viewed as link prediction task.  In this section, we first introduce the background of graph-based recommender systems~\cite{lim2020stpudgat, li2020hierarchical, liu2020deoscillated} and the motivation for counterfactual recommendation. We then review recent advances in counterfactual recommendation.

\subsubsection{Background and Motivation}
% In the information age,  information overload makes it difficult for users to filter information and find useful information. 
Recommender systems can rescue users from information overload and find the information they need in a timely manner. 
% Thus, recommender systems have been widely used in many services such as e-commerce~\cite{chen2020copurchaser}, social networks~\cite{fan2019graph}, and digital streaming~\cite{qiu2020gag}.
During the past decades, the mainstream of recommender systems has evolved from neighborhood-based methods~\cite{linden2003amazon, sarwar2001itembased} to representation learning-based methods~\cite{wang2015collaborative, sedhain2015autorec}. Neighborhood-based methods record historical-interacted items and recommend similar items to users~\cite{sarwar2001itembased}. 
% However, the success of representation-based methods in Netflix Prize competition has drawn a surge of interest in representation-based models~\cite{bennett2007the}. Since then, 
In recent years, a variety of recommendation models have been proposed to learn user and item representations, ranging from matrix factorization techniques~\cite{koren2021advances, koren2009matrix,xiao2019bayesian} to deep learning methods~\cite{sedhain2015autorec, zhang2019deep}. Generally, user-item interactions can be modeled as a bipartite graph, with users and items as nodes and edges representing complex relationships between them. Leveraging these graphs, graph-based representation learning approaches~\cite{wu2019dual, wang2019knowledgeaware, fan2019graph, li2020hierarchical} have demonstrated promising performance in recommendation tasks. For instance, HiGNN~\cite{li2020hierarchical} constructs new coarsened user-item graphs by clustering similar users or items and using the clustered centers as new nodes, enabling the model to capture hierarchical relationships among users and items more effectively. 

Despite the success of existing graph-based recommender systems, they might make predictions with misleading spurious factors~\cite{gao2022causal,xiao2022representation,xiao2022towards}. For example, some of the recommender systems prefer to recommend items with attractive exposure features. But when the actual content cannot match the exposure features the recommendation will disappoint users. Here the exposure features work as the misleading spurious factor and we want to mitigate the influence of the spurious factor. Another example is the movie recommender system. Alice watched a lot of English movies, hence the recommender systems tend to think Alice enjoys English movies and continues to recommend English movies to her. However, the real reason for Alice watching those movies might be that she likes movies from the genre of Sci-Fi and many Sci-Fi movies are in English~\cite{mu2022alleviating}. The language factor is the misleading spurious factor in this case. Many of the existing recommender systems are fooled by these misleading factors and fail to provide high-quality recommendations. Inspired by the causal theory, counterfactual recommendation attempts to build determined causal relationships among involved variables and remove the misleading spurious factors.

\subsubsection{Methodologies}
There are several attempts on counterfactual recommendation to remove spurious factors in various scenarios~\cite{SongW0L0Y23,wang2021clicks, mu2022alleviating, liu2021improving, chen2022grease}. Next, we will briefly introduce them.

\begin{figure}
    \centering
    \includegraphics[width=0.7\linewidth]{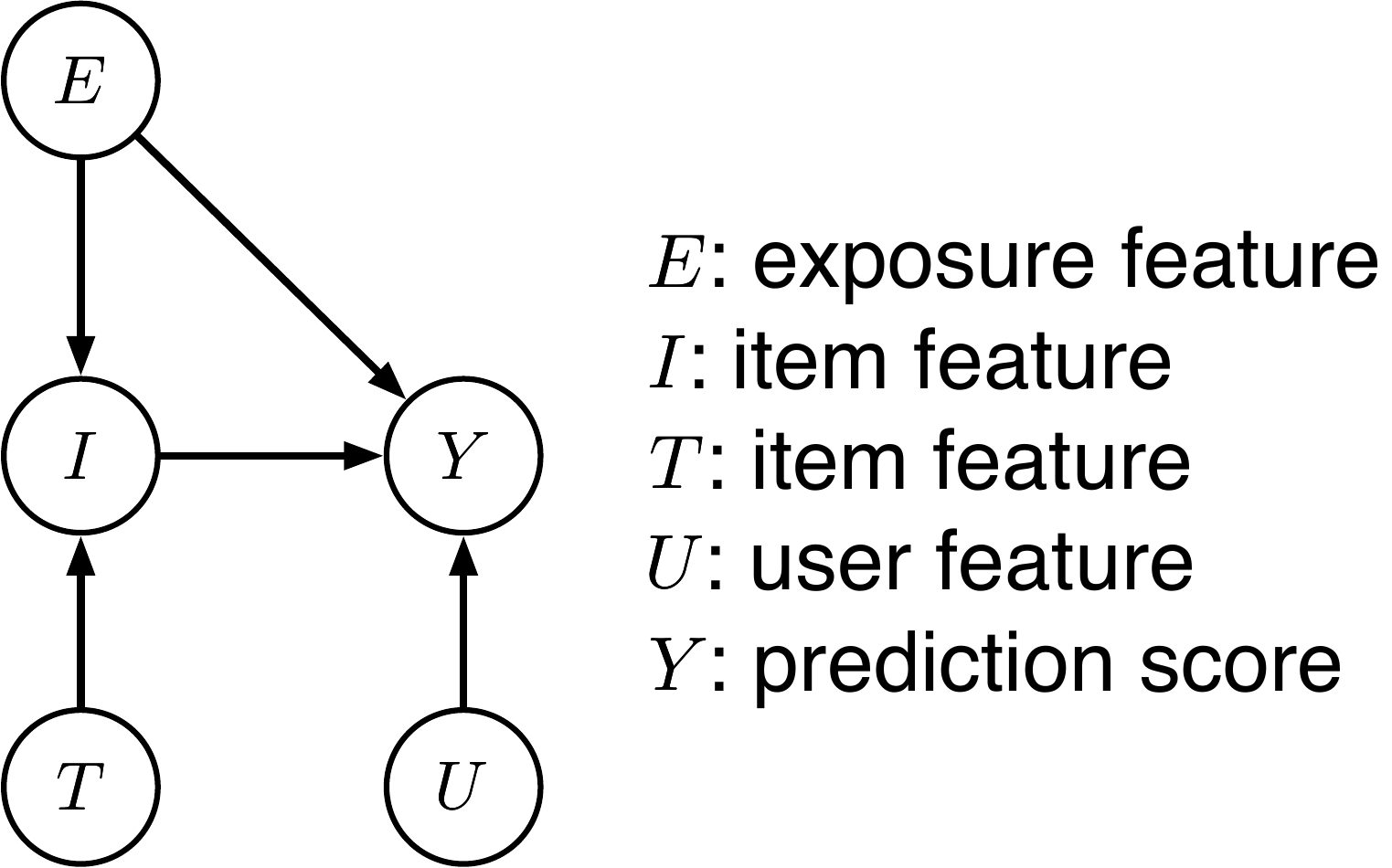}
    \caption{Causal Graph for Clickbait Issue~\cite{wang2021clicks}.}
    \label{fig:recommend}
\end{figure}
\textbf{Counterfactual Recommendation for Mitigating Clickbait Issue} 
Many recommender systems try to optimize the Click-Through-Rate (CTR)~\cite{zhou2018deep, zhang2021deep}, i.e., maximizing the likelihood that a user clicks the recommended items. However, the clickbait issue widely exists in this scenario, which utilizes attractive exposure features $E$ to attract clicks but shows disappointing content $T$ after clicking~\cite{wang2021clicks}. Given the misleading exposure features, a click from user $u$ to item $i$ cannot reflect the true preference of $u$ to $i$. Hence, recommender systems trained on such datasets will tend to recommend items with attractive exposure features with potentially disappointing content, which significantly hurts user experiences. As shown in Figure~\ref{fig:recommend}, the exposure features $E$ can influence click $Y$ in two ways: (i) natural direct effect (NDE) that causes clickbait issue, i.e., $E \rightarrow Y$; and (ii) total indirect effect (TIE) as part of item features to influence the prediction score, i.e., $E \rightarrow I \rightarrow Y$. The total effect of exposure features $E$ on prediction score $Y$ is modeled as total effect (TE). The relationship of these three effects is $\operatorname{TE}=\operatorname{NDE}+\operatorname{TIE}$.
Based on this causal graph, \citeauthor{wang2021clicks}~\cite{wang2021clicks} propose CR to alleviate the issue by removing the direct influence of exposure features, i.e., $\operatorname{TIE}=\operatorname{TE}-\operatorname{NDE}$. The influence of item features on prediction score TIE is kept as the prediction score $\hat{Y}$, which provides a desired recommendation without clickbait issue. 

Specifically, %let $u \in \mathcal{U}$ denote a user from the user set and $i \in \mathcal{I}$ denote an item from the item set.
let $Y_{u,i} \in \{0,1\}$ be whether user $u$ clicks on item $i$ ($Y_{u,i}=1$) or not ($Y_{u,i}=0$). The predicted clicking probability of user $u$ on item $i$ is $\hat{Y}_{u,i}=s_{\theta}(u,i)$, with $s_\theta(\cdot,\cdot)$ being a scoring function parameterized by $\theta$. The recommendation task involves learning an effective scoring function on the training set by minimizing the loss:
\begin{equation}
\theta=\underset{\theta}{\arg \min } {\sum}_{\left(u, i, Y_{u, i}\right) \in \mathcal{D}} l\left(s_\theta(u, i), Y_{u, i}\right),
\end{equation}
where $l(\cdot, \cdot)$ is the cross-entropy loss and $\mathcal{D}$ denotes the training set. In the clickbait context, item features comprise exposure features $e$ and content features $t$, i.e., $i = (e, t)$. Both item and exposure features affect the prediction score, as $\hat{Y}_{u,i,e} = f(u,i,e)$. To model the direct effect of one variable on another while keeping other variables fixed, we should maintain $I$ constant and prevent it from contributing to $Y$. The authors use reference value $I=i^*$ to represent the absence of $I$ information, i.e., $Y_{u,i^*,e}:=Y_{u,e}$, to signify that the variable $I$ lacks distinctive characteristics.
% In the implementation, they adopt a fusion strategy $Y_{u,i,e}=Y_{u, i} * \sigma(Y_{u, e})$, enabling the modeling of $Y_{u, e}$ and $Y_{u,i,e}$ within the same model. 
Consequently, the natural direct effect of $I$ on $Y$ can be expressed as $\operatorname{NDE}=Y_{u,i^*,e}-Y_{u,i^*,e^*}$, in the absence of $i^*,e^*$. %Building upon the previous analysis, 
Then, the prediction score can be obtained as $\hat{Y}_{\text{CR}}=\operatorname{TE}-\operatorname{NDE}$. CR estimates NDE and TE as:
\begin{equation}
\fontsize{9pt}{10pt}\selectfont
\operatorname{NDE} =\hat{Y}_{u, i^*, e}-\hat{Y}_{u, i^*, e^*}, \operatorname{TE} =\hat{Y}_{u, i, e}-\hat{Y}_{u, i^*, e^*}.
\end{equation}
Thus, CR acquires the desired prediction by subtracting NDE from TE, i.e.,
\begin{equation}
\fontsize{9pt}{10pt}\selectfont
\hat{Y}_{\text{CR}}=\operatorname{TE}-\operatorname{NDE}=\hat{Y}_{u,i,e}-\hat{Y}_{u,i^*,e}.
\end{equation}
Since user-item interactions are modeled as a graph, CR employs a GNN model called MMGCN~\cite{wei2019mmgcn} as the base model. Both $\hat{Y}_{u,i}=f_{\theta_1}(U=u,I=i)$ and $\hat{Y}_{u,e}=f_{\theta_2}(U=u,E=e)$ are modeled separately, and the fusion strategy is applied as $\hat{Y}_{u,i,e}=\hat{Y}_{u, i} \cdot \sigma\left(\hat{Y}_{u, e}\right)$. To optimize the model, both $\hat{Y}_{u,i,e}$ and $\hat{Y}_{u,i^*,e}$ need improvement. Noting that $\hat{Y}_{u,i^*,e}=\hat{Y}_{u,e}$, considering we only have $Y_{u, i}$ as the supervision, the objective function can be expressed as:
\begin{equation}
\fontsize{8pt}{10pt}\selectfont
\min_{\theta} {\sum}_{(u, i, Y_{u, i}) \in \mathcal{D}} l(\hat{Y}_{u, i, e}, Y_{u, i})+\alpha \cdot l(\hat{Y}_{u, e}, Y_{u, i}),
\end{equation}
where $\alpha$ is a hyper-parameter. Through this design, both $\hat{Y}_{u,i,e}$ and $\hat{Y}_{u,i^*,e}$ can be estimated, enabling the acquisition of an accurate prediction, $\hat{Y}_{\text{CR}}$.

\textbf{Counterfactual Recommendation for Mitigating Geographical Factor}
%UKGC~\cite{liu2021improving} focus on the location recommendation problem, i.e., recommending 
Point-of-Interests (POIs) recommendations are widely used in location-based services~\cite{feng2015personalized, yin2017spatialaware, zhao2016a}.
% \teng{missing citations}.
%Many existing location recommendation models fail to explicitly model the geographical factor, which may lead to sub-optimal recommendation results~\cite{lim2020stpudgat}. 
In many existing location recommendation systems, the interactions between users and POIs are determined by the interests of users and the functional attributes of POIs. Recently, \citeauthor{liu2021improving}~\cite{liu2021improving} notice that the geographical factor works as a confounder of this interaction, i.e., the geographical factor not only affects the POIs, but also influences the user-POI interactions. For example, in downtown, there are many restaurants but Alice prefers to go to restaurant $r$. The reasons may differ: Alice may be attracted by the restaurant, or maybe Alice chooses this restaurant just because it is near her office (geographical factor). Taking this into consideration, \citeauthor{liu2021improving}~\cite{liu2021improving} proposes UKGC, which can remove the direct influence of geographical factor on user-POI interactions with counterfactual learning and make recommendations merely based on user features and POI features. The core idea of UKGC is to model the geographical factor separately with disentangled representation learning, and then use counterfactual learning to remove the direct effect of geographical bias. Specifically, in the first step, UKGC collects various POI information and cities' geographical data from Tencent Map and builds a large-scale Urban Knowledge Graph (UrbanKG). %The graph contains 7 types of entities including POIs, business areas, regions, brands and categories, and 16 types of relations. The relations can be divided into geographical type, i.e., BelongTo, and functional type, i.e., BrandOf. 
It then divides UrbanKG into two subgraphs, the geographical graph and the functional graph, with respect to the type of relations. Both graphs contain the same POIs. UKGC adopts graph convolutions to learn geographical embeddings and functional embeddings for users and POIs then integrates two sets of embeddings via linear combination and gets the final embedding of users and POIs. In the second step, UKGC shares the same design of CR~\cite{wang2021clicks} to mitigate geographical factors with counterfactual learning. Formally, let $Y_{u, p, g}$ denote the recommendation score between use $u$ and POI $p$, where $g$ is the geographical attribute of $p$. Similar to CR, the natural direct effect (NDE) of the geographical factor on recommendation is: 
\begin{equation}
\operatorname{NDE}=Y_{u,p^*,g}-Y_{u,p^*,g^*},    
\end{equation}
where $p^*$ and $g^*$ are the reference value of $p$ and $g$. The reference value signifies that the variable lacks distinctive characteristics, and it is identical for each variable, computed as the average value.
Similarly, the total effect (TE) and total indirect effect (TIE) of geographical factors on recommendation are defined as:
\begin{equation}
\fontsize{9pt}{10pt}\selectfont
\begin{aligned}
        \operatorname{TE} &= Y_{u,p,g}-Y_{u,p^*,g^*},  \\
        \operatorname{TIE} &= \operatorname{TE} - \operatorname{NDE} = Y_{u,p,g}-Y_{u,p^*,g},
\end{aligned}
\end{equation}
where TIE is the desired recommendation score between user $u$ and POI $p$ without the direct influence of geographical factors. The model optimization closely resembles the CR approach~\cite{wang2021clicks}. Interested readers can refer to the original paper~\cite{liu2021improving} for more details.

\textbf{Counterfactual Recommendation for Alleviating Spurious Correlation}
Instead of mitigating a single predetermined spurious factor in recommender systems, CGKR~\cite{mu2022alleviating} aims to integrate information from knowledge graphs (KGs) to simultaneously identify and alleviate potential spurious correlations. For instance, John has watched numerous action movies featuring car chases, causing the recommender system to suggest more movies with car chases for him. However, the actual reason behind John's preference for these action movies is his admiration for a particular actor who often stars in films with car chase scenes. Thus, the co-occurrence patterns may mislead the recommender to make recommendations based on the spurious factor, i.e., the car chase factor in this case. CGKR addresses this problem by designing counterfactual generators. The mutual collaboration mechanism between the recommender and the generators helps to find potential spurious correlations and weaken their influence. The whole process can be illustrated using John's example. Intuitively, knowing that ``John watched a movie'' and the movie features a car chase from the training data, CGKR generates a counterfactual interaction by asking and answering the question, ``Would John still watch the movie if it didn't have a car chase scene?''. If the recommender answers yes, the spurious correlation of car chases will be identified and removed. %The answers mean that the recommender gives a high prediction score.

In the first step, CGKR specifically designs counterfactual generators to modify item attributes, enabling the creation of desired counterfactual interactions. Given the vast space of potential modifications, CGKR employs a Markov Decision Process~\cite{sutton1998reinforcement} (MDP) to streamline the process. To guide the MDP in discovering spurious counterfactual interactions, two rewards are designed: the information-based reward, which helps identify essential attributes, and the prediction-based reward, which facilitates generating more informative counterfactual interactions. For the second step, CGKR employs a knowledge-aware recommender that jointly leverages knowledge graphs and user-item interactions to learn item and user representations. It utilizes an inductive Graph Neural Network (GNN) to learn both factual and counterfactual item representations from the knowledge graph. Subsequently, a GNN is employed to aggregate user-item interactions and obtain user representations. The recommender is supervised to provide similar predictions for both factual and counterfactual interactions, ensuring reliable recommendations without spurious interactions.

\textbf{Counterfactual Recommendation for Using Outer-Session Causes}
Most session-based recommender systems (SBRSs) primarily concentrate on utilizing information from the items viewed in a user's current session to predict the next item, overlooking external factors (referred to as outer-session causes, OSCs) that affect the user's choice of items. In order to solve this issue, COCO-SBRS~\cite{SongW0L0Y23} tries to learn the causality between OSCs and user-item interactions in SBRS.

COCO-SBRS initially pre-trains a recommendation model to understand the relationships between Inner-session causes (ISCs), Outer-session causes (OSCs), and user-item interactions in SBRSs. It then forecasts the next item for a session by considering neighboring sessions that share the same ISCs and OSCs as the current session. The recommendation model is utilized to simulate the user's selection in these neighboring sessions.

\textbf{Counterfactual Explanation in Recommender Systems}
GREASE~\cite{chen2022grease} is a natural extension of GNN counterfactual explanation method CF$^2$~\cite{chen2022grease} for recommender systems. To mitigate the gap between recommendation and explanation, \citeauthor{chen2022grease}~\cite{chen2022grease} propose GREASE to find the sufficient and necessary conditions for an item to be recommended, respectively. 
% For factual explanation, the model tries to answer ``You have interacted with Y so that you are likely to like item X''. In this case, Y is the factual explanation for X. For counterfactual explanation, the model tries to answer ``If you had not interacted with Y, then you would not have recommended item X''. In this case, Y is the counterfactual example for X. 
The proposed GREASE has two steps, i.e., training a surrogate model for the target model and generating explanations with adjacency matrix perturbation. For the first step, GREASE adopts a relational graph convolutional network (R-GCN)~\cite{zhang2020relational} as the surrogate model by minimizing mean squared error of the node embeddings of the target model and that of the surrogate model. For the second step, GREASE defines two adjacency matrix masks for factual explanation and counterfactual explanation, separately. Considering the case that recommends item $i$ to user $u$, the mask for factual explanation is used to perturb the adjacency matrix and seek to maximize the recommendation score for $i$ until $i$ is recommended. Similarly, \citeauthor{chen2022grease} perturb the adjacency matrix with a counterfactual mask and try to minimize the recommendation score until $i$ is no longer recommended. These learned factual and counterfactual masks can be used to indicate the importance of user-item interactions in the adjacency matrix and explain the model.

\section{Real-World Applications of Graph Counterfactual Learning} \label{application}

Due to its superior performance in many tasks~\cite{agarwal2021towards, lucic2022cf}, graph counterfactual learning has wide applications in real-world scenarios. In this section, we will first review  applications in various domains, including physical systems, medical and molecular, which is summarized in Table~\ref{tab:application}. We will then review the adoption of graph counterfactual learning to facilitate other machine learning techniques, which is summarized in Table~\ref{tab:application2}.

\subsection{Application of Graph Counterfactual Learning in Various Domains}
\subsubsection{Physical Systems}

\begin{table*}[t]
\centering
\caption{Summary of real-world applications of graph counterfactual learning.}
\fontsize{6.5pt}{7pt}\selectfont
\begin{tabular}{ccccc}
\toprule
\textbf{Category} & \textbf{Method} & \textbf{Graph Type} & \textbf{Task}  \\
\midrule
\multirow{4}{*}{Physical Systems} & CoPhy~\cite{baradel2019cophy}, V-CDN~\cite{li2020causal}, \citeauthor{li2022deconfounding}~\cite{li2022deconfounding} & Objects with interactions & Position prediction \\
& XHGP~\cite{limeros2022towards} & Vehicles, pedestrians with interactions & Position prediction \\
& TGV-CRN~\cite{fujii2022estimating}, \citeauthor{su2020counterfactual}~\cite{su2020counterfactual} & Multi-agents and interactions & Position prediction \\
& \citeauthor{naderializadeh2020wireless}~\cite{naderializadeh2020wireless} & Wireless interference network & Power allocation \\ \midrule
\multirow{3}{*}{Medical} & CMGE~\cite{wu2021counterfactual} & EMR text graph & Explain diagnosis result \\
& CACHE~\cite{xu2022counterfactual} & Hypergraph of EMR & Explain clinical predictions \\
& \citeauthor{zhang2022counterfactual}~\cite{zhang2022counterfactual} & Medical images and their relations & Disease prediction \\ \midrule
\multirow{2}{*}{Molecular} & SolvGNN~\cite{qin2022capturing} & Molecule graph & Molecule property prediction \\
& MMACE~\cite{wellawatte2022model} & Molecule graph & Explain property prediction \\
\bottomrule
\end{tabular}
\label{tab:application}
\end{table*}

The ability to find the underneath causal relations is a representative feature of intelligent agents. As a human, we have the capacity to discover causal effects in the real-world physical environment. For example, having observed a ball being thrown in a certain manner, we can predict its parabolic trajectory. Counterfactual learning together with graph neural networks makes it possible to model the underlying causal structure of complex systems, enabling intelligent agents to make more accurate predictions and reason about potential interventions~\cite{baradel2019cophy, li2020causal, li2022deconfounding}. 

%In physical systems, inference of the potential trajectories from video inputs is a way to evaluate the model intelligence .
%CoPhy~\cite{baradel2019cophy} is an early work that introduces a novel task in this domain that demands counterfactual reasoning capabilities, as well as a method to tackle this challenge. The objective is to 

In physical systems, predicting alternative future trajectories for objects within a 3D simulation is an important problem~\cite{baradel2019cophy}. Specifically, with a video depicting the motion of objects in one scenario, we aim to predict the trajectory of the object given a single altered frame that modifies object positions at the initial timestep of the input video. The task can be an effective way to evaluate the model performance in terms of counterfactual reasoning and their ability to model complex systems. CoPhy~\cite{baradel2019cophy} and V-CDN~\cite{li2020causal} adopt end-to-end approaches to learn latent representations of confounders and perform counterfactual reasoning, allowing for more accurate predictions of alternative future trajectories when compared to feedforward video prediction baselines.
% \suhang{adopts xxx. Can you briefly talk about the approach proposed by CoPhy. If CoPhy is not relevant to graph counterfactual learning, we just delete it} 
Specifically, these two models build graph structure from the video to model the complex interactions between objects and perform interventions on the graph to get counterfactual outcomes. This enables precise predictions and possible interventions in the real-world physical environment, such as predicting alternative future trajectories of objects in the aforementioned task. 
% \suhang{very confusing. what is the difference? which one has end-to-end, unsupervised manner? how about the other?}. 
Though they are effective in modeling causal relationships and counterfactual reasoning, CoPhy and V-CDN are prone to overfitting.
To alleviate the overfitting issue, \citeauthor{li2022deconfounding}~\cite{li2022deconfounding} takes the unobservable confounders into consideration, such as masses and gravity that are tied to physics laws and independent of the data. By incorporating confounders into the model, \cite{li2022deconfounding} enhances the model's ability to capture the true underlying causal structure, reducing the risk of overfitting.

In addition to the ability to perform counterfactual reasoning, another advantage of counterfactual learning is to help investigate the model behavior in different scenarios. The intuition is that properly modeling the causal relationships of variables in a system can help to identify the critical variables that strongly affect the outcomes~\cite{kaddour2022causal}.
One example is interpretable motion prediction models, which can enable autonomous vehicles to perform safe and transparent decisions~\cite{gulzar2021a}. Specifically, the motion prediction system is to estimate future trajectories of the vehicle of interest with input features describing the past trajectory of the vehicle as well as the scene context.  \citeauthor{limeros2022towards}~\cite{limeros2022towards} propose XHGP, which uses a graph-based model to model heterogeneous objects in traffic scenes and interactions between objects. It adopts counterfactual learning to examine the trained models' sensitivity to changes made in the input data, such as masking scene elements, modifying trajectories, and adding or removing dynamic agents.

Graph counterfactual learning has also been applied to the domain of multi-agent systems. Multi-agent systems involve the interaction and cooperation of multiple autonomous agents, each with their own goals, to achieve a common objective or solve a problem~\cite{dorri2018multiagent}. For example, in the network packet routing problem, multiple autonomous routers work together to forward packets through a network efficiently and quickly~\cite{dorri2018multiagent}. In multi-agent systems, partial observability and communication constraints cause two crucial problems: how to make efficient communications between objects and how to assign credit from a global reward function. \citeauthor{su2020counterfactual}~\cite{su2020counterfactual} tackle the first problem by modeling agents and interactions with graphs and utilizing graph convolution to integrate desired information from neighboring nodes. They settle the second problem by incorporating COMA~\cite{foerster2018counterfactual}, which makes counterfactual reasoning for each agent to understand the sole contribution and assign credit accordingly. \citeauthor{fujii2022estimating}~\cite{fujii2022estimating} propose TGV-CRN to estimate the effect of intervention in multi-agent systems in interpretable ways. They model confounders with graph representation learning techniques and make predictions with the intervened representations to estimate the effect. The result can help researchers understand the potential effect of an intervention, e.g., in autonomous vehicle simulation, what is the potential effect of the acceleration of a vehicle. Wireless power control problem has multiple transmitter-receiver pairs communicating with each other, which can also be treated as a multi-agent system. \citeauthor{naderializadeh2020wireless}~\cite{naderializadeh2020wireless} use a GNN-based framework to learn optimal power allocation decisions with the help of counterfactuals.

\subsubsection{Medical}

In medical domain, it is critical to know the supporting facts of the clinical predictions to ensure the transparency and reliability of the diagnoses and treatments~\cite{wu2021counterfactual}. By employing graph counterfactual learning in medical research, it becomes possible to identify the underlying causal relationships between different variables, such as symptoms, diagnoses, and treatments, allowing for more accurate predictions and evidence-based decision-making~\cite{wu2021counterfactual, xu2022counterfactual, zhang2022counterfactual}.

One example is explaianble Electronic Medical Records (EMR) diagnosis, which aims to make explaianble clinical diagnosis with machine learning models~\cite{huang2020fusion}. The common approach is to treat the EMR as a text sequence and utilize an external expert knowledge base to find explanations for clinical diagnosis. CMGE~\cite{wu2021counterfactual} leverages counterfactual learning to gradually weaken the features until the diagnosis changes dramatically, thus the feature could be considered as the supporting fact. It adopts graph modeling techniques to provide multi-granularity entities, such as sentences, clauses and words, together with the relationships at different levels. Thus, CMGE can avoid external knowledge and try to provide explanations appropriately for the diagnosis solely with the EMR itself. Facing the same problem of providing explanations for EMR-based clinical diagnosis, CACHE~\cite{xu2022counterfactual} models the relationships between medical entities with hypergraphs. Hypergraphs are generalizations of traditional graph structure that allow edges to connect more than two nodes, which are called hyperedges. In CMGE, hyperedges connect multiple entities, such as chest pain, obesity and gout together, to represent the complex relations among these units. Counterfactual learning is also used to find features that cause the change of predictions as the critical supporting fact. In~\cite{zhang2022counterfactual}, \citeauthor{zhang2022counterfactual} endeavor to improve the disease prediction performance by learning the cause-effect relationship from data. They ask counterfactual questions ``Would someone be inferred to be free of the disease if he/she has some attributes?''. By answering the counterfactual questions, their model can learn the latent relationships and integrate knowledge into their prediction model.

%In addition to the EMR or other single source of information-based research, 
\citeauthor{holzinger2021towards}~\cite{holzinger2021towards} highlight the importance of incorporating various modalities in medical domain. They propose using GNNs to construct multi-model representations, spanning images, text and genomics data, where each sample works as a node and they are connected based on various relationships. The flexibility of GNNs can effectively enable information fusion for multi models. In this process, counterfactual learning works as an interpretable tool to investigate the model behavior in various scenarios by exploring how changes made to the input data affect the outcomes.

\subsubsection{Molecular}

In molecular studies, it is important to understand the impact of functional groups and composition on equilibrium behavior~\cite{wang2022molecular}. A deep understanding of this problem can greatly help in designing new materials, optimizing chemical processes, and predicting the properties of complex systems. GNNs have gained wide popularity in molecular studies due to their ability to capture and represent intricate relationships between different molecular components in multi-component systems~\cite{wieder2020a}. However, the lack of transparency severely limits the border application of GNNs in molecular studies~\cite{aouichaoui2023combining}. SolvGNN~\cite{qin2022capturing} tackles this problem by incorporating counterfactual learning. It performs counterfactual analysis by perturbing the input graph structure until the prediction changes. The counterfactual analysis helps gain a further understanding of which chemical structures and functional groups can lead to certain activity coefficient predictions. Continuing this line of research, MMACE~\cite{wellawatte2022model} is proposed to generate experimentally available molecules as counterfactuals by enumerating chemical space. Thus, MMACE is independent of the target model architecture and can be applied to explain various models.

\subsection{Graph Counterfactual Learning for Other Techniques}

\begin{table}
\scriptsize
    \centering
    \caption{Graph counterfactual learning for other techniques.}
    \begin{tabular}{cc}
    \toprule
     \textbf{Techniques}    &  \textbf{References} \\
     \midrule
     Transfer Learning & \citeauthor{li2022counterfactual}~\cite{li2022counterfactual} \\
     \midrule
     Data Augmentation & CRL~\cite{pham2022counterfactual} \\
     \midrule
     Anomaly Detection & CFAD~\cite{xiao2023counterfactual} \\
    \bottomrule
    \end{tabular}
    \label{tab:application2}
\end{table}

\subsubsection{Transfer Learning}

Transfer learning aims to transfer knowledge from source domains to target domains to facilitate the task in the target domain. \citeauthor{li2022counterfactual}~\cite{li2022counterfactual} focus on a specific setting, the relational transfer learning task, which extracts semantic-meaningful entities to build knowledge graphs in the source domain and then transfer the structured knowledge to the target domain. However, a critical challenge of relational transfer learning is that it needs support from human experts to select appropriate relations to transfer, which is time-consuming and laborious. \citeauthor{li2022counterfactual}~\cite{li2022counterfactual} tackle this problem by employing counterfactual inference. By performing counterfactual inference, one can discover the latent causal relationships, which can help to predict causal knowledge graphs from the source domain to the target domain for relational transfer learning.

\subsubsection{Data Augmentation}

In previous sections, we have shown the effectiveness of performing counterfactual augmentation, such as counterfactual link prediction~\cite{zhao2022learning}. The intuition of is that the counterfactual augmented data can help the model learn the latent causal relationships better. For example, in the image classifications task, an image with a bird on it is labeled as ``bird'' class. Counterfactual augmentation aims to remove the bird in the image and label the augmented image as ``non-bird'' class.  By doing so, the model learns to identify the bird as the key  causal feature associated with the ``bird'' class, thereby improving its understanding of the causal relationships between image features and class labels. In CRL~\cite{pham2022counterfactual}, \citeauthor{pham2022counterfactual} employ counterfactual learning in a reinforcement learning-based manner for graph classification task. They integrate external domain knowledge to guide the counterfactual data augmentation and then use GNNs to learn from both the original and counterfactual augmented data.

\subsubsection{Anomaly Detection}

Anomaly detection in attributed networks aims to identify unusual or suspicious nodes within a network that contains both topological structure and node attribute. It is widely adopted in various domains such as social networks and transaction networks~\cite{ma2021a}. However, distribution discrepancy of training data and test data may hurt the generalization ability of existing anomaly detection approaches since they are trained by learning environment-dependent correlations. That means, training nodes and test nodes are different with respect to their environments, e.g., subgraph topology and neighboring nodes.  To solve this problem, CFAD~\cite{xiao2023counterfactual} constructs the causal graph to learn the causal relations among latent and observed variables. Specifically, CFAD constructs the counterfactual examples in different environments to help the model generalize to unseen scenarios. With the root causes captured and spurious information filtered, CFAD can make robust and stable predictions in anomaly detection setting.

\section{Future Directions} \label{future}

Despite various efforts taken on graph counterfactual learning, it is still in its early stage. Existing approaches mainly focus on static graphs and lack the ability to generalize to other settings, such as large-scale graphs and dynamic graphs. Moreover, most existing datasets and metrics are developed for factual learning, not dedicated to counterfactual learning, which may not accurately reflect the performance of counterfactual learning methods. Additionally, some directions are not yet fully explored and require further investigation, such as unsupervised graph counterfactual learning and counterfactual data augmentation. We summarize the existing open-source implementations in Table~\ref{tab:implement} to help readers have a better knowledge of current progress. Next, we give the details of future works, aiming to provide researchers with insights into crucial directions in this field.

\begin{table*}[t]
    \tiny
    \scriptsize
    % \small
    \centering
    \caption{Summary of open-source implementations.}
    \resizebox{0.9\textwidth}{!}{\begin{tabular}{llll}
    \toprule
    \textbf{Methods}  & \textbf{Category} & \textbf{Framework} &  \textbf{Code} \\
    \midrule
    GEAR~\cite{ma2022learning} & Fairness & PyTorch &\url{https://github.com/jma712/GEAR} \\
    \midrule
    NIFTY~\cite{agarwal2021towards} & Fairness & PyTorch & \url{https://github.com/chirag126/nifty} \\
    \midrule
    CAF~\cite{Guo0X0W23} & Fairness & PyTorch & \url{https://github.com/TimeLovercc/CAF-GNN} \\
    \midrule
    GraphXAI~\cite{agarwal2023evaluating} & Explanation & PyTorch & \url{https://github.com/mims-harvard/GraphXAI} \\
    \midrule
    GCFExplainer~\cite{huang2023global} & Explanation & PyTorch & \url{https://github.com/mertkosan/GCFExplainer} \\
    \midrule
    \citeauthor{wellawatte2022model}~\cite{wellawatte2022model} & Explanation & PyTorch & \url{https://github.com/ur-whitelab/exmol} \\
    \midrule
    GRETEL~\cite{prado2022gretel} & Explanation & TensorFlow & \url{https://github.com/MarioTheOne/GRETEL} \\
    \midrule
    CF$^2$~\cite{verma2020counterfactual} & Explanation & PyTorch & \url{https://github.com/chrisjtan/gnn_cff} \\
    \midrule
    CF-GNNExplainer~\cite{lucic2022cf} & Explanation & PyTorch & \url{https://github.com/a-lucic/cf-gnnexplainer} \\
    \midrule
    RCExplainer~\cite{bajaj2021robust} & Explanation & PyTorch & \url{https://github.com/RomanOort/FACTAI} \\
    \midrule
    MEG~\cite{numeroso2021meg} & Explanation & PyTorch & \url{https://github.com/danilonumeroso/meg} \\
    \midrule
    \citeauthor{abrate2021counterfactual}~\cite{abrate2021counterfactual} & Explanation & PyTorch & \url{https://github.com/carlo-abrate/CounterfactualGraphs} \\
    \midrule
    CFLP~\cite{zhao2022learning} & Link Prediction & PyTorch & \url{https://github.com/DM2-ND/CFLP} \\
    \midrule
    CR~\cite{wang2021clicks} & Recommendation & PyTorch & \url{https://github.com/WenjieWWJ/Clickbait/} \\
    \midrule
    CGKR~\cite{mu2022alleviating} & Recommendation & PyTorch & \url{https://github.com/RUCAIBox/CGKR} \\
    \midrule
    XHGP~\cite{limeros2022towards} & Physical System & PyTorch & \url{https://github.com/sancarlim/Explainable-MP/tree/v1.1} \\
    \midrule
    TGV-CRN~\cite{fujii2022estimating} & Physical System & PyTorch & \url{https://github.com/TGV-CRN/TGV-CRN} \\
    \midrule
    V-CDN~\cite{li2020causal} & Physical System & PyTorch & \url{https://github.com/pairlab/v-cdn} \\
    \midrule
    CoPhy~\cite{baradel2019cophy} & Physical System & PyTorch & \url{https://projet.liris.cnrs.fr/cophy/} \\
    \midrule
    SolvGNN~\cite{qin2022capturing} & Molecular & PyTorch & \url{https://github.com/zavalab/ML/tree/SolvGNN} \\
    \midrule
    CMGE~\cite{wu2021counterfactual} & Medical & PyTorch & \url{https://github.com/CKRE/CMGE/blob/main/node_mask/gat.py} \\
    \bottomrule
    \end{tabular}}

    \label{tab:implement}
\end{table*}

\begin{itemize}[]
%\subsection{Benchmark Datasets}
\item{\bf Benchmark Datasets.} A well-curated dataset can significantly accelerate the development of a research area~\cite{deng2009imagenet}. Currently, graph counterfactual learning lacks benchmark datasets. For real-world datasets, we typically have access only to observed data, making it challenging to evaluate the performance of counterfactual learning methods. Thus, one important future research direction is constructing synthetic or semi-synthetic datasets that encompass various characteristics of real-world graphs while providing ground truth for counterfactual scenarios~\cite{kaddour2022causal}. For example, in counterfactual fairness, researchers may come up with synthetic social networks where each node has node attributes and connections in the counterfactual world where the node sensitive attribute is flipped. For semi-synthetic datasets, in computer vision, researchers use deepfake to change the face in an image to a different gender~\cite{westerlund2019the}, such that they can get the counterfactuals for fair face recognition tasks. For graph domain, we need to design specific approaches to generate counterfactual nodes or edges within the context of a specific application or domain. In addition, the datasets should be diverse and cover a wide range of applications, such as social networks, biological systems, and financial markets, to enable a comprehensive evaluation of different graph counterfactual learning techniques. 

%\subsection{Evaluation Metrics}
\item{\bf Evaluation Metrics.} Another important problem is that graph counterfactual learning lacks dedicated evaluation metrics. Currently, existing graph counterfactual learning methods adopt evaluation metrics designed for conventional graph learning. For example, statistical parity is a group fairness metric and most of graph counterfactual learning works also adopt this metric for counterfactual fairness. Although it can effectively compare the prediction result of different sub-populations, it has been found not able to detect discrimination in presence of statistical anomalies~\cite{makhlouf2020survey}. For graph counterfactual explanation, it also faces the same problem that existing metrics may not be well-suited for evaluating the quality of counterfactual explanations generated for graph-structured data. To address these issues, the development of tailored metrics to facilitate evaluation of counterfactual learning on graphs is a promising and urgent direction. These metrics should align well with the causal structure of the problem formulation.

\item{\bf Counterfactual Data Augmentation.} Counterfactual learning typically involves: (a) modeling counterfactual outcomes~\cite{zhao2022learning, pham2022counterfactual}; (b) identifying root causes of model predictions~\cite{chen2022grease, numeroso2021meg}. Counterfactual augmentation focuses on modeling outcomes, providing additional knowledge to enhance input-prediction relationships. For example, in animal image classification, counterfactual augmentation removes a bird, labeling the altered image as ``non-bird.'' In graph domains, such as recommendations based on user connections, augmentation may involve removing a user's influential connection and labeling altered graph as a negative example. This highlights certain connections and improves causal understanding between connections and labels. Despite limited initial efforts in graph domains~\cite{zhao2022learning}, ample opportunities exist for further research, including tailored strategies for specific applications like recommendation systems~\cite{ying2018graph,xiao2022towards,xiao2022towards,zhang2019deep,xiao2019hierarchical}, social network analysis~\cite{bhagat2011node,xiao2022decoupled}, or molecular graph models~\cite{xiong2019pushing}.

%\subsection{Scalability}
\item{\bf Scalability.} 
%Scalability is another important direction in graph domain, especially considering the size of many real-world graphs~\cite{jia2020improving}. 
Large-scale graphs are pervasive in real-world such as Facebook or Twitter~\cite{traud2012social, suh2010want}, web graphs~\cite{mernyei2020wikics} and knowledge graph~\cite{shalaby2019beyond, wang2018acekg}. Due to the complex graph structure, the iterative process of generating counterfactuals and the need to retrain models with augmented data, existing graph counterfactual learning approaches are time-consuming and often struggle with scalability, making them unsuitable for large-scale graphs~\cite{frasca2020sign}. %The time-consuming nature of these methods can be attributed to several factors, such as the  
Despite the importance of scalability, works on addressing this issue in graph counterfactual learning are relatively limited.

%\subsection{Dynamicity}
\item{\bf Dynamicity.} %Dynamicity is another essential aspect to consider when examining graph counterfactual learning methods, 
Many real-world graphs are inherently dynamic~\cite{kazemi2020representation}, such as social networks where new relationships form and old ones dissolve over time~\cite{Perry2003social}; transportation networks where traffic patterns and infrastructure change~\cite{ran1996modeling} and user-item interaction networks in recommender systems where relationships between users and items evolve ~\cite{xiao2019dynamic,xiao2019dynamic}. However, existing graph counterfactual learning approaches mainly focus on static graphs. To deal with the dynamic graphs effectively, we need to address several challenges, such as the complexity of modeling the graph structure over time and the lack of high-quality dynamic graph datasets.

%\subsection{Unsupervised Learning}
\item{\bf Unsupervised Learning.} Unsupervised/self-supervised learning on graphs have achieved comparable performance with supervised methods~\cite{you2020graph, zhang2021from,xiao2022decoupled}. However, there are very few explorations for unsupervised/self-supervised graph counterfactual learning. The lack of labeled data in unsupervised setting makes it difficult to investigate the causal structure of latent variables and observed variables. Moreover, it is challenging to generate counterfactual outcomes and compare them with factual outcomes without the necessary ground truth label~\cite{wang2021desiderata}. 
Given the success of unsupervised learning in graph domain, it is crucial to address this gap and explore innovative approaches for unsupervised graph counterfactual learning. %Some potential topics can be: (1) Adapting unsupervised generative models, such as Variational Autoencoders (VAEs)~\cite{kingma2013autoencoding}. These models can learn to generate realistic graph samples, enabling the estimation of counterfactual outcomes and causal effects without labeled data. (2) Investigating the use of other self-supervised learning techniques, like contrastive learning~\cite{you2021graph}. These techniques can learn meaningful representations by contrasting positive and negative samples, which could potentially be used to extract meaningful representations for latent variables and observed variables, together with the causal relationships among them. (3) Exploiting the inherent structure and connectivity of graphs or integrating external domain knowledge for unsupervised graph counterfactual learning. The internal knowledge and external knowledge can help to guide the discovery of causal relationships and enable more effective graph counterfactual learning, even in the absence of labeled data~\cite{kaddour2022causal}.
\end{itemize}

\section{Conclusion} \label{con}

% In this survey, we conduct a comprehensive review on the topic of counterfactual learning on graphs from different categories, counterfactual fairness, counterfactual explanation, counterfactual link prediction and recommendation and applications. As we know this is the first survey to talk about counterfactual learning on graphs. We want to use this survey to give starters the fundamental knowledge in this area and inspire the domain experts to solve the urgent challenges in this area. Concretely, for counterfactual fairness, counterfactual explanation, counterfactual link prediction and recommendation we use general framework to give unified understanding on the problem. We also summarise the datasets and metrics used in each category. Then we go beyond the counterfactual learning on graphs to its applications in many areas, such as physics, chemical science and etc. Finally, we propose the future researcher directions and encourage the domain experts to contribute to essential and urgent topics in this area. 
In this survey, we present a comprehensive review of counterfactual learning on graphs from the problems of counterfactual fairness, counterfactual explanation, counterfactual link prediction and the real-world applications of graph counterfactual learning. This is the first survey for counterfactual learning on graphs. In particular, we first introduce the basic concept of counterfactual learning, then introduce a framework to give a unified understanding of the problems. We also summarise the datasets and metrics used in each category. Then we go beyond the counterfactual learning on graphs to its applications in many areas, such as physical systems, medical, etc. Finally, we also discuss future directions and encourage domain experts to contribute to essential and urgent topics in this area. We believe this survey can give starters the fundamental knowledge and inspire the domain experts to solve the urgent challenges in this area.

% \section*{Acknowledgments}

% Acknowledgments are not compulsory. Where included they should be brief. Grant or contribution numbers may be acknowledged. The form should be:

% This work was supported by ***(No. ***) and ***(No. ***).

% Please refer to Journal-level guidance for any specific requirements.

% \section*{Declarations of Conflict of Interest}

% The authors declared that they have no conflicts of interest to this work.

% \noindent

%%===========================================================================================%%
%% If you are submitting to one of the Nature Portfolio journals, using the eJP submission   %%
%% system, please include the references within the manuscript file itself. You may do this  %%
%% by copying the reference list from your .bbl file, paste it into the main manuscript .tex %%
%% file, and delete the associated \verb+\bibliography+ commands.                            %%
%%===========================================================================================%%

%\bibliography{sn-bibliography}% common bib file
%% if required, the content of .bbl file can be included here once bbl is generated
%%\input sn-article.bbl

%% Default %%
%%\input sn-sample-bib.tex%
\bibliographystyle{ACM-Reference-Format}
\bibliography{a.bib}

%%% -*-BibTeX-*-
%%% Do NOT edit. File created by BibTeX with style
%%% ACM-Reference-Format-Journals [18-Jan-2012].

\begin{thebibliography}{240}

%%% ====================================================================
%%% NOTE TO THE USER: you can override these defaults by providing
%%% customized versions of any of these macros before the \bibliography
%%% command.  Each of them MUST provide its own final punctuation,
%%% except for \shownote{}, \showDOI{}, and \showURL{}.  The latter two
%%% do not use final punctuation, in order to avoid confusing it with
%%% the Web address.
%%%
%%% To suppress output of a particular field, define its macro to expand
%%% to an empty string, or better, \unskip, like this:
%%%
%%% \newcommand{\showDOI}[1]{\unskip}   % LaTeX syntax
%%%
%%% \def \showDOI #1{\unskip}           % plain TeX syntax
%%%
%%% ====================================================================

\ifx \showCODEN    \undefined \def \showCODEN     #1{\unskip}     \fi
\ifx \showDOI      \undefined \def \showDOI       #1{#1}\fi
\ifx \showISBNx    \undefined \def \showISBNx     #1{\unskip}     \fi
\ifx \showISBNxiii \undefined \def \showISBNxiii  #1{\unskip}     \fi
\ifx \showISSN     \undefined \def \showISSN      #1{\unskip}     \fi
\ifx \showLCCN     \undefined \def \showLCCN      #1{\unskip}     \fi
\ifx \shownote     \undefined \def \shownote      #1{#1}          \fi
\ifx \showarticletitle \undefined \def \showarticletitle #1{#1}   \fi
\ifx \showURL      \undefined \def \showURL       {\relax}        \fi
% The following commands are used for tagged output and should be
% invisible to TeX
\providecommand\bibfield[2]{#2}
\providecommand\bibinfo[2]{#2}
\providecommand\natexlab[1]{#1}
\providecommand\showeprint[2][]{arXiv:#2}

\bibitem[Abbas et~al\mbox{.}(2021)]%
        {abbas2021application}
\bibfield{author}{\bibinfo{person}{Khushnood Abbas}, \bibinfo{person}{Alireza Abbasi}, \bibinfo{person}{Shi Dong}, \bibinfo{person}{Ling Niu}, \bibinfo{person}{Laihang Yu}, \bibinfo{person}{Bolun Chen}, \bibinfo{person}{Shi-Min Cai}, {and} \bibinfo{person}{Qambar Hasan}.} \bibinfo{year}{2021}\natexlab{}.
\newblock \showarticletitle{Application of network link prediction in drug discovery}.
\newblock \bibinfo{journal}{\emph{BMC bioinformatics}}  \bibinfo{volume}{22} (\bibinfo{year}{2021}), \bibinfo{pages}{1--21}.
\newblock


\bibitem[Abrate and Bonchi(2021)]%
        {abrate2021counterfactual}
\bibfield{author}{\bibinfo{person}{Carlo Abrate} {and} \bibinfo{person}{Francesco Bonchi}.} \bibinfo{year}{2021}\natexlab{}.
\newblock \showarticletitle{Counterfactual graphs for explainable classification of brain networks}. In \bibinfo{booktitle}{\emph{Proceedings of the 27th ACM SIGKDD Conference on Knowledge Discovery \& Data Mining}}. \bibinfo{pages}{2495--2504}.
\newblock


\bibitem[Agarwal et~al\mbox{.}(2021)]%
        {agarwal2021towards}
\bibfield{author}{\bibinfo{person}{Chirag Agarwal}, \bibinfo{person}{Himabindu Lakkaraju}, {and} \bibinfo{person}{Marinka Zitnik}.} \bibinfo{year}{2021}\natexlab{}.
\newblock \showarticletitle{Towards a unified framework for fair and stable graph representation learning}. In \bibinfo{booktitle}{\emph{Uncertainty in Artificial Intelligence}}. PMLR, \bibinfo{pages}{2114--2124}.
\newblock


\bibitem[Agarwal et~al\mbox{.}(2023)]%
        {agarwal2023evaluating}
\bibfield{author}{\bibinfo{person}{Chirag Agarwal}, \bibinfo{person}{Owen Queen}, \bibinfo{person}{Himabindu Lakkaraju}, {and} \bibinfo{person}{Marinka Zitnik}.} \bibinfo{year}{2023}\natexlab{}.
\newblock \showarticletitle{Evaluating explainability for graph neural networks}.
\newblock \bibinfo{journal}{\emph{Scientific Data}} \bibinfo{volume}{10}, \bibinfo{number}{1} (\bibinfo{year}{2023}), \bibinfo{pages}{144}.
\newblock


\bibitem[Agarwal et~al\mbox{.}(2022)]%
        {agarwal2022probing}
\bibfield{author}{\bibinfo{person}{Chirag Agarwal}, \bibinfo{person}{Marinka Zitnik}, {and} \bibinfo{person}{Himabindu Lakkaraju}.} \bibinfo{year}{2022}\natexlab{}.
\newblock \showarticletitle{Probing gnn explainers: A rigorous theoretical and empirical analysis of gnn explanation methods}. In \bibinfo{booktitle}{\emph{International Conference on Artificial Intelligence and Statistics}}. PMLR, \bibinfo{pages}{8969--8996}.
\newblock


\bibitem[{Anonymous}(2023)]%
        {anonymous2023learning}
\bibfield{author}{\bibinfo{person}{{Anonymous}}.} \bibinfo{year}{2023}\natexlab{}.
\newblock \showarticletitle{Learning {Fair} {Graph} {Representations} via {Automated} {Data} {Augmentations}}.
\newblock \bibinfo{journal}{\emph{openreview.net}} (\bibinfo{year}{2023}).
\newblock


\bibitem[Aouichaoui et~al\mbox{.}(2023)]%
        {aouichaoui2023combining}
\bibfield{author}{\bibinfo{person}{Adem R.~N. Aouichaoui}, \bibinfo{person}{Fan Fan}, \bibinfo{person}{Seyed~Soheil Mansouri}, \bibinfo{person}{Jens Abildskov}, {and} \bibinfo{person}{G{\" u}rkan Sin}.} \bibinfo{year}{2023}\natexlab{}.
\newblock \showarticletitle{Combining {Group}-{Contribution} {Concept} and {Graph} {Neural} {Networks} {Toward} {Interpretable} {Molecular} {Property} {Models}.}
\newblock \bibinfo{journal}{\emph{Journal of Chemical Information and Modeling}} \bibinfo{volume}{63}, \bibinfo{number}{3} (\bibinfo{year}{2023}), \bibinfo{pages}{725--744}.
\newblock


\bibitem[Arjovsky et~al\mbox{.}(2019)]%
        {arjovsky2019invariant}
\bibfield{author}{\bibinfo{person}{Martin Arjovsky}, \bibinfo{person}{Leon Bottou}, \bibinfo{person}{Ishaan Gulrajani}, {and} \bibinfo{person}{David Lopez-Paz}.} \bibinfo{year}{2019}\natexlab{}.
\newblock \showarticletitle{Invariant risk minimization}.
\newblock \bibinfo{journal}{\emph{arXiv}} (\bibinfo{year}{2019}).
\newblock


\bibitem[Artelt and Hammer(2019)]%
        {artelt2019computation}
\bibfield{author}{\bibinfo{person}{Andr{\'e} Artelt} {and} \bibinfo{person}{Barbara Hammer}.} \bibinfo{year}{2019}\natexlab{}.
\newblock \showarticletitle{On the computation of counterfactual explanations--A survey}.
\newblock \bibinfo{journal}{\emph{arXiv preprint arXiv:1911.07749}} (\bibinfo{year}{2019}).
\newblock


\bibitem[Asuncion and Newman(2007)]%
        {asuncion2007uci}
\bibfield{author}{\bibinfo{person}{Arthur Asuncion} {and} \bibinfo{person}{David Newman}.} \bibinfo{year}{2007}\natexlab{}.
\newblock \showarticletitle{UCI machine learning repository}.
\newblock  (\bibinfo{year}{2007}).
\newblock


\bibitem[Bajaj et~al\mbox{.}(2021)]%
        {bajaj2021robust}
\bibfield{author}{\bibinfo{person}{Mohit Bajaj}, \bibinfo{person}{Lingyang Chu}, \bibinfo{person}{Zi~Yu Xue}, \bibinfo{person}{Jian Pei}, \bibinfo{person}{Lanjun Wang}, \bibinfo{person}{Peter Cho-Ho Lam}, {and} \bibinfo{person}{Yong Zhang}.} \bibinfo{year}{2021}\natexlab{}.
\newblock \showarticletitle{Robust counterfactual explanations on graph neural networks}.
\newblock \bibinfo{journal}{\emph{Advances in Neural Information Processing Systems}}  \bibinfo{volume}{34} (\bibinfo{year}{2021}), \bibinfo{pages}{5644--5655}.
\newblock


\bibitem[Baldassarre and Azizpour(2019)]%
        {baldassarre2019explainability}
\bibfield{author}{\bibinfo{person}{Federico Baldassarre} {and} \bibinfo{person}{Hossein Azizpour}.} \bibinfo{year}{2019}\natexlab{}.
\newblock \showarticletitle{Explainability techniques for graph convolutional networks}.
\newblock \bibinfo{journal}{\emph{arXiv preprint arXiv:1905.13686}} (\bibinfo{year}{2019}).
\newblock


\bibitem[Baradel et~al\mbox{.}(2019)]%
        {baradel2019cophy}
\bibfield{author}{\bibinfo{person}{Fabien Baradel}, \bibinfo{person}{Natalia Neverova}, \bibinfo{person}{Julien Mille}, \bibinfo{person}{Greg Mori}, {and} \bibinfo{person}{Christian Wolf}.} \bibinfo{year}{2019}\natexlab{}.
\newblock \showarticletitle{Cophy: Counterfactual learning of physical dynamics}.
\newblock \bibinfo{journal}{\emph{arXiv preprint arXiv:1909.12000}} (\bibinfo{year}{2019}).
\newblock


\bibitem[Barbieri et~al\mbox{.}(2014)]%
        {barbieri2014who}
\bibfield{author}{\bibinfo{person}{Nicola Barbieri}, \bibinfo{person}{Francesco Bonchi}, {and} \bibinfo{person}{Giuseppe Manco}.} \bibinfo{year}{2014}\natexlab{}.
\newblock \showarticletitle{Who to follow and why: link prediction with explanations}. In \bibinfo{booktitle}{\emph{Proceedings of the 20th {ACM} {SIGKDD} international conference on {Knowledge} discovery and data mining}}. \bibinfo{pages}{1266--1275}.
\newblock


\bibitem[{Barredo Arrieta} et~al\mbox{.}(2020)]%
        {BARREDOARRIETA202082}
\bibfield{author}{\bibinfo{person}{Alejandro {Barredo Arrieta}}, \bibinfo{person}{Natalia Díaz-Rodríguez}, \bibinfo{person}{Javier {Del Ser}}, \bibinfo{person}{Adrien Bennetot}, \bibinfo{person}{Siham Tabik}, \bibinfo{person}{Alberto Barbado}, \bibinfo{person}{Salvador Garcia}, \bibinfo{person}{Sergio Gil-Lopez}, \bibinfo{person}{Daniel Molina}, \bibinfo{person}{Richard Benjamins}, \bibinfo{person}{Raja Chatila}, {and} \bibinfo{person}{Francisco Herrera}.} \bibinfo{year}{2020}\natexlab{}.
\newblock \showarticletitle{Explainable Artificial Intelligence (XAI): Concepts, taxonomies, opportunities and challenges toward responsible AI}.
\newblock \bibinfo{journal}{\emph{Information Fusion}}  \bibinfo{volume}{58} (\bibinfo{year}{2020}), \bibinfo{pages}{82--115}.
\newblock
\showISSN{1566-2535}
\urldef\tempurl%
\url{https://doi.org/10.1016/j.inffus.2019.12.012}
\showDOI{\tempurl}


\bibitem[Bhagat et~al\mbox{.}(2011)]%
        {bhagat2011node}
\bibfield{author}{\bibinfo{person}{Smriti Bhagat}, \bibinfo{person}{Graham Cormode}, {and} \bibinfo{person}{S Muthukrishnan}.} \bibinfo{year}{2011}\natexlab{}.
\newblock \showarticletitle{Node classification in social networks}.
\newblock In \bibinfo{booktitle}{\emph{Social network data analytics}}. \bibinfo{publisher}{Springer}, \bibinfo{pages}{115--148}.
\newblock


\bibitem[Bickel et~al\mbox{.}(1975)]%
        {bickel1975sex}
\bibfield{author}{\bibinfo{person}{Peter~J Bickel}, \bibinfo{person}{Eugene~A Hammel}, {and} \bibinfo{person}{J~William O'Connell}.} \bibinfo{year}{1975}\natexlab{}.
\newblock \showarticletitle{Sex {Bias} in {Graduate} {Admissions}: Data from {Berkeley}: Measuring bias is harder than is usually assumed, and the evidence is sometimes contrary to expectation.}
\newblock \bibinfo{journal}{\emph{Science}} \bibinfo{volume}{187}, \bibinfo{number}{4175} (\bibinfo{year}{1975}), \bibinfo{pages}{398--404}.
\newblock


\bibitem[Bishop and Nasrabadi(2007)]%
        {bishop2007pattern}
\bibfield{author}{\bibinfo{person}{Christopher~M. Bishop} {and} \bibinfo{person}{Nasser~M. Nasrabadi}.} \bibinfo{year}{2007}\natexlab{}.
\newblock \bibinfo{booktitle}{\emph{Pattern {Recognition} and {Machine} {Learning}.}} Vol.~\bibinfo{volume}{16}.
\newblock \bibinfo{publisher}{Springer}. 049901 pages.
\newblock


\bibitem[Blondel et~al\mbox{.}(2008)]%
        {blondel2008fast}
\bibfield{author}{\bibinfo{person}{Vincent~D Blondel}, \bibinfo{person}{Jean-Loup Guillaume}, \bibinfo{person}{Renaud Lambiotte}, {and} \bibinfo{person}{Etienne Lefebvre}.} \bibinfo{year}{2008}\natexlab{}.
\newblock \showarticletitle{Fast unfolding of communities in large networks}.
\newblock \bibinfo{journal}{\emph{Journal of Statistical Mechanics: Theory and Experiment}} \bibinfo{volume}{2008}, \bibinfo{number}{10} (\bibinfo{year}{2008}), \bibinfo{pages}{P10008}.
\newblock


\bibitem[Bose and Hamilton(2019)]%
        {bose2019compositional}
\bibfield{author}{\bibinfo{person}{Avishek~Joey Bose} {and} \bibinfo{person}{William~L. Hamilton}.} \bibinfo{year}{2019}\natexlab{}.
\newblock \showarticletitle{Compositional {Fairness} {Constraints} for {Graph} {Embeddings}}.
\newblock  (\bibinfo{year}{2019}).
\newblock


\bibitem[Brown et~al\mbox{.}(2012)]%
        {brown2012the}
\bibfield{author}{\bibinfo{person}{Jesse~A Brown}, \bibinfo{person}{Jeffrey~D Rudie}, \bibinfo{person}{Anita Bandrowski}, \bibinfo{person}{John D~Van Horn}, {and} \bibinfo{person}{Susan~Y Bookheimer}.} \bibinfo{year}{2012}\natexlab{}.
\newblock \showarticletitle{The {UCLA} multimodal connectivity database: a web-based platform for brain connectivity matrix sharing and analysis}.
\newblock \bibinfo{journal}{\emph{Frontiers in neuroinformatics}}  \bibinfo{volume}{6} (\bibinfo{year}{2012}), \bibinfo{pages}{28}.
\newblock


\bibitem[Bullmore and Bassett(2011)]%
        {bullmore2011brain}
\bibfield{author}{\bibinfo{person}{Edward~T Bullmore} {and} \bibinfo{person}{Danielle~S Bassett}.} \bibinfo{year}{2011}\natexlab{}.
\newblock \showarticletitle{Brain graphs: graphical models of the human brain connectome}.
\newblock \bibinfo{journal}{\emph{Annual review of clinical psychology}}  \bibinfo{volume}{7} (\bibinfo{year}{2011}), \bibinfo{pages}{113--140}.
\newblock


\bibitem[Cai et~al\mbox{.}(2022a)]%
        {cai2022fpgnn}
\bibfield{author}{\bibinfo{person}{Hanxuan Cai}, \bibinfo{person}{Huimin Zhang}, \bibinfo{person}{Duancheng Zhao}, \bibinfo{person}{Jingxing Wu}, {and} \bibinfo{person}{Ling Wang}.} \bibinfo{year}{2022}\natexlab{a}.
\newblock \showarticletitle{FP-{GNN}: a versatile deep learning architecture for enhanced molecular property prediction}.
\newblock \bibinfo{journal}{\emph{Briefings in bioinformatics}} (\bibinfo{year}{2022}).
\newblock


\bibitem[Cai et~al\mbox{.}(2022b)]%
        {cai2022on}
\bibfield{author}{\bibinfo{person}{Ruichu Cai}, \bibinfo{person}{Yuxuan Zhu}, \bibinfo{person}{Xuexin Chen}, \bibinfo{person}{Yuan Fang}, \bibinfo{person}{Min Wu}, \bibinfo{person}{Jie Qiao}, {and} \bibinfo{person}{Z. Hao}.} \bibinfo{year}{2022}\natexlab{b}.
\newblock \showarticletitle{On the {Probability} of {Necessity} and {Sufficiency} of {Explaining} {Graph} {Neural} {Networks}: A {Lower} {Bound} {Optimization} {Approach}}.
\newblock \bibinfo{journal}{\emph{ArXiv}}  \bibinfo{volume}{abs/2212.07056} (\bibinfo{year}{2022}).
\newblock


\bibitem[Carvalho et~al\mbox{.}(2019)]%
        {carvalho2019machine}
\bibfield{author}{\bibinfo{person}{Diogo~V. Carvalho}, \bibinfo{person}{Eduardo~M. Pereira}, {and} \bibinfo{person}{Jaime~S. Cardoso}.} \bibinfo{year}{2019}\natexlab{}.
\newblock \showarticletitle{Machine {Learning} {Interpretability}: A {Survey} on {Methods} and {Metrics}}.
\newblock \bibinfo{journal}{\emph{Electronics}} \bibinfo{volume}{8}, \bibinfo{number}{8} (\bibinfo{year}{2019}), \bibinfo{pages}{832}.
\newblock


\bibitem[Chang et~al\mbox{.}(2023)]%
        {chang2023knowledge}
\bibfield{author}{\bibinfo{person}{Heng Chang}, \bibinfo{person}{Jie Cai}, {and} \bibinfo{person}{Jia Li}.} \bibinfo{year}{2023}\natexlab{}.
\newblock \showarticletitle{Knowledge Graph Completion with Counterfactual Augmentation}. In \bibinfo{booktitle}{\emph{Proceedings of the {ACM} Web Conference, {WWW}}}. \bibinfo{publisher}{{ACM}}, \bibinfo{pages}{2611--2620}.
\newblock
\urldef\tempurl%
\url{https://doi.org/10.1145/3543507.3583401}
\showDOI{\tempurl}


\bibitem[Chen(2018)]%
        {chen2018fair}
\bibfield{author}{\bibinfo{person}{Jiahao Chen}.} \bibinfo{year}{2018}\natexlab{}.
\newblock \showarticletitle{Fair lending needs explainable models for responsible recommendation}.
\newblock \bibinfo{journal}{\emph{arXiv}} (\bibinfo{year}{2018}).
\newblock


\bibitem[Chen et~al\mbox{.}(2023)]%
        {10.1145/3564284}
\bibfield{author}{\bibinfo{person}{Jiawei Chen}, \bibinfo{person}{Hande Dong}, \bibinfo{person}{Xiang Wang}, \bibinfo{person}{Fuli Feng}, \bibinfo{person}{Meng Wang}, {and} \bibinfo{person}{Xiangnan He}.} \bibinfo{year}{2023}\natexlab{}.
\newblock \showarticletitle{Bias and Debias in Recommender System: A Survey and Future Directions}.
\newblock \bibinfo{journal}{\emph{ACM Trans. Inf. Syst.}} \bibinfo{volume}{41}, \bibinfo{number}{3}, Article \bibinfo{articleno}{67} (\bibinfo{date}{feb} \bibinfo{year}{2023}), \bibinfo{numpages}{39}~pages.
\newblock
\showISSN{1046-8188}
\urldef\tempurl%
\url{https://doi.org/10.1145/3564284}
\showDOI{\tempurl}


\bibitem[Chen et~al\mbox{.}(2020b)]%
        {chen2020simple}
\bibfield{author}{\bibinfo{person}{Ming Chen}, \bibinfo{person}{Zhewei Wei}, \bibinfo{person}{Zengfeng Huang}, \bibinfo{person}{Bolin Ding}, {and} \bibinfo{person}{Yaliang Li}.} \bibinfo{year}{2020}\natexlab{b}.
\newblock \showarticletitle{Simple and {Deep} {Graph} {Convolutional} {Networks}.}. In \bibinfo{booktitle}{\emph{International {Conference} on {Machine} {Learning} ({ICML})}}. \bibinfo{pages}{1725--1735}.
\newblock


\bibitem[Chen et~al\mbox{.}(2022)]%
        {chen2022grease}
\bibfield{author}{\bibinfo{person}{Ziheng Chen}, \bibinfo{person}{F. Silvestri}, \bibinfo{person}{Jia Wang}, \bibinfo{person}{Yongfeng Zhang}, \bibinfo{person}{Zhenhua Huang}, \bibinfo{person}{H. Ahn}, {and} \bibinfo{person}{Gabriele Tolomei}.} \bibinfo{year}{2022}\natexlab{}.
\newblock \showarticletitle{GREASE: Generate {Factual} and {Counterfactual} {Explanations} for {GNN}-based {Recommendations}}.
\newblock \bibinfo{journal}{\emph{ArXiv}}  \bibinfo{volume}{abs/2208.04222} (\bibinfo{year}{2022}).
\newblock


\bibitem[Chen et~al\mbox{.}(2020a)]%
        {chen2020knowledge}
\bibfield{author}{\bibinfo{person}{Zhe Chen}, \bibinfo{person}{Yuehan Wang}, \bibinfo{person}{Bin Zhao}, \bibinfo{person}{Jing Cheng}, \bibinfo{person}{Xin Zhao}, {and} \bibinfo{person}{Zongtao Duan}.} \bibinfo{year}{2020}\natexlab{a}.
\newblock \showarticletitle{Knowledge graph completion: A review}.
\newblock \bibinfo{journal}{\emph{Ieee Access}}  \bibinfo{volume}{8} (\bibinfo{year}{2020}), \bibinfo{pages}{192435--192456}.
\newblock


\bibitem[Cheng et~al\mbox{.}(2021)]%
        {cheng2021causal}
\bibfield{author}{\bibinfo{person}{Lu Cheng}, \bibinfo{person}{Ahmadreza Mosallanezhad}, \bibinfo{person}{Paras Sheth}, {and} \bibinfo{person}{Huan Liu}.} \bibinfo{year}{2021}\natexlab{}.
\newblock \showarticletitle{Causal Learning for Socially Responsible AI}.
\newblock \bibinfo{journal}{\emph{arXiv preprint arXiv:2104.12278}} (\bibinfo{year}{2021}).
\newblock


\bibitem[Chhablani et~al\mbox{.}(2024)]%
        {chhablani2024game}
\bibfield{author}{\bibinfo{person}{Chirag Chhablani}, \bibinfo{person}{Sarthak Jain}, \bibinfo{person}{Akshay Channesh}, \bibinfo{person}{Ian~A Kash}, {and} \bibinfo{person}{Sourav Medya}.} \bibinfo{year}{2024}\natexlab{}.
\newblock \showarticletitle{Game-theoretic Counterfactual Explanation for Graph Neural Networks}.
\newblock \bibinfo{journal}{\emph{arXiv preprint arXiv:2402.06030}} (\bibinfo{year}{2024}).
\newblock


\bibitem[Coley et~al\mbox{.}(2019)]%
        {coley2019a}
\bibfield{author}{\bibinfo{person}{Connor~W Coley}, \bibinfo{person}{Wengong Jin}, \bibinfo{person}{Luke Rogers}, \bibinfo{person}{Timothy~F Jamison}, \bibinfo{person}{Tommi~S Jaakkola}, \bibinfo{person}{William~H Green}, \bibinfo{person}{Regina Barzilay}, {and} \bibinfo{person}{Klavs~F Jensen}.} \bibinfo{year}{2019}\natexlab{}.
\newblock \showarticletitle{A graph-convolutional neural network model for the prediction of chemical reactivity}.
\newblock \bibinfo{journal}{\emph{Chemical science}} \bibinfo{volume}{10}, \bibinfo{number}{2} (\bibinfo{year}{2019}), \bibinfo{pages}{370--377}.
\newblock


\bibitem[Consortium and {others}(2013)]%
        {consortium2013identification}
\bibfield{author}{\bibinfo{person}{Cross-Disorder Group Psychiatric~Genomics Consortium} {and} \bibinfo{person}{{others}}.} \bibinfo{year}{2013}\natexlab{}.
\newblock \showarticletitle{Identification of risk loci with shared effects on five major psychiatric disorders: a genome-wide analysis}.
\newblock \bibinfo{journal}{\emph{The Lancet}} \bibinfo{volume}{381}, \bibinfo{number}{9875} (\bibinfo{year}{2013}), \bibinfo{pages}{1371--1379}.
\newblock


\bibitem[Craddock et~al\mbox{.}(2013)]%
        {craddock2013the}
\bibfield{author}{\bibinfo{person}{Cameron Craddock}, \bibinfo{person}{Yassine Benhajali}, \bibinfo{person}{Carlton Chu}, \bibinfo{person}{Francois Chouinard}, \bibinfo{person}{Alan Evans}, \bibinfo{person}{Andr{\' a}s Jakab}, \bibinfo{person}{Budhachandra~Singh Khundrakpam}, \bibinfo{person}{John~David Lewis}, \bibinfo{person}{Qingyang Li}, \bibinfo{person}{Michael Milham}, {and} \bibinfo{person}{{others}}.} \bibinfo{year}{2013}\natexlab{}.
\newblock \showarticletitle{The neuro bureau preprocessing initiative: open sharing of preprocessed neuroimaging data and derivatives}.
\newblock \bibinfo{journal}{\emph{Frontiers in Neuroinformatics}}  \bibinfo{volume}{7} (\bibinfo{year}{2013}), \bibinfo{pages}{27}.
\newblock


\bibitem[Dai et~al\mbox{.}(2021)]%
        {dai2021nrgnn}
\bibfield{author}{\bibinfo{person}{Enyan Dai}, \bibinfo{person}{Charu Aggarwal}, {and} \bibinfo{person}{Suhang Wang}.} \bibinfo{year}{2021}\natexlab{}.
\newblock \showarticletitle{Nrgnn: Learning a label noise resistant graph neural network on sparsely and noisily labeled graphs}. In \bibinfo{booktitle}{\emph{Proceedings of the 27th ACM SIGKDD Conference on Knowledge Discovery \& Data Mining}}. \bibinfo{pages}{227--236}.
\newblock


\bibitem[Dai and Wang(2021a)]%
        {dai2021say}
\bibfield{author}{\bibinfo{person}{Enyan Dai} {and} \bibinfo{person}{Suhang Wang}.} \bibinfo{year}{2021}\natexlab{a}.
\newblock \showarticletitle{Say {No} to the {Discrimination}: Learning {Fair} {Graph} {Neural} {Networks} with {Limited} {Sensitive} {Attribute} {Information}}.
\newblock  (\bibinfo{year}{2021}).
\newblock


\bibitem[Dai and Wang(2021b)]%
        {dai2021towards}
\bibfield{author}{\bibinfo{person}{Enyan Dai} {and} \bibinfo{person}{Suhang Wang}.} \bibinfo{year}{2021}\natexlab{b}.
\newblock \showarticletitle{Towards self-explainable graph neural network}. In \bibinfo{booktitle}{\emph{Proceedings of the 30th {ACM} {International} {Conference} on {Information} \& {Knowledge} {Management}}}. \bibinfo{pages}{302--311}.
\newblock


\bibitem[Dai and Wang(2022a)]%
        {dai2022learning}
\bibfield{author}{\bibinfo{person}{Enyan Dai} {and} \bibinfo{person}{Suhang Wang}.} \bibinfo{year}{2022}\natexlab{a}.
\newblock \showarticletitle{Learning fair graph neural networks with limited and private sensitive attribute information}.
\newblock \bibinfo{journal}{\emph{IEEE Transactions on Knowledge and Data Engineering}} (\bibinfo{year}{2022}).
\newblock


\bibitem[Dai and Wang(2022b)]%
        {dai2022towards}
\bibfield{author}{\bibinfo{person}{Enyan Dai} {and} \bibinfo{person}{Suhang Wang}.} \bibinfo{year}{2022}\natexlab{b}.
\newblock \showarticletitle{Towards Prototype-Based Self-Explainable Graph Neural Network}.
\newblock \bibinfo{journal}{\emph{arXiv preprint arXiv:2210.01974}} (\bibinfo{year}{2022}).
\newblock


\bibitem[Dai et~al\mbox{.}(2022)]%
        {dai2022a}
\bibfield{author}{\bibinfo{person}{Enyan Dai}, \bibinfo{person}{Tianxiang Zhao}, \bibinfo{person}{Huaisheng Zhu}, \bibinfo{person}{Junjie Xu}, \bibinfo{person}{Zhimeng Guo}, \bibinfo{person}{Hui Liu}, \bibinfo{person}{Jiliang Tang}, {and} \bibinfo{person}{Suhang Wang}.} \bibinfo{year}{2022}\natexlab{}.
\newblock \showarticletitle{A {Comprehensive} {Survey} on {Trustworthy} {Graph} {Neural} {Networks}: Privacy, {Robustness}, {Fairness}, and {Explainability}}.
\newblock \bibinfo{journal}{\emph{arXiv}} (\bibinfo{year}{2022}).
\newblock


\bibitem[Debnath et~al\mbox{.}(1991)]%
        {debnath1991structureactivity}
\bibfield{author}{\bibinfo{person}{Asim~Kumar Debnath}, \bibinfo{person}{Rosa~L Compadre}, \bibinfo{person}{Gargi Debnath}, \bibinfo{person}{Alan~J Shusterman}, {and} \bibinfo{person}{Corwin Hansch}.} \bibinfo{year}{1991}\natexlab{}.
\newblock \showarticletitle{Structure-activity relationship of mutagenic aromatic and heteroaromatic nitro compounds. correlation with molecular orbital energies and hydrophobicity}.
\newblock \bibinfo{journal}{\emph{Journal of medicinal chemistry}} \bibinfo{volume}{34}, \bibinfo{number}{2} (\bibinfo{year}{1991}), \bibinfo{pages}{786--797}.
\newblock


\bibitem[Dehejia and Wahba(1999)]%
        {dehejia1999causal}
\bibfield{author}{\bibinfo{person}{Rajeev~H Dehejia} {and} \bibinfo{person}{Sadek Wahba}.} \bibinfo{year}{1999}\natexlab{}.
\newblock \showarticletitle{Causal effects in nonexperimental studies: Reevaluating the evaluation of training programs}.
\newblock \bibinfo{journal}{\emph{Journal of the American statistical Association}} \bibinfo{volume}{94}, \bibinfo{number}{448} (\bibinfo{year}{1999}), \bibinfo{pages}{1053--1062}.
\newblock


\bibitem[Deng et~al\mbox{.}(2009)]%
        {deng2009imagenet}
\bibfield{author}{\bibinfo{person}{Jia Deng}, \bibinfo{person}{Wei Dong}, \bibinfo{person}{Richard Socher}, \bibinfo{person}{Li-Jia Li}, \bibinfo{person}{Kai Li}, {and} \bibinfo{person}{Li Fei-Fei}.} \bibinfo{year}{2009}\natexlab{}.
\newblock \showarticletitle{Imagenet: A large-scale hierarchical image database}. In \bibinfo{booktitle}{\emph{2009 {IEEE} conference on computer vision and pattern recognition}}. Ieee, \bibinfo{pages}{248--255}.
\newblock


\bibitem[Devlin et~al\mbox{.}(2019)]%
        {devlin2019bert}
\bibfield{author}{\bibinfo{person}{Jacob Devlin}, \bibinfo{person}{Ming-Wei Chang}, \bibinfo{person}{Kenton Lee}, {and} \bibinfo{person}{Kristina Toutanova}.} \bibinfo{year}{2019}\natexlab{}.
\newblock \showarticletitle{BERT: Pre-training of {Deep} {Bidirectional} {Transformers} for {Language} {Understanding}.}. In \bibinfo{booktitle}{\emph{North {American} {Chapter} of the {Association} for {Computational} {Linguistics} ({NAACL})}}. \bibinfo{pages}{4171--4186}.
\newblock


\bibitem[Ding et~al\mbox{.}(2022)]%
        {ding2022data}
\bibfield{author}{\bibinfo{person}{Kaize Ding}, \bibinfo{person}{Zhe Xu}, \bibinfo{person}{Hanghang Tong}, {and} \bibinfo{person}{Huan Liu}.} \bibinfo{year}{2022}\natexlab{}.
\newblock \showarticletitle{Data augmentation for deep graph learning: A survey}.
\newblock \bibinfo{journal}{\emph{arXiv}} (\bibinfo{year}{2022}).
\newblock


\bibitem[Dong et~al\mbox{.}(2022a)]%
        {dong2022edits}
\bibfield{author}{\bibinfo{person}{Yushun Dong}, \bibinfo{person}{Ninghao Liu}, \bibinfo{person}{Brian Jalaian}, {and} \bibinfo{person}{Jundong Li}.} \bibinfo{year}{2022}\natexlab{a}.
\newblock \showarticletitle{Edits: Modeling and mitigating data bias for graph neural networks}. In \bibinfo{booktitle}{\emph{Proceedings of the ACM web conference, {WWW} 2022}}. \bibinfo{pages}{1259--1269}.
\newblock


\bibitem[Dong et~al\mbox{.}(2022b)]%
        {dong2022fairness}
\bibfield{author}{\bibinfo{person}{Yushun Dong}, \bibinfo{person}{Jing Ma}, \bibinfo{person}{Chen Chen}, {and} \bibinfo{person}{Jundong Li}.} \bibinfo{year}{2022}\natexlab{b}.
\newblock \showarticletitle{Fairness in {Graph} {Mining}: A {Survey}}.
\newblock \bibinfo{journal}{\emph{arXiv}} (\bibinfo{year}{2022}).
\newblock


\bibitem[Dorri et~al\mbox{.}(2018)]%
        {dorri2018multiagent}
\bibfield{author}{\bibinfo{person}{Ali Dorri}, \bibinfo{person}{Salil~S Kanhere}, {and} \bibinfo{person}{Raja Jurdak}.} \bibinfo{year}{2018}\natexlab{}.
\newblock \showarticletitle{Multi-agent systems: A survey}.
\newblock \bibinfo{journal}{\emph{Ieee Access}}  \bibinfo{volume}{6} (\bibinfo{year}{2018}), \bibinfo{pages}{28573--28593}.
\newblock


\bibitem[Dubey and Shapley(1979)]%
        {dubey1979mathematical}
\bibfield{author}{\bibinfo{person}{Pradeep Dubey} {and} \bibinfo{person}{Lloyd~S Shapley}.} \bibinfo{year}{1979}\natexlab{}.
\newblock \showarticletitle{Mathematical properties of the Banzhaf power index}.
\newblock \bibinfo{journal}{\emph{Mathematics of Operations Research}}  \bibinfo{volume}{4} (\bibinfo{year}{1979}), \bibinfo{pages}{99--131}.
\newblock


\bibitem[Dwork et~al\mbox{.}(2012)]%
        {dwork2012fairness}
\bibfield{author}{\bibinfo{person}{Cynthia Dwork}, \bibinfo{person}{Moritz Hardt}, \bibinfo{person}{Toniann Pitassi}, \bibinfo{person}{Omer Reingold}, {and} \bibinfo{person}{Richard Zemel}.} \bibinfo{year}{2012}\natexlab{}.
\newblock \showarticletitle{Fairness through awareness}. In \bibinfo{booktitle}{\emph{Proceedings of the 3rd innovations in theoretical computer science conference}}. \bibinfo{pages}{214--226}.
\newblock


\bibitem[Fan et~al\mbox{.}(2021)]%
        {fan2021interpretability}
\bibfield{author}{\bibinfo{person}{Feng-Lei Fan}, \bibinfo{person}{Jinjun Xiong}, \bibinfo{person}{Mengzhou Li}, {and} \bibinfo{person}{Ge Wang}.} \bibinfo{year}{2021}\natexlab{}.
\newblock \showarticletitle{On interpretability of artificial neural networks: A survey}.
\newblock \bibinfo{journal}{\emph{IEEE Transactions on Radiation and Plasma Medical Sciences}} \bibinfo{volume}{5}, \bibinfo{number}{6} (\bibinfo{year}{2021}), \bibinfo{pages}{741--760}.
\newblock


\bibitem[Fan et~al\mbox{.}(2019)]%
        {fan2019graph}
\bibfield{author}{\bibinfo{person}{Wenqi Fan}, \bibinfo{person}{Yao Ma}, \bibinfo{person}{Qing Li}, \bibinfo{person}{Yuan He}, \bibinfo{person}{Eric Zhao}, \bibinfo{person}{Jiliang Tang}, {and} \bibinfo{person}{Dawei Yin}.} \bibinfo{year}{2019}\natexlab{}.
\newblock \showarticletitle{Graph neural networks for social recommendation}. In \bibinfo{booktitle}{\emph{The world wide web conference}}. \bibinfo{pages}{417--426}.
\newblock


\bibitem[Feder et~al\mbox{.}(2021)]%
        {feder2021causal}
\bibfield{author}{\bibinfo{person}{Amir Feder}, \bibinfo{person}{Katherine~A Keith}, \bibinfo{person}{Emaad Manzoor}, \bibinfo{person}{Reid Pryzant}, \bibinfo{person}{Dhanya Sridhar}, \bibinfo{person}{Zach Wood-Doughty}, \bibinfo{person}{Jacob Eisenstein}, \bibinfo{person}{Justin Grimmer}, \bibinfo{person}{Roi Reichart}, \bibinfo{person}{Margaret~E Roberts}, {et~al\mbox{.}}} \bibinfo{year}{2021}\natexlab{}.
\newblock \showarticletitle{Causal inference in natural language processing: Estimation, prediction, interpretation and beyond}.
\newblock \bibinfo{journal}{\emph{arXiv preprint arXiv:2109.00725}} (\bibinfo{year}{2021}).
\newblock


\bibitem[Feng et~al\mbox{.}(2015)]%
        {feng2015personalized}
\bibfield{author}{\bibinfo{person}{Shanshan Feng}, \bibinfo{person}{Xutao Li}, \bibinfo{person}{Yifeng Zeng}, \bibinfo{person}{Gao Cong}, \bibinfo{person}{Yeow~Meng Chee}, {and} \bibinfo{person}{Quan Yuan}.} \bibinfo{year}{2015}\natexlab{}.
\newblock \showarticletitle{Personalized {Ranking} {Metric} {Embedding} for {Next} {New} {POI} {Recommendation}.}. In \bibinfo{booktitle}{\emph{International {Joint} {Conference} on {Artificial} {Intelligence} ({IJCAI})}}. ACM, \bibinfo{pages}{2069--2075}.
\newblock


\bibitem[Foerster et~al\mbox{.}(2018)]%
        {foerster2018counterfactual}
\bibfield{author}{\bibinfo{person}{Jakob~N. Foerster}, \bibinfo{person}{Gregory Farquhar}, \bibinfo{person}{Triantafyllos Afouras}, \bibinfo{person}{Nantas Nardelli}, {and} \bibinfo{person}{Shimon Whiteson}.} \bibinfo{year}{2018}\natexlab{}.
\newblock \showarticletitle{Counterfactual {Multi}-{Agent} {Policy} {Gradients}.}. In \bibinfo{booktitle}{\emph{AAAI {Conference} on {Artificial} {Intelligence} ({AAAI})}}. \bibinfo{pages}{2974--2982}.
\newblock


\bibitem[Frasca et~al\mbox{.}(2020)]%
        {frasca2020sign}
\bibfield{author}{\bibinfo{person}{Fabrizio Frasca}, \bibinfo{person}{Emanuele Rossi}, \bibinfo{person}{Davide Eynard}, \bibinfo{person}{Ben Chamberlain}, \bibinfo{person}{Michael Bronstein}, {and} \bibinfo{person}{Federico Monti}.} \bibinfo{year}{2020}\natexlab{}.
\newblock \showarticletitle{Sign: Scalable inception graph neural networks}.
\newblock \bibinfo{journal}{\emph{arXiv}} (\bibinfo{year}{2020}).
\newblock


\bibitem[Fujii et~al\mbox{.}(2022)]%
        {fujii2022estimating}
\bibfield{author}{\bibinfo{person}{Keisuke Fujii}, \bibinfo{person}{Koh Takeuchi}, \bibinfo{person}{Atsushi Kuribayashi}, \bibinfo{person}{Naoya Takeishi}, \bibinfo{person}{Yoshinobu Kawahara}, {and} \bibinfo{person}{Kazuya Takeda}.} \bibinfo{year}{2022}\natexlab{}.
\newblock \showarticletitle{Estimating counterfactual treatment outcomes over time in complex multi-agent scenarios}.
\newblock \bibinfo{journal}{\emph{arXiv preprint arXiv:2206.01900}} (\bibinfo{year}{2022}).
\newblock


\bibitem[Gao et~al\mbox{.}(2022)]%
        {gao2022causal}
\bibfield{author}{\bibinfo{person}{Chen Gao}, \bibinfo{person}{Yu Zheng}, \bibinfo{person}{Wenjie Wang}, \bibinfo{person}{Fuli Feng}, \bibinfo{person}{Xiangnan He}, {and} \bibinfo{person}{Yong Li}.} \bibinfo{year}{2022}\natexlab{}.
\newblock \showarticletitle{Causal inference in recommender systems: A survey and future directions}.
\newblock \bibinfo{journal}{\emph{arXiv preprint arXiv:2208.12397}} (\bibinfo{year}{2022}).
\newblock


\bibitem[Garg et~al\mbox{.}(2019)]%
        {garg2019counterfactual}
\bibfield{author}{\bibinfo{person}{Sahaj Garg}, \bibinfo{person}{Vincent Perot}, \bibinfo{person}{Nicole Limtiaco}, \bibinfo{person}{Ankur Taly}, \bibinfo{person}{Ed~H Chi}, {and} \bibinfo{person}{Alex Beutel}.} \bibinfo{year}{2019}\natexlab{}.
\newblock \showarticletitle{Counterfactual fairness in text classification through robustness}. In \bibinfo{booktitle}{\emph{Proceedings of the 2019 {AAAI}/{ACM} {Conference} on {AI}, {Ethics}, and {Society}}}. \bibinfo{pages}{219--226}.
\newblock


\bibitem[Glymour et~al\mbox{.}(2016)]%
        {glymour2016causal}
\bibfield{author}{\bibinfo{person}{Madelyn Glymour}, \bibinfo{person}{Judea Pearl}, {and} \bibinfo{person}{Nicholas~P Jewell}.} \bibinfo{year}{2016}\natexlab{}.
\newblock \bibinfo{booktitle}{\emph{Causal inference in statistics: A primer}}.
\newblock \bibinfo{publisher}{John Wiley \& Sons}.
\newblock


\bibitem[Grover and Leskovec(2016)]%
        {grover2016nodevec}
\bibfield{author}{\bibinfo{person}{Aditya Grover} {and} \bibinfo{person}{Jure Leskovec}.} \bibinfo{year}{2016}\natexlab{}.
\newblock \showarticletitle{node2vec - {Scalable} {Feature} {Learning} for {Networks}}. In \bibinfo{booktitle}{\emph{Proceedings of the 22nd {ACM} {SIGKDD} {International} {Conference} on {Knowledge} {Discovery} and {Data} {Mining}}}. ACM.
\newblock


\bibitem[Gulzar et~al\mbox{.}(2021)]%
        {gulzar2021a}
\bibfield{author}{\bibinfo{person}{Mahir Gulzar}, \bibinfo{person}{Yar Muhammad}, {and} \bibinfo{person}{Naveed Muhammad}.} \bibinfo{year}{2021}\natexlab{}.
\newblock \showarticletitle{A survey on motion prediction of pedestrians and vehicles for autonomous driving}.
\newblock \bibinfo{journal}{\emph{IEEE Access}}  \bibinfo{volume}{9} (\bibinfo{year}{2021}), \bibinfo{pages}{137957--137969}.
\newblock


\bibitem[Guo et~al\mbox{.}(2020a)]%
        {guo2020a}
\bibfield{author}{\bibinfo{person}{Ruocheng Guo}, \bibinfo{person}{Lu Cheng}, \bibinfo{person}{Jundong Li}, \bibinfo{person}{P.~Richard Hahn}, {and} \bibinfo{person}{Huan Liu}.} \bibinfo{year}{2020}\natexlab{a}.
\newblock \showarticletitle{A {Survey} of {Learning} {Causality} with {Data}: Problems and {Methods}.}
\newblock \bibinfo{journal}{\emph{Comput. Surveys}} \bibinfo{volume}{53}, \bibinfo{number}{4} (\bibinfo{year}{2020}), \bibinfo{pages}{75:1--75:37}.
\newblock


\bibitem[Guo et~al\mbox{.}(2020b)]%
        {guo2020survey}
\bibfield{author}{\bibinfo{person}{Ruocheng Guo}, \bibinfo{person}{Lu Cheng}, \bibinfo{person}{Jundong Li}, \bibinfo{person}{P~Richard Hahn}, {and} \bibinfo{person}{Huan Liu}.} \bibinfo{year}{2020}\natexlab{b}.
\newblock \showarticletitle{A survey of learning causality with data: Problems and methods}.
\newblock \bibinfo{journal}{\emph{ACM Computing Surveys (CSUR)}} \bibinfo{volume}{53}, \bibinfo{number}{4} (\bibinfo{year}{2020}), \bibinfo{pages}{1--37}.
\newblock


\bibitem[Guo et~al\mbox{.}(2023)]%
        {Guo0X0W23}
\bibfield{author}{\bibinfo{person}{Zhimeng Guo}, \bibinfo{person}{Jialiang Li}, \bibinfo{person}{Teng Xiao}, \bibinfo{person}{Yao Ma}, {and} \bibinfo{person}{Suhang Wang}.} \bibinfo{year}{2023}\natexlab{}.
\newblock \showarticletitle{Towards Fair Graph Neural Networks via Graph Counterfactual}. In \bibinfo{booktitle}{\emph{Proceedings of the 32nd {ACM} International Conference on Information and Knowledge Management, {CIKM} 2023}}. \bibinfo{pages}{669--678}.
\newblock
\urldef\tempurl%
\url{https://doi.org/10.1145/3583780.3615092}
\showDOI{\tempurl}


\bibitem[Hamilton(2020)]%
        {hamilton2020graph}
\bibfield{author}{\bibinfo{person}{William~L Hamilton}.} \bibinfo{year}{2020}\natexlab{}.
\newblock \showarticletitle{Graph representation learning}.
\newblock \bibinfo{journal}{\emph{Synthesis Lectures on Artifical Intelligence and Machine Learning}} \bibinfo{volume}{14}, \bibinfo{number}{3} (\bibinfo{year}{2020}), \bibinfo{pages}{1--159}.
\newblock


\bibitem[Hardt et~al\mbox{.}(2016)]%
        {hardt2016equality}
\bibfield{author}{\bibinfo{person}{Moritz Hardt}, \bibinfo{person}{Eric Price}, {and} \bibinfo{person}{Nati Srebro}.} \bibinfo{year}{2016}\natexlab{}.
\newblock \showarticletitle{Equality of {Opportunity} in {Supervised} {Learning}.}. In \bibinfo{booktitle}{\emph{Conference on {Neural} {Information} {Processing} {Systems} ({NeurIPS})}}. \bibinfo{pages}{3315--3323}.
\newblock


\bibitem[Hassani and Ahmadi(2020)]%
        {hassani2020contrastive}
\bibfield{author}{\bibinfo{person}{Kaveh Hassani} {and} \bibinfo{person}{Amir Hosein~Khas Ahmadi}.} \bibinfo{year}{2020}\natexlab{}.
\newblock \showarticletitle{Contrastive {Multi}-{View} {Representation} {Learning} on {Graphs}.}. In \bibinfo{booktitle}{\emph{International Conference on Machine Learning, {ICML}}}. PMLR, \bibinfo{pages}{4116--4126}.
\newblock


\bibitem[He et~al\mbox{.}(2016)]%
        {he2016deep}
\bibfield{author}{\bibinfo{person}{Kaiming He}, \bibinfo{person}{Xiangyu Zhang}, \bibinfo{person}{Shaoqing Ren}, {and} \bibinfo{person}{Jian Sun}.} \bibinfo{year}{2016}\natexlab{}.
\newblock \showarticletitle{Deep {Residual} {Learning} for {Image} {Recognition}.}. In \bibinfo{booktitle}{\emph{Computer {Vision} and {Pattern} {Recognition} ({CVPR})}}. \bibinfo{pages}{770--778}.
\newblock


\bibitem[Holzinger et~al\mbox{.}(2021)]%
        {holzinger2021towards}
\bibfield{author}{\bibinfo{person}{Andreas Holzinger}, \bibinfo{person}{Bernd Malle}, \bibinfo{person}{Anna Saranti}, {and} \bibinfo{person}{Bastian Pfeifer}.} \bibinfo{year}{2021}\natexlab{}.
\newblock \showarticletitle{Towards multi-modal causability with graph neural networks enabling information fusion for explainable AI}.
\newblock \bibinfo{journal}{\emph{Information Fusion}}  \bibinfo{volume}{71} (\bibinfo{year}{2021}), \bibinfo{pages}{28--37}.
\newblock


\bibitem[Huang et~al\mbox{.}(2022)]%
        {huang2022graphlime}
\bibfield{author}{\bibinfo{person}{Qiang Huang}, \bibinfo{person}{Makoto Yamada}, \bibinfo{person}{Yuan Tian}, \bibinfo{person}{Dinesh Singh}, {and} \bibinfo{person}{Yi Chang}.} \bibinfo{year}{2022}\natexlab{}.
\newblock \showarticletitle{Graphlime: Local interpretable model explanations for graph neural networks}.
\newblock \bibinfo{journal}{\emph{IEEE Transactions on Knowledge and Data Engineering}} (\bibinfo{year}{2022}).
\newblock


\bibitem[Huang et~al\mbox{.}(2020)]%
        {huang2020fusion}
\bibfield{author}{\bibinfo{person}{Shih-Cheng Huang}, \bibinfo{person}{Anuj Pareek}, \bibinfo{person}{Saeed Seyyedi}, \bibinfo{person}{Imon Banerjee}, {and} \bibinfo{person}{Matthew~P Lungren}.} \bibinfo{year}{2020}\natexlab{}.
\newblock \showarticletitle{Fusion of medical imaging and electronic health records using deep learning: a systematic review and implementation guidelines}.
\newblock \bibinfo{journal}{\emph{NPJ digital medicine}} \bibinfo{volume}{3}, \bibinfo{number}{1} (\bibinfo{year}{2020}), \bibinfo{pages}{136}.
\newblock


\bibitem[Huang et~al\mbox{.}(2023)]%
        {huang2023global}
\bibfield{author}{\bibinfo{person}{Zexi Huang}, \bibinfo{person}{Mert Kosan}, \bibinfo{person}{Sourav Medya}, \bibinfo{person}{Sayan Ranu}, {and} \bibinfo{person}{Ambuj Singh}.} \bibinfo{year}{2023}\natexlab{}.
\newblock \showarticletitle{Global {Counterfactual} {Explainer} for {Graph} {Neural} {Networks}}. In \bibinfo{booktitle}{\emph{Proceedings of the {Sixteenth} {ACM} {International} {Conference} on {Web} {Search} and {Data} {Mining}}}. ACM.
\newblock


\bibitem[Hvilsh\o{}j et~al\mbox{.}(2021)]%
        {hvilshoj2021ecinn}
\bibfield{author}{\bibinfo{person}{Frederik Hvilsh\o{}j}, \bibinfo{person}{Alexandros Iosifidis}, {and} \bibinfo{person}{Ira Assent}.} \bibinfo{year}{2021}\natexlab{}.
\newblock \showarticletitle{ECINN: Efficient {Counterfactuals} from {Invertible} {Neural} {Networks}.}. In \bibinfo{booktitle}{\emph{British {Machine} {Vision} {Conference} ({BMVC})}}. \bibinfo{pages}{43}.
\newblock


\bibitem[Imbens(2004)]%
        {imbens2004nonparametric}
\bibfield{author}{\bibinfo{person}{Guido~W Imbens}.} \bibinfo{year}{2004}\natexlab{}.
\newblock \showarticletitle{Nonparametric estimation of average treatment effects under exogeneity: A review}.
\newblock \bibinfo{journal}{\emph{Review of Economics and statistics}} \bibinfo{volume}{86}, \bibinfo{number}{1} (\bibinfo{year}{2004}), \bibinfo{pages}{4--29}.
\newblock


\bibitem[Jadidinejad et~al\mbox{.}(2021)]%
        {10.1145/3458509}
\bibfield{author}{\bibinfo{person}{Amir~H. Jadidinejad}, \bibinfo{person}{Craig Macdonald}, {and} \bibinfo{person}{Iadh Ounis}.} \bibinfo{year}{2021}\natexlab{}.
\newblock \showarticletitle{The Simpson’s Paradox in the Offline Evaluation of Recommendation Systems}.
\newblock \bibinfo{journal}{\emph{ACM Trans. Inf. Syst.}} \bibinfo{volume}{40}, \bibinfo{number}{1}, Article \bibinfo{articleno}{4} (\bibinfo{date}{sep} \bibinfo{year}{2021}), \bibinfo{numpages}{22}~pages.
\newblock
\showISSN{1046-8188}
\urldef\tempurl%
\url{https://doi.org/10.1145/3458509}
\showDOI{\tempurl}


\bibitem[Jalal et~al\mbox{.}(2021)]%
        {jalal2021fairness}
\bibfield{author}{\bibinfo{person}{Ajil Jalal}, \bibinfo{person}{Sushrut Karmalkar}, \bibinfo{person}{Jessica Hoffmann}, \bibinfo{person}{Alex Dimakis}, {and} \bibinfo{person}{Eric Price}.} \bibinfo{year}{2021}\natexlab{}.
\newblock \showarticletitle{Fairness for image generation with uncertain sensitive attributes}. In \bibinfo{booktitle}{\emph{International {Conference} on {Machine} {Learning}}}. PMLR, \bibinfo{pages}{4721--4732}.
\newblock


\bibitem[Ji et~al\mbox{.}(2021)]%
        {ji2021a}
\bibfield{author}{\bibinfo{person}{Shaoxiong Ji}, \bibinfo{person}{Shirui Pan}, \bibinfo{person}{Erik Cambria}, \bibinfo{person}{Pekka Marttinen}, {and} \bibinfo{person}{S~Yu Philip}.} \bibinfo{year}{2021}\natexlab{}.
\newblock \showarticletitle{A survey on knowledge graphs: Representation, acquisition, and applications}.
\newblock \bibinfo{journal}{\emph{IEEE transactions on neural networks and learning systems}} \bibinfo{volume}{33}, \bibinfo{number}{2} (\bibinfo{year}{2021}), \bibinfo{pages}{494--514}.
\newblock


\bibitem[Jiang et~al\mbox{.}(2022)]%
        {jiang2022fmp}
\bibfield{author}{\bibinfo{person}{Zhimeng Jiang}, \bibinfo{person}{Xiaotian Han}, \bibinfo{person}{Chao Fan}, \bibinfo{person}{Zirui Liu}, \bibinfo{person}{Na Zou}, \bibinfo{person}{Ali Mostafavi}, {and} \bibinfo{person}{Xia Hu}.} \bibinfo{year}{2022}\natexlab{}.
\newblock \showarticletitle{Fmp: Toward fair graph message passing against topology bias}.
\newblock \bibinfo{journal}{\emph{ArXiv}}  \bibinfo{volume}{abs/2202.04187} (\bibinfo{year}{2022}).
\newblock


\bibitem[Jordan and Freiburger(2015)]%
        {jordan2015the}
\bibfield{author}{\bibinfo{person}{Kareem~L Jordan} {and} \bibinfo{person}{Tina~L Freiburger}.} \bibinfo{year}{2015}\natexlab{}.
\newblock \showarticletitle{The effect of race/ethnicity on sentencing: Examining sentence type, jail length, and prison length}.
\newblock \bibinfo{journal}{\emph{Journal of Ethnicity in Criminal Justice}} \bibinfo{volume}{13}, \bibinfo{number}{3} (\bibinfo{year}{2015}), \bibinfo{pages}{179--196}.
\newblock


\bibitem[Joshi et~al\mbox{.}(2019)]%
        {joshi2019towards}
\bibfield{author}{\bibinfo{person}{Shalmali Joshi}, \bibinfo{person}{Oluwasanmi Koyejo}, \bibinfo{person}{Warut Vijitbenjaronk}, \bibinfo{person}{Been Kim}, {and} \bibinfo{person}{Joydeep Ghosh}.} \bibinfo{year}{2019}\natexlab{}.
\newblock \showarticletitle{Towards realistic individual recourse and actionable explanations in black-box decision making systems}.
\newblock \bibinfo{journal}{\emph{ArXiv}}  \bibinfo{volume}{abs/1907.09615} (\bibinfo{year}{2019}).
\newblock


\bibitem[Kaddour et~al\mbox{.}(2022)]%
        {kaddour2022causal}
\bibfield{author}{\bibinfo{person}{Jean Kaddour}, \bibinfo{person}{Aengus Lynch}, \bibinfo{person}{Qi Liu}, \bibinfo{person}{Matt~J Kusner}, {and} \bibinfo{person}{Ricardo Silva}.} \bibinfo{year}{2022}\natexlab{}.
\newblock \showarticletitle{Causal {Machine} {Learning}: A {Survey} and {Open} {Problems}}.
\newblock \bibinfo{journal}{\emph{arXiv}} (\bibinfo{year}{2022}).
\newblock


\bibitem[Katz(1953)]%
        {katz1953a}
\bibfield{author}{\bibinfo{person}{Leo Katz}.} \bibinfo{year}{1953}\natexlab{}.
\newblock \showarticletitle{A new status index derived from sociometric analysis}.
\newblock \bibinfo{journal}{\emph{Psychometrika}} \bibinfo{volume}{18}, \bibinfo{number}{1} (\bibinfo{year}{1953}), \bibinfo{pages}{39--43}.
\newblock


\bibitem[Kazemi et~al\mbox{.}(2020)]%
        {kazemi2020representation}
\bibfield{author}{\bibinfo{person}{Seyed~Mehran Kazemi}, \bibinfo{person}{Rishab Goel}, \bibinfo{person}{Kshitij Jain}, \bibinfo{person}{Ivan Kobyzev}, \bibinfo{person}{Akshay Sethi}, \bibinfo{person}{Peter Forsyth}, {and} \bibinfo{person}{Pascal Poupart}.} \bibinfo{year}{2020}\natexlab{}.
\newblock \showarticletitle{Representation {Learning} for {Dynamic} {Graphs}: A {Survey}.}
\newblock \bibinfo{journal}{\emph{Journal of Machine Learning Research (JMLR)}} \bibinfo{volume}{21}, \bibinfo{number}{1} (\bibinfo{year}{2020}), \bibinfo{pages}{70:1--70:73}.
\newblock


\bibitem[Kersting et~al\mbox{.}(2016)]%
        {KKMMN2016}
\bibfield{author}{\bibinfo{person}{Kristian Kersting}, \bibinfo{person}{Nils~M. Kriege}, \bibinfo{person}{Christopher Morris}, \bibinfo{person}{Petra Mutzel}, {and} \bibinfo{person}{Marion Neumann}.} \bibinfo{year}{2016}\natexlab{}.
\newblock \bibinfo{title}{Benchmark Data Sets for Graph Kernels}.
\newblock
\newblock
\urldef\tempurl%
\url{http://graphkernels.cs.tu-dortmund.de}
\showURL{%
\tempurl}


\bibitem[Khajehnejad et~al\mbox{.}(2022)]%
        {khajehnejad2022crosswalk}
\bibfield{author}{\bibinfo{person}{Ahmad Khajehnejad}, \bibinfo{person}{Moein Khajehnejad}, \bibinfo{person}{Mahmoudreza Babaei}, \bibinfo{person}{Krishna~P. Gummadi}, \bibinfo{person}{Adrian Weller}, {and} \bibinfo{person}{Baharan Mirzasoleiman}.} \bibinfo{year}{2022}\natexlab{}.
\newblock \showarticletitle{CrossWalk: Fairness-{Enhanced} {Node} {Representation} {Learning}.}. In \bibinfo{booktitle}{\emph{AAAI {Conference} on {Artificial} {Intelligence} ({AAAI})}}. \bibinfo{pages}{11963--11970}.
\newblock


\bibitem[Kim and Oh(2021)]%
        {kim2021how}
\bibfield{author}{\bibinfo{person}{Dongkwan Kim} {and} \bibinfo{person}{Alice Oh}.} \bibinfo{year}{2021}\natexlab{}.
\newblock \showarticletitle{How to {Find} {Your} {Friendly} {Neighborhood}: Graph {Attention} {Design} with {Self}-{Supervision}}.
\newblock \bibinfo{journal}{\emph{dblp://conf/iclr}} (\bibinfo{year}{2021}).
\newblock


\bibitem[Kingma and Welling(2013)]%
        {kingma2013autoencoding}
\bibfield{author}{\bibinfo{person}{Diederik~P Kingma} {and} \bibinfo{person}{Max Welling}.} \bibinfo{year}{2013}\natexlab{}.
\newblock \showarticletitle{Auto-encoding variational bayes}.
\newblock \bibinfo{journal}{\emph{arXiv preprint arXiv:1312.6114}} (\bibinfo{year}{2013}).
\newblock


\bibitem[Kipf and Welling(2016a)]%
        {kipf2016semi}
\bibfield{author}{\bibinfo{person}{Thomas~N Kipf} {and} \bibinfo{person}{Max Welling}.} \bibinfo{year}{2016}\natexlab{a}.
\newblock \showarticletitle{Semi-supervised classification with graph convolutional networks}.
\newblock \bibinfo{journal}{\emph{arXiv preprint arXiv:1609.02907}} (\bibinfo{year}{2016}).
\newblock


\bibitem[Kipf and Welling(2016b)]%
        {kipf2016variational}
\bibfield{author}{\bibinfo{person}{Thomas~N Kipf} {and} \bibinfo{person}{Max Welling}.} \bibinfo{year}{2016}\natexlab{b}.
\newblock \showarticletitle{Variational graph auto-encoders}.
\newblock \bibinfo{journal}{\emph{arXiv preprint arXiv:1611.07308}} (\bibinfo{year}{2016}).
\newblock


\bibitem[Kipf and Welling(2017)]%
        {kipf2017semisupervised}
\bibfield{author}{\bibinfo{person}{Thomas~N. Kipf} {and} \bibinfo{person}{Max Welling}.} \bibinfo{year}{2017}\natexlab{}.
\newblock \showarticletitle{Semi-{Supervised} {Classification} with {Graph} {Convolutional} {Networks}.}. In \bibinfo{booktitle}{\emph{International {Conference} on {Learning} {Representations} ({ICLR})}}.
\newblock


\bibitem[Klicpera et~al\mbox{.}(2019)]%
        {klicpera2019predict}
\bibfield{author}{\bibinfo{person}{Johannes Klicpera}, \bibinfo{person}{Aleksandar Bojchevski}, {and} \bibinfo{person}{Stephan G{\" u}nnemann}.} \bibinfo{year}{2019}\natexlab{}.
\newblock \showarticletitle{Predict then {Propagate}: Graph {Neural} {Networks} meet {Personalized} {PageRank}.}. In \bibinfo{booktitle}{\emph{International {Conference} on {Learning} {Representations} ({ICLR})}}.
\newblock


\bibitem[Koren et~al\mbox{.}(2009)]%
        {koren2009matrix}
\bibfield{author}{\bibinfo{person}{Yehuda Koren}, \bibinfo{person}{Robert Bell}, {and} \bibinfo{person}{Chris Volinsky}.} \bibinfo{year}{2009}\natexlab{}.
\newblock \showarticletitle{Matrix factorization techniques for recommender systems}.
\newblock \bibinfo{journal}{\emph{Computer}} \bibinfo{volume}{42}, \bibinfo{number}{8} (\bibinfo{year}{2009}), \bibinfo{pages}{30--37}.
\newblock


\bibitem[Koren et~al\mbox{.}(2021)]%
        {koren2021advances}
\bibfield{author}{\bibinfo{person}{Yehuda Koren}, \bibinfo{person}{Steffen Rendle}, {and} \bibinfo{person}{Robert Bell}.} \bibinfo{year}{2021}\natexlab{}.
\newblock \showarticletitle{Advances in collaborative filtering}.
\newblock \bibinfo{journal}{\emph{Recommender systems handbook}} (\bibinfo{year}{2021}), \bibinfo{pages}{91--142}.
\newblock


\bibitem[Kose and Shen(2021)]%
        {kose2021fairnessaware}
\bibfield{author}{\bibinfo{person}{{\" O}yk{\" u}~Deniz Kose} {and} \bibinfo{person}{Yanning Shen}.} \bibinfo{year}{2021}\natexlab{}.
\newblock \showarticletitle{Fairness-aware node representation learning}.
\newblock \bibinfo{journal}{\emph{arXiv}} (\bibinfo{year}{2021}).
\newblock


\bibitem[Kumar et~al\mbox{.}(2020)]%
        {kumar2020link}
\bibfield{author}{\bibinfo{person}{Ajay Kumar}, \bibinfo{person}{Shashank~Sheshar Singh}, \bibinfo{person}{Kuldeep Singh}, {and} \bibinfo{person}{Bhaskar Biswas}.} \bibinfo{year}{2020}\natexlab{}.
\newblock \showarticletitle{Link prediction techniques, applications, and performance: A survey}.
\newblock \bibinfo{journal}{\emph{Physica A: Statistical Mechanics and its Applications}}  \bibinfo{volume}{553} (\bibinfo{year}{2020}), \bibinfo{pages}{124289}.
\newblock


\bibitem[Kusner et~al\mbox{.}(2017)]%
        {kusner2017counterfactual}
\bibfield{author}{\bibinfo{person}{Matt~J Kusner}, \bibinfo{person}{Joshua Loftus}, \bibinfo{person}{Chris Russell}, {and} \bibinfo{person}{Ricardo Silva}.} \bibinfo{year}{2017}\natexlab{}.
\newblock \showarticletitle{Counterfactual fairness}.
\newblock \bibinfo{journal}{\emph{Advances in neural information processing systems}}  \bibinfo{volume}{30} (\bibinfo{year}{2017}).
\newblock


\bibitem[Li et~al\mbox{.}(2022a)]%
        {li2022counterfactual}
\bibfield{author}{\bibinfo{person}{Jiarui Li}, \bibinfo{person}{Yukio Horiguchi}, {and} \bibinfo{person}{Tetsuo Sawaragi}.} \bibinfo{year}{2022}\natexlab{a}.
\newblock \showarticletitle{Counterfactual inference to predict causal knowledge graph for relational transfer learning by assimilating expert knowledge--{Relational} feature transfer learning algorithm}.
\newblock \bibinfo{journal}{\emph{Advanced Engineering Informatics}}  \bibinfo{volume}{51} (\bibinfo{year}{2022}), \bibinfo{pages}{101516}.
\newblock


\bibitem[Li et~al\mbox{.}(2021)]%
        {li2021on}
\bibfield{author}{\bibinfo{person}{Peizhao Li}, \bibinfo{person}{Yifei Wang}, \bibinfo{person}{Han Zhao}, \bibinfo{person}{Pengyu Hong}, {and} \bibinfo{person}{Hongfu Liu}.} \bibinfo{year}{2021}\natexlab{}.
\newblock \showarticletitle{ON {DYADIC} {FAIRNESS}: EXPLORING {AND} {MITIGATING} {BIAS} {IN} {GRAPH} {CONNECTIONS}}.
\newblock  (\bibinfo{year}{2021}), \bibinfo{pages}{18}.
\newblock


\bibitem[Li et~al\mbox{.}(2020b)]%
        {li2020hierarchical}
\bibfield{author}{\bibinfo{person}{Xingchen Li}, \bibinfo{person}{Xiang Wang}, \bibinfo{person}{Xiangnan He}, \bibinfo{person}{Long Chen}, \bibinfo{person}{Jun Xiao}, {and} \bibinfo{person}{Tat-Seng Chua}.} \bibinfo{year}{2020}\natexlab{b}.
\newblock \showarticletitle{Hierarchical fashion graph network for personalized outfit recommendation}. In \bibinfo{booktitle}{\emph{Proceedings of the 43rd {International} {ACM} {SIGIR} {Conference} on {Research} and {Development} in {Information} {Retrieval}}}. \bibinfo{pages}{159--168}.
\newblock


\bibitem[Li et~al\mbox{.}(2020a)]%
        {li2020causal}
\bibfield{author}{\bibinfo{person}{Yunzhu Li}, \bibinfo{person}{Antonio Torralba}, \bibinfo{person}{Anima Anandkumar}, \bibinfo{person}{Dieter Fox}, {and} \bibinfo{person}{Animesh Garg}.} \bibinfo{year}{2020}\natexlab{a}.
\newblock \showarticletitle{Causal discovery in physical systems from videos}.
\newblock \bibinfo{journal}{\emph{Advances in Neural Information Processing Systems}}  \bibinfo{volume}{33} (\bibinfo{year}{2020}), \bibinfo{pages}{9180--9192}.
\newblock


\bibitem[Li et~al\mbox{.}(2022b)]%
        {li2022deconfounding}
\bibfield{author}{\bibinfo{person}{Zongzhao Li}, \bibinfo{person}{Xiangyu Zhu}, \bibinfo{person}{Zhen Lei}, {and} \bibinfo{person}{Zhaoxiang Zhang}.} \bibinfo{year}{2022}\natexlab{b}.
\newblock \showarticletitle{Deconfounding Physical Dynamics with Global Causal Relation and Confounder Transmission for Counterfactual Prediction}.
\newblock  (\bibinfo{year}{2022}).
\newblock


\bibitem[Liben-Nowell and Kleinberg(2007)]%
        {liben2007link}
\bibfield{author}{\bibinfo{person}{David Liben-Nowell} {and} \bibinfo{person}{Jon Kleinberg}.} \bibinfo{year}{2007}\natexlab{}.
\newblock \showarticletitle{The link-prediction problem for social networks}.
\newblock \bibinfo{journal}{\emph{Journal of the American society for information science and technology}} \bibinfo{volume}{58}, \bibinfo{number}{7} (\bibinfo{year}{2007}), \bibinfo{pages}{1019--1031}.
\newblock


\bibitem[Lim et~al\mbox{.}(2020)]%
        {lim2020stpudgat}
\bibfield{author}{\bibinfo{person}{Nicholas Lim}, \bibinfo{person}{Bryan Hooi}, \bibinfo{person}{See-Kiong Ng}, \bibinfo{person}{Xueou Wang}, \bibinfo{person}{Yong~Liang Goh}, \bibinfo{person}{Renrong Weng}, {and} \bibinfo{person}{Jagannadan Varadarajan}.} \bibinfo{year}{2020}\natexlab{}.
\newblock \showarticletitle{STP-{UDGAT}: Spatial-temporal-preference user dimensional graph attention network for next {POI} recommendation}. In \bibinfo{booktitle}{\emph{Proceedings of the 29th {ACM} {International} {Conference} on {Information} \& {Knowledge} {Management}}}. \bibinfo{pages}{845--854}.
\newblock


\bibitem[Limeros et~al\mbox{.}(2022)]%
        {limeros2022towards}
\bibfield{author}{\bibinfo{person}{Sandra~Carrasco Limeros}, \bibinfo{person}{Sylwia Majchrowska}, \bibinfo{person}{Joakim Johnander}, \bibinfo{person}{Christoffer Petersson}, {and} \bibinfo{person}{D.~F. Llorca}.} \bibinfo{year}{2022}\natexlab{}.
\newblock \showarticletitle{Towards {Explainable} {Motion} {Prediction} using {Heterogeneous} {Graph} {Representations}}.
\newblock \bibinfo{journal}{\emph{ArXiv}}  \bibinfo{volume}{abs/2212.03806} (\bibinfo{year}{2022}).
\newblock


\bibitem[Lin et~al\mbox{.}(2021)]%
        {lin2021generative}
\bibfield{author}{\bibinfo{person}{Wanyu Lin}, \bibinfo{person}{Hao Lan}, {and} \bibinfo{person}{Baochun Li}.} \bibinfo{year}{2021}\natexlab{}.
\newblock \showarticletitle{Generative causal explanations for graph neural networks}. In \bibinfo{booktitle}{\emph{International {Conference} on {Machine} {Learning}}}. PMLR, \bibinfo{pages}{6666--6679}.
\newblock


\bibitem[Linden et~al\mbox{.}(2003)]%
        {linden2003amazon}
\bibfield{author}{\bibinfo{person}{Greg Linden}, \bibinfo{person}{Brent Smith}, {and} \bibinfo{person}{Jeremy York}.} \bibinfo{year}{2003}\natexlab{}.
\newblock \showarticletitle{Amazon. com recommendations: Item-to-item collaborative filtering}.
\newblock \bibinfo{journal}{\emph{IEEE Internet computing}} \bibinfo{volume}{7}, \bibinfo{number}{1} (\bibinfo{year}{2003}), \bibinfo{pages}{76--80}.
\newblock


\bibitem[Liu et~al\mbox{.}(2021b)]%
        {liu2021improving}
\bibfield{author}{\bibinfo{person}{Chang Liu}, \bibinfo{person}{Chen Gao}, \bibinfo{person}{Depeng Jin}, {and} \bibinfo{person}{Yong Li}.} \bibinfo{year}{2021}\natexlab{b}.
\newblock \showarticletitle{Improving {Location} {Recommendation} with {Urban} {Knowledge} {Graph}}.
\newblock \bibinfo{journal}{\emph{arXiv}} (\bibinfo{year}{2021}).
\newblock


\bibitem[Liu et~al\mbox{.}(2021c)]%
        {liu2021just}
\bibfield{author}{\bibinfo{person}{Evan~Z Liu}, \bibinfo{person}{Behzad Haghgoo}, \bibinfo{person}{Annie~S Chen}, \bibinfo{person}{Aditi Raghunathan}, \bibinfo{person}{Pang~Wei Koh}, \bibinfo{person}{Shiori Sagawa}, \bibinfo{person}{Percy Liang}, {and} \bibinfo{person}{Chelsea Finn}.} \bibinfo{year}{2021}\natexlab{c}.
\newblock \showarticletitle{Just train twice: Improving group robustness without training group information}. In \bibinfo{booktitle}{\emph{International Conference on Machine Learning}}. \bibinfo{pages}{6781--6792}.
\newblock


\bibitem[Liu et~al\mbox{.}(2021a)]%
        {liu2021multiobjective}
\bibfield{author}{\bibinfo{person}{Yifei Liu}, \bibinfo{person}{Chao Chen}, \bibinfo{person}{Yazheng Liu}, \bibinfo{person}{Xi Zhang}, {and} \bibinfo{person}{Sihong Xie}.} \bibinfo{year}{2021}\natexlab{a}.
\newblock \showarticletitle{Multi-objective {Explanations} of {GNN} {Predictions}}. In \bibinfo{booktitle}{\emph{2021 {IEEE} {International} {Conference} on {Data} {Mining} ({ICDM})}}. IEEE, \bibinfo{pages}{409--418}.
\newblock


\bibitem[Liu et~al\mbox{.}({[n.\,d.]})]%
        {liucausal}
\bibfield{author}{\bibinfo{person}{Yang Liu}, \bibinfo{person}{Yushen Wei}, \bibinfo{person}{Hong Yan}, \bibinfo{person}{Guanbin Li}, {and} \bibinfo{person}{Liang Lin}.} \bibinfo{year}{[n.\,d.]}\natexlab{}.
\newblock \showarticletitle{Causal Reasoning Meets Visual Representation Learning: A Prospective Study}.
\newblock  (\bibinfo{year}{[n.\,d.]}).
\newblock


\bibitem[Liu et~al\mbox{.}(2020)]%
        {liu2020deoscillated}
\bibfield{author}{\bibinfo{person}{Zhiwei Liu}, \bibinfo{person}{Lin Meng}, \bibinfo{person}{Fei Jiang}, \bibinfo{person}{Jiawei Zhang}, {and} \bibinfo{person}{Philip~S Yu}.} \bibinfo{year}{2020}\natexlab{}.
\newblock \showarticletitle{Deoscillated graph collaborative filtering}.
\newblock \bibinfo{journal}{\emph{ArXiv}}  \bibinfo{volume}{abs/2011.02100} (\bibinfo{year}{2020}).
\newblock


\bibitem[Liu and Zhou(2020)]%
        {liu2020graph}
\bibfield{author}{\bibinfo{person}{Zhiyuan Liu} {and} \bibinfo{person}{Jie Zhou}.} \bibinfo{year}{2020}\natexlab{}.
\newblock \showarticletitle{Graph {Attention} {Networks}}.
\newblock \bibinfo{journal}{\emph{Introduction to Graph Neural Networks}} (\bibinfo{year}{2020}), \bibinfo{pages}{39--41}.
\newblock


\bibitem[L{\" u} and Zhou(2011)]%
        {lv2011link}
\bibfield{author}{\bibinfo{person}{Linyuan L{\" u}} {and} \bibinfo{person}{Tao Zhou}.} \bibinfo{year}{2011}\natexlab{}.
\newblock \showarticletitle{Link prediction in complex networks: A survey}.
\newblock \bibinfo{journal}{\emph{Physica A: statistical mechanics and its applications}} \bibinfo{volume}{390}, \bibinfo{number}{6} (\bibinfo{year}{2011}), \bibinfo{pages}{1150--1170}.
\newblock


\bibitem[Lucic et~al\mbox{.}(2022)]%
        {lucic2022cf}
\bibfield{author}{\bibinfo{person}{Ana Lucic}, \bibinfo{person}{Maartje~A Ter~Hoeve}, \bibinfo{person}{Gabriele Tolomei}, \bibinfo{person}{Maarten De~Rijke}, {and} \bibinfo{person}{Fabrizio Silvestri}.} \bibinfo{year}{2022}\natexlab{}.
\newblock \showarticletitle{Cf-gnnexplainer: Counterfactual explanations for graph neural networks}. In \bibinfo{booktitle}{\emph{International Conference on Artificial Intelligence and Statistics}}. PMLR, \bibinfo{pages}{4499--4511}.
\newblock


\bibitem[Luo et~al\mbox{.}(2020)]%
        {luo2020parameterized}
\bibfield{author}{\bibinfo{person}{Dongsheng Luo}, \bibinfo{person}{Wei Cheng}, \bibinfo{person}{Dongkuan Xu}, \bibinfo{person}{Wenchao Yu}, \bibinfo{person}{Bo Zong}, \bibinfo{person}{Haifeng Chen}, {and} \bibinfo{person}{Xiang Zhang}.} \bibinfo{year}{2020}\natexlab{}.
\newblock \showarticletitle{Parameterized explainer for graph neural network}.
\newblock \bibinfo{journal}{\emph{Advances in neural information processing systems}}  \bibinfo{volume}{33} (\bibinfo{year}{2020}), \bibinfo{pages}{19620--19631}.
\newblock


\bibitem[Ma et~al\mbox{.}(2022a)]%
        {ma2022clear}
\bibfield{author}{\bibinfo{person}{Jing Ma}, \bibinfo{person}{Ruocheng Guo}, \bibinfo{person}{Saumitra Mishra}, \bibinfo{person}{Aidong Zhang}, {and} \bibinfo{person}{Jundong Li}.} \bibinfo{year}{2022}\natexlab{a}.
\newblock \showarticletitle{CLEAR: Generative {Counterfactual} {Explanations} on {Graphs}}. In \bibinfo{booktitle}{\emph{Conference on {Neural} {Information} {Processing} {Systems} ({NeurIPS})}}.
\newblock


\bibitem[Ma et~al\mbox{.}(2022b)]%
        {ma2022learning}
\bibfield{author}{\bibinfo{person}{Jing Ma}, \bibinfo{person}{Ruocheng Guo}, \bibinfo{person}{Mengting Wan}, \bibinfo{person}{Longqi Yang}, \bibinfo{person}{Aidong Zhang}, {and} \bibinfo{person}{Jundong Li}.} \bibinfo{year}{2022}\natexlab{b}.
\newblock \showarticletitle{Learning fair node representations with graph counterfactual fairness}. In \bibinfo{booktitle}{\emph{Proceedings of the Fifteenth ACM International Conference on Web Search and Data Mining}}. \bibinfo{pages}{695--703}.
\newblock


\bibitem[Ma et~al\mbox{.}(2021)]%
        {ma2021a}
\bibfield{author}{\bibinfo{person}{Xiaoxiao Ma}, \bibinfo{person}{Jia Wu}, \bibinfo{person}{Shan Xue}, \bibinfo{person}{Jian Yang}, \bibinfo{person}{Chuan Zhou}, \bibinfo{person}{Quan~Z Sheng}, \bibinfo{person}{Hui Xiong}, {and} \bibinfo{person}{Leman Akoglu}.} \bibinfo{year}{2021}\natexlab{}.
\newblock \showarticletitle{A comprehensive survey on graph anomaly detection with deep learning}.
\newblock \bibinfo{journal}{\emph{IEEE Transactions on Knowledge and Data Engineering}} (\bibinfo{year}{2021}).
\newblock


\bibitem[Makhlouf et~al\mbox{.}(2020)]%
        {makhlouf2020survey}
\bibfield{author}{\bibinfo{person}{Karima Makhlouf}, \bibinfo{person}{Sami Zhioua}, {and} \bibinfo{person}{Catuscia Palamidessi}.} \bibinfo{year}{2020}\natexlab{}.
\newblock \showarticletitle{Survey on causal-based machine learning fairness notions}.
\newblock \bibinfo{journal}{\emph{arXiv}} (\bibinfo{year}{2020}).
\newblock


\bibitem[Martin(2019)]%
        {martin2019interpretable}
\bibfield{author}{\bibinfo{person}{Tyler Martin}.} \bibinfo{year}{2019}\natexlab{}.
\newblock \bibinfo{booktitle}{\emph{Interpretable machine learning}}.
\newblock \bibinfo{publisher}{Lulu. com}.
\newblock


\bibitem[Masrour et~al\mbox{.}(2020)]%
        {masrour2020bursting}
\bibfield{author}{\bibinfo{person}{Farzan Masrour}, \bibinfo{person}{Tyler Wilson}, \bibinfo{person}{Heng Yan}, \bibinfo{person}{Pang-Ning Tan}, {and} \bibinfo{person}{Abdol Esfahanian}.} \bibinfo{year}{2020}\natexlab{}.
\newblock \showarticletitle{Bursting the {Filter} {Bubble}: Fairness-{Aware} {Network} {Link} {Prediction}}.
\newblock \bibinfo{journal}{\emph{Proceedings of the AAAI Conference on Artificial Intelligence}} \bibinfo{volume}{34}, \bibinfo{number}{01} (\bibinfo{year}{2020}), \bibinfo{pages}{841--848}.
\newblock


\bibitem[Mehrabi et~al\mbox{.}(2021)]%
        {mehrabi2021a}
\bibfield{author}{\bibinfo{person}{Ninareh Mehrabi}, \bibinfo{person}{Fred Morstatter}, \bibinfo{person}{Nripsuta Saxena}, \bibinfo{person}{Kristina Lerman}, {and} \bibinfo{person}{Aram Galstyan}.} \bibinfo{year}{2021}\natexlab{}.
\newblock \showarticletitle{A {Survey} on {Bias} and {Fairness} in {Machine} {Learning}}.
\newblock \bibinfo{journal}{\emph{Comput. Surveys}} \bibinfo{volume}{54}, \bibinfo{number}{6} (\bibinfo{year}{2021}), \bibinfo{pages}{1--35}.
\newblock


\bibitem[Mernyei and Cangea(2020)]%
        {mernyei2020wikics}
\bibfield{author}{\bibinfo{person}{Peter Mernyei} {and} \bibinfo{person}{C{\u a}t{\u a}lina Cangea}.} \bibinfo{year}{2020}\natexlab{}.
\newblock \showarticletitle{Wiki-cs: A wikipedia-based benchmark for graph neural networks}.
\newblock \bibinfo{journal}{\emph{arXiv}} (\bibinfo{year}{2020}).
\newblock


\bibitem[Miao et~al\mbox{.}(2022)]%
        {miao2022interpretable}
\bibfield{author}{\bibinfo{person}{Siqi Miao}, \bibinfo{person}{Mia Liu}, {and} \bibinfo{person}{Pan Li}.} \bibinfo{year}{2022}\natexlab{}.
\newblock \showarticletitle{Interpretable and {Generalizable} {Graph} {Learning} via {Stochastic} {Attention} {Mechanism}.}. In \bibinfo{booktitle}{\emph{International {Conference} on {Machine} {Learning} ({ICML})}}. \bibinfo{pages}{15524--15543}.
\newblock


\bibitem[Mitrovic et~al\mbox{.}(2021)]%
        {mitrovic2021representation}
\bibfield{author}{\bibinfo{person}{Jovana Mitrovic}, \bibinfo{person}{Brian McWilliams}, \bibinfo{person}{Jacob~C. Walker}, \bibinfo{person}{Lars~Holger Buesing}, {and} \bibinfo{person}{Charles Blundell}.} \bibinfo{year}{2021}\natexlab{}.
\newblock \showarticletitle{Representation {Learning} via {Invariant} {Causal} {Mechanisms}.}. In \bibinfo{booktitle}{\emph{International {Conference} on {Learning} {Representations} ({ICLR})}}.
\newblock


\bibitem[Mu et~al\mbox{.}(2022)]%
        {mu2022alleviating}
\bibfield{author}{\bibinfo{person}{Shanlei Mu}, \bibinfo{person}{Yaliang Li}, \bibinfo{person}{Wayne~Xin Zhao}, \bibinfo{person}{Jingyuan Wang}, \bibinfo{person}{Bolin Ding}, {and} \bibinfo{person}{Ji-Rong Wen}.} \bibinfo{year}{2022}\natexlab{}.
\newblock \showarticletitle{Alleviating {Spurious} {Correlations} in {Knowledge}-aware {Recommendations} through {Counterfactual} {Generator}}. In \bibinfo{booktitle}{\emph{Proceedings of the 45th {International} {ACM} {SIGIR} {Conference} on {Research} and {Development} in {Information} {Retrieval}}}. \bibinfo{pages}{1401--1411}.
\newblock


\bibitem[Naderializadeh et~al\mbox{.}(2020)]%
        {naderializadeh2020wireless}
\bibfield{author}{\bibinfo{person}{Navid Naderializadeh}, \bibinfo{person}{Mark Eisen}, {and} \bibinfo{person}{Alejandro Ribeiro}.} \bibinfo{year}{2020}\natexlab{}.
\newblock \showarticletitle{Wireless power control via counterfactual optimization of graph neural networks}. In \bibinfo{booktitle}{\emph{2020 IEEE 21st International Workshop on Signal Processing Advances in Wireless Communications (SPAWC)}}. IEEE, \bibinfo{pages}{1--5}.
\newblock


\bibitem[Nassar et~al\mbox{.}(2019)]%
        {nassar2019pairwise}
\bibfield{author}{\bibinfo{person}{Huda Nassar}, \bibinfo{person}{Austin~R Benson}, {and} \bibinfo{person}{David~F Gleich}.} \bibinfo{year}{2019}\natexlab{}.
\newblock \showarticletitle{Pairwise link prediction}. In \bibinfo{booktitle}{\emph{Proceedings of the 2019 {IEEE}/{ACM} {International} {Conference} on {Advances} in {Social} {Networks} {Analysis} and {Mining}}}. \bibinfo{pages}{386--393}.
\newblock


\bibitem[Nathan et~al\mbox{.}(2019)]%
        {nathan2019endtoend}
\bibfield{author}{\bibinfo{person}{Vikram Nathan}, \bibinfo{person}{Vibhaalakshmi Sivaraman}, \bibinfo{person}{Ravichandra Addanki}, \bibinfo{person}{Mehrdad Khani}, \bibinfo{person}{Prateesh Goyal}, {and} \bibinfo{person}{Mohammad Alizadeh}.} \bibinfo{year}{2019}\natexlab{}.
\newblock \showarticletitle{End-to-end transport for video {QoE} fairness}. In \bibinfo{booktitle}{\emph{Proceedings of the {ACM} {Special} {Interest} {Group} on {Data} {Communication}}}. \bibinfo{pages}{408--423}.
\newblock


\bibitem[No{\' e} et~al\mbox{.}(2020)]%
        {noe2020machine}
\bibfield{author}{\bibinfo{person}{Frank No{\' e}}, \bibinfo{person}{Alexandre Tkatchenko}, \bibinfo{person}{Klaus-Robert M{\" u}ller}, {and} \bibinfo{person}{Cecilia Clementi}.} \bibinfo{year}{2020}\natexlab{}.
\newblock \showarticletitle{Machine learning for molecular simulation}.
\newblock \bibinfo{journal}{\emph{Annual review of physical chemistry}}  \bibinfo{volume}{71} (\bibinfo{year}{2020}), \bibinfo{pages}{361--390}.
\newblock


\bibitem[Numeroso and Bacciu(2021)]%
        {numeroso2021meg}
\bibfield{author}{\bibinfo{person}{Danilo Numeroso} {and} \bibinfo{person}{Davide Bacciu}.} \bibinfo{year}{2021}\natexlab{}.
\newblock \showarticletitle{MEG: Generating {Molecular} {Counterfactual} {Explanations} for {Deep} {Graph} {Networks}}. In \bibinfo{booktitle}{\emph{2021 {International} {Joint} {Conference} on {Neural} {Networks} ({IJCNN})}}. IEEE, \bibinfo{pages}{1--8}.
\newblock


\bibitem[Ohly(2022)]%
        {ohly2022flowbased}
\bibfield{author}{\bibinfo{person}{Lorenz Ohly}.} \bibinfo{year}{2022}\natexlab{}.
\newblock \showarticletitle{Flow-based counterfactuals for interpretable graph node classification}.
\newblock  (\bibinfo{year}{2022}).
\newblock


\bibitem[Oneto et~al\mbox{.}(2022)]%
        {oneto2022towards}
\bibfield{author}{\bibinfo{person}{Luca Oneto}, \bibinfo{person}{Nicol{\'o} Navarin}, \bibinfo{person}{Battista Biggio}, \bibinfo{person}{Federico Errica}, \bibinfo{person}{Alessio Micheli}, \bibinfo{person}{Franco Scarselli}, \bibinfo{person}{Monica Bianchini}, \bibinfo{person}{Luca Demetrio}, \bibinfo{person}{Pietro Bongini}, \bibinfo{person}{Armando Tacchella}, {et~al\mbox{.}}} \bibinfo{year}{2022}\natexlab{}.
\newblock \showarticletitle{Towards learning trustworthily, automatically, and with guarantees on graphs: An overview}.
\newblock \bibinfo{journal}{\emph{Neurocomputing}} (\bibinfo{year}{2022}).
\newblock


\bibitem[Pan et~al\mbox{.}(2022)]%
        {pan2022neural}
\bibfield{author}{\bibinfo{person}{Liming Pan}, \bibinfo{person}{Cheng Shi}, {and} \bibinfo{person}{Ivan Dokmanic}.} \bibinfo{year}{2022}\natexlab{}.
\newblock \showarticletitle{Neural {Link} {Prediction} with {Walk} {Pooling}.}. In \bibinfo{booktitle}{\emph{International {Conference} on {Learning} {Representations} ({ICLR})}}.
\newblock


\bibitem[Papp et~al\mbox{.}(2021)]%
        {papp2021dropgnn}
\bibfield{author}{\bibinfo{person}{P{\' a}l~Andr{\' a}s Papp}, \bibinfo{person}{Karolis Martinkus}, \bibinfo{person}{Lukas Faber}, {and} \bibinfo{person}{Roger Wattenhofer}.} \bibinfo{year}{2021}\natexlab{}.
\newblock \showarticletitle{DropGNN: Random {Dropouts} {Increase} the {Expressiveness} of {Graph} {Neural} {Networks}.}. In \bibinfo{booktitle}{\emph{Conference on {Neural} {Information} {Processing} {Systems} ({NeurIPS})}}. \bibinfo{pages}{21997--22009}.
\newblock


\bibitem[Pearl(2009)]%
        {pearl2009causal}
\bibfield{author}{\bibinfo{person}{Judea Pearl}.} \bibinfo{year}{2009}\natexlab{}.
\newblock \showarticletitle{Causal inference in statistics: An overview}.
\newblock \bibinfo{journal}{\emph{Statistics Surveys}}  \bibinfo{volume}{3} (\bibinfo{year}{2009}), \bibinfo{pages}{96--146}.
\newblock


\bibitem[Perry-Smith and Shalley(2003)]%
        {Perry2003social}
\bibfield{author}{\bibinfo{person}{Jill~E Perry-Smith} {and} \bibinfo{person}{Christina~E Shalley}.} \bibinfo{year}{2003}\natexlab{}.
\newblock \showarticletitle{The social side of creativity: A static and dynamic social network perspective}.
\newblock \bibinfo{journal}{\emph{Academy of management review}} \bibinfo{volume}{28}, \bibinfo{number}{1} (\bibinfo{year}{2003}), \bibinfo{pages}{89--106}.
\newblock


\bibitem[Pham and Zhang(2022)]%
        {pham2022counterfactual}
\bibfield{author}{\bibinfo{person}{David Pham} {and} \bibinfo{person}{Yongfeng Zhang}.} \bibinfo{year}{2022}\natexlab{}.
\newblock \showarticletitle{Counterfactual based reinforcement learning for graph neural networks}.
\newblock \bibinfo{journal}{\emph{Annals of Operations Research}} (\bibinfo{year}{2022}), \bibinfo{pages}{1--17}.
\newblock


\bibitem[Pitis et~al\mbox{.}(2020)]%
        {pitis2020counterfactual}
\bibfield{author}{\bibinfo{person}{Silviu Pitis}, \bibinfo{person}{Elliot Creager}, {and} \bibinfo{person}{Animesh Garg}.} \bibinfo{year}{2020}\natexlab{}.
\newblock \showarticletitle{Counterfactual data augmentation using locally factored dynamics}.
\newblock \bibinfo{journal}{\emph{Advances in Neural Information Processing Systems}}  \bibinfo{volume}{33} (\bibinfo{year}{2020}), \bibinfo{pages}{3976--3990}.
\newblock


\bibitem[Prado-Romero et~al\mbox{.}(2022b)]%
        {Prado2022Ensemble}
\bibfield{author}{\bibinfo{person}{Mario~Alfonso Prado-Romero}, \bibinfo{person}{Bardh Prenkaj}, \bibinfo{person}{Giovanni Stilo}, \bibinfo{person}{Alessandro Celi}, \bibinfo{person}{Ernesto~Luis Estevanell-Valladares}, {and} \bibinfo{person}{Daniel Alejandro~Vald{\' e}s P{\' e}rez}.} \bibinfo{year}{2022}\natexlab{b}.
\newblock \showarticletitle{Ensemble {Approaches} for {Graph} {Counterfactual} {Explanations}.}. In \bibinfo{booktitle}{\emph{International {Conference} of the {Italian} {Association} for {Artificial} {Intelligence} ({AI}*{IA}) {Workshop}}}. \bibinfo{pages}{88--97}.
\newblock


\bibitem[Prado-Romero et~al\mbox{.}(2022a)]%
        {Prado2022Survey}
\bibfield{author}{\bibinfo{person}{Mario~Alfonso Prado-Romero}, \bibinfo{person}{Bardh Prenkaj}, \bibinfo{person}{Giovanni Stilo}, {and} \bibinfo{person}{Fosca Giannotti}.} \bibinfo{year}{2022}\natexlab{a}.
\newblock \showarticletitle{A {Survey} on {Graph} {Counterfactual} {Explanations}: Definitions, {Methods}, {Evaluation}}.
\newblock \bibinfo{journal}{\emph{openreview.net}}  \bibinfo{volume}{abs/2210.12089} (\bibinfo{year}{2022}).
\newblock


\bibitem[Prado-Romero and Stilo(2022)]%
        {prado2022gretel}
\bibfield{author}{\bibinfo{person}{Mario~Alfonso Prado-Romero} {and} \bibinfo{person}{Giovanni Stilo}.} \bibinfo{year}{2022}\natexlab{}.
\newblock \showarticletitle{GRETEL: A unified framework for Graph Counterfactual Explanation Evaluation}.
\newblock \bibinfo{journal}{\emph{arXiv preprint arXiv:2206.02957}} (\bibinfo{year}{2022}).
\newblock


\bibitem[Pr{\v{z}}ulj(2007)]%
        {prvzulj2007biological}
\bibfield{author}{\bibinfo{person}{Nata{\v{s}}a Pr{\v{z}}ulj}.} \bibinfo{year}{2007}\natexlab{}.
\newblock \showarticletitle{Biological network comparison using graphlet degree distribution}.
\newblock \bibinfo{journal}{\emph{Bioinformatics}} \bibinfo{volume}{23}, \bibinfo{number}{2} (\bibinfo{year}{2007}), \bibinfo{pages}{e177--e183}.
\newblock


\bibitem[Qin et~al\mbox{.}(2022)]%
        {qin2022capturing}
\bibfield{author}{\bibinfo{person}{Shiyi Qin}, \bibinfo{person}{Shengli Jiang}, \bibinfo{person}{Jianping Li}, \bibinfo{person}{Prasanna Balaprakash}, \bibinfo{person}{Reid Van~Lehn}, {and} \bibinfo{person}{Victor Zavala}.} \bibinfo{year}{2022}\natexlab{}.
\newblock \showarticletitle{Capturing Molecular Interactions in Graph Neural Networks: A Case Study in Multi-Component Phase Equilibrium}.
\newblock  (\bibinfo{year}{2022}).
\newblock


\bibitem[Qiu et~al\mbox{.}(2018)]%
        {qiu2018network}
\bibfield{author}{\bibinfo{person}{Jiezhong Qiu}, \bibinfo{person}{Yuxiao Dong}, \bibinfo{person}{Hao Ma}, \bibinfo{person}{Jian Li}, \bibinfo{person}{Kuansan Wang}, {and} \bibinfo{person}{Jie Tang}.} \bibinfo{year}{2018}\natexlab{}.
\newblock \showarticletitle{Network embedding as matrix factorization: Unifying deepwalk, line, pte, and node2vec}. In \bibinfo{booktitle}{\emph{Proceedings of the eleventh {ACM} international conference on web search and data mining}}. \bibinfo{pages}{459--467}.
\newblock


\bibitem[Rahman et~al\mbox{.}(2019)]%
        {rahman2019fairwalk}
\bibfield{author}{\bibinfo{person}{Tahleen~A. Rahman}, \bibinfo{person}{Bartlomiej Surma}, \bibinfo{person}{Michael Backes}, {and} \bibinfo{person}{Yang Zhang}.} \bibinfo{year}{2019}\natexlab{}.
\newblock \showarticletitle{Fairwalk: Towards {Fair} {Graph} {Embedding}.}. In \bibinfo{booktitle}{\emph{International Joint Conference on Artificial Intelligence, {IJCAI}}}. \bibinfo{pages}{3289--3295}.
\newblock


\bibitem[Ran and Boyce(1996)]%
        {ran1996modeling}
\bibfield{author}{\bibinfo{person}{B. Ran} {and} \bibinfo{person}{D. Boyce}.} \bibinfo{year}{1996}\natexlab{}.
\newblock \bibinfo{booktitle}{\emph{Modeling dynamic transportation networks: an intelligent transportation system oriented approach}}.
\newblock \bibinfo{publisher}{Springer Science \& Business Media}.
\newblock


\bibitem[Rubin(1974)]%
        {rubin1974estimating}
\bibfield{author}{\bibinfo{person}{Donald~B Rubin}.} \bibinfo{year}{1974}\natexlab{}.
\newblock \showarticletitle{Estimating causal effects of treatments in randomized and nonrandomized studies.}
\newblock \bibinfo{journal}{\emph{Journal of educational Psychology}} \bibinfo{volume}{66}, \bibinfo{number}{5} (\bibinfo{year}{1974}), \bibinfo{pages}{688}.
\newblock


\bibitem[Rubin(2005)]%
        {rubin2005causal}
\bibfield{author}{\bibinfo{person}{Donald~B Rubin}.} \bibinfo{year}{2005}\natexlab{}.
\newblock \showarticletitle{Causal inference using potential outcomes: Design, modeling, decisions}.
\newblock \bibinfo{journal}{\emph{J. Amer. Statist. Assoc.}} \bibinfo{volume}{100}, \bibinfo{number}{469} (\bibinfo{year}{2005}), \bibinfo{pages}{322--331}.
\newblock


\bibitem[Sagawa et~al\mbox{.}(2019)]%
        {sagawa2019distributionally}
\bibfield{author}{\bibinfo{person}{Shiori Sagawa}, \bibinfo{person}{Pang~Wei Koh}, \bibinfo{person}{Tatsunori~B Hashimoto}, {and} \bibinfo{person}{Percy Liang}.} \bibinfo{year}{2019}\natexlab{}.
\newblock \showarticletitle{Distributionally Robust Neural Networks}. In \bibinfo{booktitle}{\emph{International Conference on Learning Representations}}.
\newblock


\bibitem[Sanchez et~al\mbox{.}(2022)]%
        {sanchez2022causal}
\bibfield{author}{\bibinfo{person}{Pedro Sanchez}, \bibinfo{person}{Jeremy~P Voisey}, \bibinfo{person}{Tian Xia}, \bibinfo{person}{Hannah~I Watson}, \bibinfo{person}{Alison~Q O’Neil}, {and} \bibinfo{person}{Sotirios~A Tsaftaris}.} \bibinfo{year}{2022}\natexlab{}.
\newblock \showarticletitle{Causal machine learning for healthcare and precision medicine}.
\newblock \bibinfo{journal}{\emph{Royal Society Open Science}} \bibinfo{volume}{9}, \bibinfo{number}{8} (\bibinfo{year}{2022}), \bibinfo{pages}{220638}.
\newblock


\bibitem[Sarwar et~al\mbox{.}(2001)]%
        {sarwar2001itembased}
\bibfield{author}{\bibinfo{person}{Badrul Sarwar}, \bibinfo{person}{George Karypis}, \bibinfo{person}{Joseph Konstan}, {and} \bibinfo{person}{John Riedl}.} \bibinfo{year}{2001}\natexlab{}.
\newblock \showarticletitle{Item-based collaborative filtering recommendation algorithms}. In \bibinfo{booktitle}{\emph{Proceedings of the 10th international conference on {World} {Wide} {Web}}}. \bibinfo{pages}{285--295}.
\newblock


\bibitem[Schlichtkrull et~al\mbox{.}(2018)]%
        {schlichtkrull2018modeling}
\bibfield{author}{\bibinfo{person}{Michael Schlichtkrull}, \bibinfo{person}{Thomas~N. Kipf}, \bibinfo{person}{Peter Bloem}, \bibinfo{person}{Rianne van~den Berg}, \bibinfo{person}{Ivan Titov}, {and} \bibinfo{person}{Max Welling}.} \bibinfo{year}{2018}\natexlab{}.
\newblock \bibinfo{booktitle}{\emph{Modeling {Relational} {Data} with {Graph} {Convolutional} {Networks}}}.
\newblock \bibinfo{publisher}{Springer International Publishing}. 593--607 pages.
\newblock


\bibitem[Sch{\" o}lkopf and K{\" u}gelgen(2022)]%
        {scholkopf2022from}
\bibfield{author}{\bibinfo{person}{Bernhard Sch{\" o}lkopf} {and} \bibinfo{person}{Julius~von K{\" u}gelgen}.} \bibinfo{year}{2022}\natexlab{}.
\newblock \showarticletitle{From {Statistical} to {Causal} {Learning}}.
\newblock \bibinfo{journal}{\emph{arXiv}} (\bibinfo{year}{2022}).
\newblock


\bibitem[Sch{\" o}lkopf et~al\mbox{.}(2021)]%
        {scholkopf2021towards}
\bibfield{author}{\bibinfo{person}{Bernhard Sch{\" o}lkopf}, \bibinfo{person}{Francesco Locatello}, \bibinfo{person}{Stefan Bauer}, \bibinfo{person}{Nan~Rosemary Ke}, \bibinfo{person}{Nal Kalchbrenner}, \bibinfo{person}{Anirudh Goyal}, {and} \bibinfo{person}{Yoshua Bengio}.} \bibinfo{year}{2021}\natexlab{}.
\newblock \showarticletitle{Towards {Causal} {Representation} {Learning}}.
\newblock \bibinfo{journal}{\emph{arXiv}} (\bibinfo{year}{2021}).
\newblock


\bibitem[Sedhain et~al\mbox{.}(2015)]%
        {sedhain2015autorec}
\bibfield{author}{\bibinfo{person}{Suvash Sedhain}, \bibinfo{person}{Aditya~Krishna Menon}, \bibinfo{person}{Scott Sanner}, {and} \bibinfo{person}{Lexing Xie}.} \bibinfo{year}{2015}\natexlab{}.
\newblock \showarticletitle{Autorec: Autoencoders meet collaborative filtering}. In \bibinfo{booktitle}{\emph{Proceedings of the 24th international conference on {World} {Wide} {Web}}}. \bibinfo{pages}{111--112}.
\newblock


\bibitem[Shalaby et~al\mbox{.}(2019)]%
        {shalaby2019beyond}
\bibfield{author}{\bibinfo{person}{Walid Shalaby}, \bibinfo{person}{Wlodek Zadrozny}, {and} \bibinfo{person}{Hongxia Jin}.} \bibinfo{year}{2019}\natexlab{}.
\newblock \showarticletitle{Beyond word embeddings: learning entity and concept representations from large scale knowledge bases}.
\newblock \bibinfo{journal}{\emph{Information Retrieval Journal}}  \bibinfo{volume}{22} (\bibinfo{year}{2019}), \bibinfo{pages}{525--542}.
\newblock


\bibitem[Shanmugam(2018)]%
        {shanmugam2018elements}
\bibfield{author}{\bibinfo{person}{Ramalingam Shanmugam}.} \bibinfo{year}{2018}\natexlab{}.
\newblock \showarticletitle{Elements of causal inference: foundations and learning algorithms}.
\newblock \bibinfo{journal}{\emph{Journal of Statistical Computation and Simulation}} \bibinfo{volume}{88}, \bibinfo{number}{16} (\bibinfo{year}{2018}), \bibinfo{pages}{3248--3248}.
\newblock


\bibitem[Shchur and Gunnemann(2019)]%
        {shchur2019overlapping}
\bibfield{author}{\bibinfo{person}{Oleksandr Shchur} {and} \bibinfo{person}{Stephan Gunnemann}.} \bibinfo{year}{2019}\natexlab{}.
\newblock \showarticletitle{Overlapping community detection with graph neural networks}.
\newblock \bibinfo{journal}{\emph{openreview.net}} (\bibinfo{year}{2019}).
\newblock


\bibitem[Shu et~al\mbox{.}(2017)]%
        {shu2017fake}
\bibfield{author}{\bibinfo{person}{Kai Shu}, \bibinfo{person}{Amy Sliva}, \bibinfo{person}{Suhang Wang}, \bibinfo{person}{Jiliang Tang}, {and} \bibinfo{person}{Huan Liu}.} \bibinfo{year}{2017}\natexlab{}.
\newblock \showarticletitle{Fake {News} {Detection} on {Social} {Media}: A {Data} {Mining} {Perspective}.}
\newblock \bibinfo{journal}{\emph{SIGKDD Explorations}} \bibinfo{volume}{19}, \bibinfo{number}{1} (\bibinfo{year}{2017}), \bibinfo{pages}{22--36}.
\newblock


\bibitem[Song et~al\mbox{.}(2023)]%
        {SongW0L0Y23}
\bibfield{author}{\bibinfo{person}{Wenzhuo Song}, \bibinfo{person}{Shoujin Wang}, \bibinfo{person}{Yan Wang}, \bibinfo{person}{Kunpeng Liu}, \bibinfo{person}{Xueyan Liu}, {and} \bibinfo{person}{Minghao Yin}.} \bibinfo{year}{2023}\natexlab{}.
\newblock \showarticletitle{A Counterfactual Collaborative Session-based Recommender System}. In \bibinfo{booktitle}{\emph{Proceedings of the {ACM} Web Conference, {WWW}}}. \bibinfo{publisher}{{ACM}}, \bibinfo{pages}{971--982}.
\newblock
\urldef\tempurl%
\url{https://doi.org/10.1145/3543507.3583321}
\showDOI{\tempurl}


\bibitem[Stepin et~al\mbox{.}(2021)]%
        {9321372}
\bibfield{author}{\bibinfo{person}{Ilia Stepin}, \bibinfo{person}{Jose~M. Alonso}, \bibinfo{person}{Alejandro Catala}, {and} \bibinfo{person}{Martín Pereira-Fariña}.} \bibinfo{year}{2021}\natexlab{}.
\newblock \showarticletitle{A Survey of Contrastive and Counterfactual Explanation Generation Methods for Explainable Artificial Intelligence}.
\newblock \bibinfo{journal}{\emph{IEEE Access}}  \bibinfo{volume}{9} (\bibinfo{year}{2021}), \bibinfo{pages}{11974--12001}.
\newblock
\urldef\tempurl%
\url{https://doi.org/10.1109/ACCESS.2021.3051315}
\showDOI{\tempurl}


\bibitem[Su et~al\mbox{.}(2020)]%
        {su2020counterfactual}
\bibfield{author}{\bibinfo{person}{Jianyu Su}, \bibinfo{person}{Stephen Adams}, {and} \bibinfo{person}{Peter~A Beling}.} \bibinfo{year}{2020}\natexlab{}.
\newblock \showarticletitle{Counterfactual multi-agent reinforcement learning with graph convolution communication}.
\newblock \bibinfo{journal}{\emph{arXiv preprint arXiv:2004.00470}} (\bibinfo{year}{2020}).
\newblock


\bibitem[Suh et~al\mbox{.}(2010)]%
        {suh2010want}
\bibfield{author}{\bibinfo{person}{Bongwon Suh}, \bibinfo{person}{Lichan Hong}, \bibinfo{person}{Peter Pirolli}, {and} \bibinfo{person}{Ed~H Chi}.} \bibinfo{year}{2010}\natexlab{}.
\newblock \showarticletitle{Want to be retweeted? large scale analytics on factors impacting retweet in twitter network}. In \bibinfo{booktitle}{\emph{2010 {IEEE} second international conference on social computing}}. IEEE, \bibinfo{pages}{177--184}.
\newblock


\bibitem[Sui et~al\mbox{.}(2022)]%
        {sui2022causal}
\bibfield{author}{\bibinfo{person}{Yongduo Sui}, \bibinfo{person}{Xiang Wang}, \bibinfo{person}{Jiancan Wu}, \bibinfo{person}{Min Lin}, \bibinfo{person}{Xiangnan He}, {and} \bibinfo{person}{Tat-Seng Chua}.} \bibinfo{year}{2022}\natexlab{}.
\newblock \showarticletitle{Causal attention for interpretable and generalizable graph classification}. In \bibinfo{booktitle}{\emph{Proceedings of the 28th {ACM} {SIGKDD} {Conference} on {Knowledge} {Discovery} and {Data} {Mining}}}. \bibinfo{pages}{1696--1705}.
\newblock


\bibitem[Sun et~al\mbox{.}(2021)]%
        {sun2021preserve}
\bibfield{author}{\bibinfo{person}{Yi Sun}, \bibinfo{person}{Abel Valente}, \bibinfo{person}{Sijia Liu}, {and} \bibinfo{person}{Dakuo Wang}.} \bibinfo{year}{2021}\natexlab{}.
\newblock \showarticletitle{Preserve, promote, or attack? gnn explanation via topology perturbation}.
\newblock \bibinfo{journal}{\emph{arXiv preprint arXiv:2103.13944}} (\bibinfo{year}{2021}).
\newblock


\bibitem[Sun et~al\mbox{.}(2019)]%
        {sun2019rotate}
\bibfield{author}{\bibinfo{person}{Zhiqing Sun}, \bibinfo{person}{Zhi-Hong Deng}, \bibinfo{person}{Jian-Yun Nie}, {and} \bibinfo{person}{Jian Tang}.} \bibinfo{year}{2019}\natexlab{}.
\newblock \showarticletitle{RotatE: Knowledge {Graph} {Embedding} by {Relational} {Rotation} in {Complex} {Space}.}. In \bibinfo{booktitle}{\emph{International {Conference} on {Learning} {Representations} ({ICLR})}}.
\newblock


\bibitem[Sutton and Barto(1998)]%
        {sutton1998reinforcement}
\bibfield{author}{\bibinfo{person}{R.S. Sutton} {and} \bibinfo{person}{A.G. Barto}.} \bibinfo{year}{1998}\natexlab{}.
\newblock \bibinfo{booktitle}{\emph{Reinforcement {Learning}: An {Introduction}}}. Vol.~\bibinfo{volume}{9}.
\newblock \bibinfo{publisher}{MIT press}. 1054--1054 pages.
\newblock


\bibitem[Tabassum et~al\mbox{.}(2018)]%
        {tabassum2018social}
\bibfield{author}{\bibinfo{person}{Shazia Tabassum}, \bibinfo{person}{Fabiola~SF Pereira}, \bibinfo{person}{Sofia Fernandes}, {and} \bibinfo{person}{Jo{\~a}o Gama}.} \bibinfo{year}{2018}\natexlab{}.
\newblock \showarticletitle{Social network analysis: An overview}.
\newblock \bibinfo{journal}{\emph{Wiley Interdisciplinary Reviews: Data Mining and Knowledge Discovery}} \bibinfo{volume}{8}, \bibinfo{number}{5} (\bibinfo{year}{2018}), \bibinfo{pages}{e1256}.
\newblock


\bibitem[Takac and Zabovsky(2012)]%
        {takac2012data}
\bibfield{author}{\bibinfo{person}{Lubos Takac} {and} \bibinfo{person}{Michal Zabovsky}.} \bibinfo{year}{2012}\natexlab{}.
\newblock \showarticletitle{Data analysis in public social networks}. In \bibinfo{booktitle}{\emph{International scientific conference and international workshop present day trends of innovations}}, Vol.~\bibinfo{volume}{1}. Present Day Trends of Innovations Lamza Poland.
\newblock


\bibitem[Tan et~al\mbox{.}(2022)]%
        {tan2022learning}
\bibfield{author}{\bibinfo{person}{Juntao Tan}, \bibinfo{person}{Shijie Geng}, \bibinfo{person}{Zuohui Fu}, \bibinfo{person}{Yingqiang Ge}, \bibinfo{person}{Shuyuan Xu}, \bibinfo{person}{Yunqi Li}, {and} \bibinfo{person}{Yongfeng Zhang}.} \bibinfo{year}{2022}\natexlab{}.
\newblock \showarticletitle{Learning and evaluating graph neural network explanations based on counterfactual and factual reasoning}. In \bibinfo{booktitle}{\emph{Proceedings of the ACM Web Conference 2022}}. \bibinfo{pages}{1018--1027}.
\newblock


\bibitem[Tang et~al\mbox{.}(2015)]%
        {tang2015line}
\bibfield{author}{\bibinfo{person}{Jian Tang}, \bibinfo{person}{Meng Qu}, \bibinfo{person}{Mingzhe Wang}, \bibinfo{person}{Ming Zhang}, \bibinfo{person}{Jun Yan}, {and} \bibinfo{person}{Qiaozhu Mei}.} \bibinfo{year}{2015}\natexlab{}.
\newblock \showarticletitle{LINE: Large-scale {Information} {Network} {Embedding}}. In \bibinfo{booktitle}{\emph{Proceedings of the 24th {International} {Conference} on {World} {Wide} {Web}}}. International World Wide Web Conferences Steering Committee.
\newblock


\bibitem[Traud et~al\mbox{.}(2012)]%
        {traud2012social}
\bibfield{author}{\bibinfo{person}{Amanda~L Traud}, \bibinfo{person}{Peter~J Mucha}, {and} \bibinfo{person}{Mason~A Porter}.} \bibinfo{year}{2012}\natexlab{}.
\newblock \showarticletitle{Social structure of facebook networks}.
\newblock \bibinfo{journal}{\emph{Physica A: Statistical Mechanics and its Applications}} \bibinfo{volume}{391}, \bibinfo{number}{16} (\bibinfo{year}{2012}), \bibinfo{pages}{4165--4180}.
\newblock


\bibitem[Velickovic et~al\mbox{.}(2018)]%
        {velickovic2018graph}
\bibfield{author}{\bibinfo{person}{Petar Velickovic}, \bibinfo{person}{Guillem Cucurull}, \bibinfo{person}{Arantxa Casanova}, \bibinfo{person}{Adriana Romero}, \bibinfo{person}{Pietro Li{\` o}}, {and} \bibinfo{person}{Yoshua Bengio}.} \bibinfo{year}{2018}\natexlab{}.
\newblock \showarticletitle{Graph {Attention} {Networks}.}. In \bibinfo{booktitle}{\emph{International {Conference} on {Learning} {Representations} ({ICLR})}}.
\newblock


\bibitem[Verma et~al\mbox{.}(2020)]%
        {verma2020counterfactual}
\bibfield{author}{\bibinfo{person}{Sahil Verma}, \bibinfo{person}{Varich Boonsanong}, \bibinfo{person}{Minh Hoang}, \bibinfo{person}{Keegan~E Hines}, \bibinfo{person}{John~P Dickerson}, {and} \bibinfo{person}{Chirag Shah}.} \bibinfo{year}{2020}\natexlab{}.
\newblock \showarticletitle{Counterfactual {Explanations} and {Algorithmic} {Recourses} for {Machine} {Learning}: A {Review}}.
\newblock \bibinfo{journal}{\emph{arXiv preprint arXiv:2010.10596}} (\bibinfo{year}{2020}).
\newblock


\bibitem[Vishwanathan et~al\mbox{.}(2010)]%
        {vishwanathan2010graph}
\bibfield{author}{\bibinfo{person}{S~Vichy~N Vishwanathan}, \bibinfo{person}{Nicol~N Schraudolph}, \bibinfo{person}{Risi Kondor}, {and} \bibinfo{person}{Karsten~M Borgwardt}.} \bibinfo{year}{2010}\natexlab{}.
\newblock \showarticletitle{Graph kernels}.
\newblock \bibinfo{journal}{\emph{Journal of Machine Learning Research}}  \bibinfo{volume}{11} (\bibinfo{year}{2010}), \bibinfo{pages}{1201--1242}.
\newblock


\bibitem[Vlontzos et~al\mbox{.}(2022)]%
        {vlontzos2022review}
\bibfield{author}{\bibinfo{person}{Athanasios Vlontzos}, \bibinfo{person}{Daniel Rueckert}, {and} \bibinfo{person}{Bernhard Kainz}.} \bibinfo{year}{2022}\natexlab{}.
\newblock \showarticletitle{A review of causality for learning algorithms in medical image analysis}.
\newblock \bibinfo{journal}{\emph{arXiv preprint arXiv:2206.05498}} (\bibinfo{year}{2022}).
\newblock


\bibitem[Wale et~al\mbox{.}(2008)]%
        {wale2008comparison}
\bibfield{author}{\bibinfo{person}{Nikil Wale}, \bibinfo{person}{Ian~A Watson}, {and} \bibinfo{person}{George Karypis}.} \bibinfo{year}{2008}\natexlab{}.
\newblock \showarticletitle{Comparison of descriptor spaces for chemical compound retrieval and classification}.
\newblock \bibinfo{journal}{\emph{Knowledge and Information Systems}} \bibinfo{volume}{14}, \bibinfo{number}{3} (\bibinfo{year}{2008}), \bibinfo{pages}{347--375}.
\newblock


\bibitem[Wang et~al\mbox{.}(2015)]%
        {wang2015collaborative}
\bibfield{author}{\bibinfo{person}{Hao Wang}, \bibinfo{person}{Naiyan Wang}, {and} \bibinfo{person}{Dit-Yan Yeung}.} \bibinfo{year}{2015}\natexlab{}.
\newblock \showarticletitle{Collaborative deep learning for recommender systems}. In \bibinfo{booktitle}{\emph{Proceedings of the 21th {ACM} {SIGKDD} international conference on knowledge discovery and data mining}}. \bibinfo{pages}{1235--1244}.
\newblock


\bibitem[Wang et~al\mbox{.}(2019)]%
        {wang2019knowledgeaware}
\bibfield{author}{\bibinfo{person}{Hongwei Wang}, \bibinfo{person}{Fuzheng Zhang}, \bibinfo{person}{Mengdi Zhang}, \bibinfo{person}{Jure Leskovec}, \bibinfo{person}{Miao Zhao}, \bibinfo{person}{Wenjie Li}, {and} \bibinfo{person}{Zhongyuan Wang}.} \bibinfo{year}{2019}\natexlab{}.
\newblock \showarticletitle{Knowledge-aware graph neural networks with label smoothness regularization for recommender systems}. In \bibinfo{booktitle}{\emph{Proceedings of the 25th {ACM} {SIGKDD} international conference on knowledge discovery \& data mining}}. \bibinfo{pages}{968--977}.
\newblock


\bibitem[Wang et~al\mbox{.}(2021b)]%
        {wang2021generalizing}
\bibfield{author}{\bibinfo{person}{Jindong Wang}, \bibinfo{person}{Cuiling Lan}, \bibinfo{person}{Chang Liu}, \bibinfo{person}{Yidong Ouyang}, {and} \bibinfo{person}{Tao Qin}.} \bibinfo{year}{2021}\natexlab{b}.
\newblock \showarticletitle{Generalizing to {Unseen} {Domains}: A {Survey} on {Domain} {Generalization}.}. In \bibinfo{booktitle}{\emph{International {Joint} {Conference} on {Artificial} {Intelligence} ({IJCAI})}}. International Joint Conferences on Artificial Intelligence Organization, \bibinfo{pages}{4627--4635}.
\newblock


\bibitem[Wang et~al\mbox{.}(2022a)]%
        {wang2022unbiased}
\bibfield{author}{\bibinfo{person}{Nan Wang}, \bibinfo{person}{Lu Lin}, \bibinfo{person}{Jundong Li}, {and} \bibinfo{person}{Hongning Wang}.} \bibinfo{year}{2022}\natexlab{a}.
\newblock \showarticletitle{Unbiased {Graph} {Embedding} with {Biased} {Graph} {Observations}}. In \bibinfo{booktitle}{\emph{Proceedings of the {ACM} {Web} {Conference} 2022}}. \bibinfo{pages}{1423--1433}.
\newblock


\bibitem[Wang et~al\mbox{.}(2018)]%
        {wang2018acekg}
\bibfield{author}{\bibinfo{person}{Ruijie Wang}, \bibinfo{person}{Yuchen Yan}, \bibinfo{person}{Jialu Wang}, \bibinfo{person}{Yuting Jia}, \bibinfo{person}{Ye Zhang}, \bibinfo{person}{Weinan Zhang}, {and} \bibinfo{person}{Xinbing Wang}.} \bibinfo{year}{2018}\natexlab{}.
\newblock \showarticletitle{Acekg: A large-scale knowledge graph for academic data mining}. In \bibinfo{booktitle}{\emph{Proceedings of the 27th {ACM} international conference on information and knowledge management}}. \bibinfo{pages}{1487--1490}.
\newblock


\bibitem[Wang et~al\mbox{.}(2021a)]%
        {wang2021clicks}
\bibfield{author}{\bibinfo{person}{Wenjie Wang}, \bibinfo{person}{Fuli Feng}, \bibinfo{person}{Xiangnan He}, \bibinfo{person}{Hanwang Zhang}, {and} \bibinfo{person}{Tat-Seng Chua}.} \bibinfo{year}{2021}\natexlab{a}.
\newblock \showarticletitle{Clicks can be {Cheating}: Counterfactual {Recommendation} for {Mitigating} {Clickbait} {Issue}}. In \bibinfo{booktitle}{\emph{Proceedings of the 44th {International} {ACM} {SIGIR} {Conference} on {Research} and {Development} in {Information} {Retrieval}}}. ACM.
\newblock


\bibitem[Wang and Jordan(2021)]%
        {wang2021desiderata}
\bibfield{author}{\bibinfo{person}{Yixin Wang} {and} \bibinfo{person}{Michael~I Jordan}.} \bibinfo{year}{2021}\natexlab{}.
\newblock \showarticletitle{Desiderata for representation learning: A causal perspective}.
\newblock \bibinfo{journal}{\emph{arXiv}} (\bibinfo{year}{2021}).
\newblock


\bibitem[Wang et~al\mbox{.}(2021c)]%
        {wang2021dskreg}
\bibfield{author}{\bibinfo{person}{Yu Wang}, \bibinfo{person}{Zhiwei Liu}, \bibinfo{person}{Ziwei Fan}, \bibinfo{person}{Lichao Sun}, {and} \bibinfo{person}{Philip~S Yu}.} \bibinfo{year}{2021}\natexlab{c}.
\newblock \showarticletitle{Dskreg: Differentiable sampling on knowledge graph for recommendation with relational gnn}. In \bibinfo{booktitle}{\emph{Proceedings of the 30th {ACM} {International} {Conference} on {Information} \& {Knowledge} {Management}}}. \bibinfo{pages}{3513--3517}.
\newblock


\bibitem[Wang et~al\mbox{.}(2023)]%
        {10.1145/3547333}
\bibfield{author}{\bibinfo{person}{Yifan Wang}, \bibinfo{person}{Weizhi Ma}, \bibinfo{person}{Min Zhang}, \bibinfo{person}{Yiqun Liu}, {and} \bibinfo{person}{Shaoping Ma}.} \bibinfo{year}{2023}\natexlab{}.
\newblock \showarticletitle{A Survey on the Fairness of Recommender Systems}.
\newblock \bibinfo{journal}{\emph{ACM Trans. Inf. Syst.}} \bibinfo{volume}{41}, \bibinfo{number}{3}, Article \bibinfo{articleno}{52} (\bibinfo{date}{feb} \bibinfo{year}{2023}), \bibinfo{numpages}{43}~pages.
\newblock
\showISSN{1046-8188}
\urldef\tempurl%
\url{https://doi.org/10.1145/3547333}
\showDOI{\tempurl}


\bibitem[Wang et~al\mbox{.}(2022b)]%
        {wang2022molecular}
\bibfield{author}{\bibinfo{person}{Yuyang Wang}, \bibinfo{person}{Jianren Wang}, \bibinfo{person}{Zhonglin Cao}, {and} \bibinfo{person}{Amir~Barati Farimani}.} \bibinfo{year}{2022}\natexlab{b}.
\newblock \showarticletitle{Molecular contrastive learning of representations via graph neural networks.}
\newblock \bibinfo{journal}{\emph{Nature Machine Intelligence}} \bibinfo{volume}{4}, \bibinfo{number}{3} (\bibinfo{year}{2022}), \bibinfo{pages}{279--287}.
\newblock


\bibitem[Wei et~al\mbox{.}(2019)]%
        {wei2019mmgcn}
\bibfield{author}{\bibinfo{person}{Yinwei Wei}, \bibinfo{person}{Xiang Wang}, \bibinfo{person}{Liqiang Nie}, \bibinfo{person}{Xiangnan He}, \bibinfo{person}{Richang Hong}, {and} \bibinfo{person}{Tat-Seng Chua}.} \bibinfo{year}{2019}\natexlab{}.
\newblock \showarticletitle{MMGCN - {Multi}-modal {Graph} {Convolution} {Network} for {Personalized} {Recommendation} of {Micro}-video}. In \bibinfo{booktitle}{\emph{Proceedings of the 27th {ACM} {International} {Conference} on {Multimedia}}}. ACM, \bibinfo{pages}{1437--1445}.
\newblock


\bibitem[Wellawatte et~al\mbox{.}(2022)]%
        {wellawatte2022model}
\bibfield{author}{\bibinfo{person}{Geemi~P Wellawatte}, \bibinfo{person}{Aditi Seshadri}, {and} \bibinfo{person}{Andrew~D White}.} \bibinfo{year}{2022}\natexlab{}.
\newblock \showarticletitle{Model agnostic generation of counterfactual explanations for molecules}.
\newblock \bibinfo{journal}{\emph{Chemical science}} \bibinfo{volume}{13}, \bibinfo{number}{13} (\bibinfo{year}{2022}), \bibinfo{pages}{3697--3705}.
\newblock


\bibitem[Westerlund(2019)]%
        {westerlund2019the}
\bibfield{author}{\bibinfo{person}{Mika Westerlund}.} \bibinfo{year}{2019}\natexlab{}.
\newblock \showarticletitle{The emergence of deepfake technology: A review}.
\newblock \bibinfo{journal}{\emph{Technology innovation management review}} \bibinfo{volume}{9}, \bibinfo{number}{11} (\bibinfo{year}{2019}).
\newblock


\bibitem[Wieder et~al\mbox{.}(2020)]%
        {wieder2020a}
\bibfield{author}{\bibinfo{person}{Oliver Wieder}, \bibinfo{person}{Stefan Kohlbacher}, \bibinfo{person}{M{\' e}laine Kuenemann}, \bibinfo{person}{Arthur Garon}, \bibinfo{person}{Pierre Ducrot}, \bibinfo{person}{Thomas Seidel}, {and} \bibinfo{person}{Thierry Langer}.} \bibinfo{year}{2020}\natexlab{}.
\newblock \showarticletitle{A compact review of molecular property prediction with graph neural networks}.
\newblock \bibinfo{journal}{\emph{Drug Discovery Today: Technologies}}  \bibinfo{volume}{37} (\bibinfo{year}{2020}), \bibinfo{pages}{1--12}.
\newblock


\bibitem[Wu et~al\mbox{.}(2021b)]%
        {wu2021fedgnn}
\bibfield{author}{\bibinfo{person}{Chuhan Wu}, \bibinfo{person}{Fangzhao Wu}, \bibinfo{person}{Yang Cao}, \bibinfo{person}{Yongfeng Huang}, {and} \bibinfo{person}{Xing Xie}.} \bibinfo{year}{2021}\natexlab{b}.
\newblock \showarticletitle{Fedgnn: Federated graph neural network for privacy-preserving recommendation}.
\newblock \bibinfo{journal}{\emph{arXiv}} (\bibinfo{year}{2021}).
\newblock


\bibitem[Wu et~al\mbox{.}(2021a)]%
        {wu2021counterfactual}
\bibfield{author}{\bibinfo{person}{Haoran Wu}, \bibinfo{person}{Wei Chen}, \bibinfo{person}{Shuang Xu}, {and} \bibinfo{person}{Bo Xu}.} \bibinfo{year}{2021}\natexlab{a}.
\newblock \showarticletitle{Counterfactual supporting facts extraction for explainable medical record based diagnosis with graph network}. In \bibinfo{booktitle}{\emph{Proceedings of the 2021 {Conference} of the {North} {American} {Chapter} of the {Association} for {Computational} {Linguistics}: Human {Language} {Technologies}}}. \bibinfo{pages}{1942--1955}.
\newblock


\bibitem[Wu et~al\mbox{.}(2019)]%
        {wu2019dual}
\bibfield{author}{\bibinfo{person}{Qitian Wu}, \bibinfo{person}{Hengrui Zhang}, \bibinfo{person}{Xiaofeng Gao}, \bibinfo{person}{Peng He}, \bibinfo{person}{Paul Weng}, \bibinfo{person}{Han Gao}, {and} \bibinfo{person}{Guihai Chen}.} \bibinfo{year}{2019}\natexlab{}.
\newblock \showarticletitle{Dual graph attention networks for deep latent representation of multifaceted social effects in recommender systems}. In \bibinfo{booktitle}{\emph{The world wide web conference}}. \bibinfo{pages}{2091--2102}.
\newblock


\bibitem[Wu et~al\mbox{.}(2022)]%
        {wu2022graph}
\bibfield{author}{\bibinfo{person}{Shiwen Wu}, \bibinfo{person}{Fei Sun}, \bibinfo{person}{Wentao Zhang}, \bibinfo{person}{Xu Xie}, {and} \bibinfo{person}{Bin Cui}.} \bibinfo{year}{2022}\natexlab{}.
\newblock \showarticletitle{Graph neural networks in recommender systems: a survey}.
\newblock \bibinfo{journal}{\emph{Comput. Surveys}} \bibinfo{volume}{55}, \bibinfo{number}{5} (\bibinfo{year}{2022}), \bibinfo{pages}{1--37}.
\newblock


\bibitem[Wu et~al\mbox{.}(2020)]%
        {wu2020a}
\bibfield{author}{\bibinfo{person}{Zonghan Wu}, \bibinfo{person}{Shirui Pan}, \bibinfo{person}{Fengwen Chen}, \bibinfo{person}{Guodong Long}, \bibinfo{person}{Chengqi Zhang}, {and} \bibinfo{person}{S~Yu Philip}.} \bibinfo{year}{2020}\natexlab{}.
\newblock \showarticletitle{A comprehensive survey on graph neural networks}.
\newblock \bibinfo{journal}{\emph{IEEE transactions on neural networks and learning systems}} \bibinfo{volume}{32}, \bibinfo{number}{1} (\bibinfo{year}{2020}), \bibinfo{pages}{4--24}.
\newblock


\bibitem[Xiao et~al\mbox{.}(2023)]%
        {xiao2023counterfactual}
\bibfield{author}{\bibinfo{person}{Chunjing Xiao}, \bibinfo{person}{Xovee Xu}, \bibinfo{person}{Yue Lei}, \bibinfo{person}{Kunpeng Zhang}, \bibinfo{person}{Siyuan Liu}, {and} \bibinfo{person}{Fan Zhou}.} \bibinfo{year}{2023}\natexlab{}.
\newblock \showarticletitle{Counterfactual {Graph} {Learning} for {Anomaly} {Detection} on {Attributed} {Networks}}.
\newblock \bibinfo{journal}{\emph{IEEE Transactions on Knowledge and Data Engineering}} (\bibinfo{year}{2023}).
\newblock


\bibitem[Xiao et~al\mbox{.}(2022b)]%
        {xiao2022decoupled}
\bibfield{author}{\bibinfo{person}{Teng Xiao}, \bibinfo{person}{Zhengyu Chen}, \bibinfo{person}{Zhimeng Guo}, \bibinfo{person}{Zeyang Zhuang}, {and} \bibinfo{person}{Suhang Wang}.} \bibinfo{year}{2022}\natexlab{b}.
\newblock \showarticletitle{Decoupled Self-supervised Learning for Graphs}.
\newblock \bibinfo{journal}{\emph{Advances in Neural Information Processing Systems}} (\bibinfo{year}{2022}), \bibinfo{pages}{620--634}.
\newblock


\bibitem[Xiao et~al\mbox{.}(2021)]%
        {xiao2021learning}
\bibfield{author}{\bibinfo{person}{Teng Xiao}, \bibinfo{person}{Zhengyu Chen}, \bibinfo{person}{Donglin Wang}, {and} \bibinfo{person}{Suhang Wang}.} \bibinfo{year}{2021}\natexlab{}.
\newblock \showarticletitle{Learning how to propagate messages in graph neural networks}. In \bibinfo{booktitle}{\emph{Proceedings of the 27th ACM SIGKDD Conference on Knowledge Discovery \& Data Mining}}. \bibinfo{pages}{1894--1903}.
\newblock


\bibitem[Xiao et~al\mbox{.}(2022a)]%
        {xiao2022representation}
\bibfield{author}{\bibinfo{person}{Teng Xiao}, \bibinfo{person}{Zhengyu Chen}, {and} \bibinfo{person}{Suhang Wang}.} \bibinfo{year}{2022}\natexlab{a}.
\newblock \showarticletitle{Representation {Matters} {When} {Learning} {From} {Biased} {Feedback} in {Recommendation}}. In \bibinfo{booktitle}{\emph{Proceedings of the 31st {ACM} {International} {Conference} on {Information} \& {Knowledge} {Management}}}. \bibinfo{pages}{2220--2229}.
\newblock


\bibitem[Xiao et~al\mbox{.}(2019a)]%
        {xiao2019dynamic}
\bibfield{author}{\bibinfo{person}{Teng Xiao}, \bibinfo{person}{Shangsong Liang}, {and} \bibinfo{person}{Zaiqiao Meng}.} \bibinfo{year}{2019}\natexlab{a}.
\newblock \showarticletitle{Dynamic collaborative recurrent learning}. In \bibinfo{booktitle}{\emph{Proceedings of the 28th ACM international conference on information and knowledge management}}. \bibinfo{pages}{1151--1160}.
\newblock


\bibitem[Xiao et~al\mbox{.}(2019b)]%
        {xiao2019hierarchical}
\bibfield{author}{\bibinfo{person}{Teng Xiao}, \bibinfo{person}{Shangsong Liang}, {and} \bibinfo{person}{Zaiqiao Meng}.} \bibinfo{year}{2019}\natexlab{b}.
\newblock \showarticletitle{Hierarchical neural variational model for personalized sequential recommendation}. In \bibinfo{booktitle}{\emph{The World Wide Web Conference}}. \bibinfo{pages}{3377--3383}.
\newblock


\bibitem[Xiao et~al\mbox{.}(2019c)]%
        {xiao2019bayesian}
\bibfield{author}{\bibinfo{person}{Teng Xiao}, \bibinfo{person}{Shangsong Liang}, \bibinfo{person}{Weizhou Shen}, {and} \bibinfo{person}{Zaiqiao Meng}.} \bibinfo{year}{2019}\natexlab{c}.
\newblock \showarticletitle{Bayesian deep collaborative matrix factorization}. In \bibinfo{booktitle}{\emph{Proceedings of the AAAI Conference on Artificial Intelligence}}. \bibinfo{pages}{5474--5481}.
\newblock


\bibitem[Xiao and Wang(2022)]%
        {xiao2022towards}
\bibfield{author}{\bibinfo{person}{Teng Xiao} {and} \bibinfo{person}{Suhang Wang}.} \bibinfo{year}{2022}\natexlab{}.
\newblock \showarticletitle{Towards unbiased and robust causal ranking for recommender systems}. In \bibinfo{booktitle}{\emph{Proceedings of the Fifteenth ACM International Conference on Web Search and Data Mining}}. \bibinfo{pages}{1158--1167}.
\newblock


\bibitem[Xiong et~al\mbox{.}(2019)]%
        {xiong2019pushing}
\bibfield{author}{\bibinfo{person}{Zhaoping Xiong}, \bibinfo{person}{Dingyan Wang}, \bibinfo{person}{Xiaohong Liu}, \bibinfo{person}{Feisheng Zhong}, \bibinfo{person}{Xiaozhe Wan}, \bibinfo{person}{Xutong Li}, \bibinfo{person}{Zhaojun Li}, \bibinfo{person}{Xiaomin Luo}, \bibinfo{person}{Kaixian Chen}, \bibinfo{person}{Hualiang Jiang}, {et~al\mbox{.}}} \bibinfo{year}{2019}\natexlab{}.
\newblock \showarticletitle{Pushing the boundaries of molecular representation for drug discovery with the graph attention mechanism}.
\newblock \bibinfo{journal}{\emph{Journal of medicinal chemistry}} \bibinfo{volume}{63}, \bibinfo{number}{16} (\bibinfo{year}{2019}), \bibinfo{pages}{8749--8760}.
\newblock


\bibitem[Xu et~al\mbox{.}(2022a)]%
        {xu2022hpgmn}
\bibfield{author}{\bibinfo{person}{Junjie Xu}, \bibinfo{person}{Enyan Dai}, \bibinfo{person}{Xiang Zhang}, {and} \bibinfo{person}{Suhang Wang}.} \bibinfo{year}{2022}\natexlab{a}.
\newblock \showarticletitle{HP-{GMN}: Graph {Memory} {Networks} for {Heterophilous} {Graphs}.}. In \bibinfo{booktitle}{\emph{IEEE {International} {Conference} on {Data} {Mining} ({ICDM})}}. IEEE, \bibinfo{pages}{1263--1268}.
\newblock


\bibitem[Xu et~al\mbox{.}(2018)]%
        {xu2018powerful}
\bibfield{author}{\bibinfo{person}{Keyulu Xu}, \bibinfo{person}{Weihua Hu}, \bibinfo{person}{Jure Leskovec}, {and} \bibinfo{person}{Stefanie Jegelka}.} \bibinfo{year}{2018}\natexlab{}.
\newblock \showarticletitle{How powerful are graph neural networks?}
\newblock \bibinfo{journal}{\emph{arXiv preprint arXiv:1810.00826}} (\bibinfo{year}{2018}).
\newblock


\bibitem[Xu et~al\mbox{.}(2019)]%
        {xu2019how}
\bibfield{author}{\bibinfo{person}{Keyulu Xu}, \bibinfo{person}{Weihua Hu}, \bibinfo{person}{Jure Leskovec}, {and} \bibinfo{person}{Stefanie Jegelka}.} \bibinfo{year}{2019}\natexlab{}.
\newblock \showarticletitle{How {Powerful} are {Graph} {Neural} {Networks}?}. In \bibinfo{booktitle}{\emph{International {Conference} on {Learning} {Representations} ({ICLR})}}.
\newblock


\bibitem[Xu et~al\mbox{.}(2022b)]%
        {xu2022counterfactual}
\bibfield{author}{\bibinfo{person}{Ran Xu}, \bibinfo{person}{Yue Yu}, \bibinfo{person}{Chao Zhang}, \bibinfo{person}{Mohammed~K Ali}, \bibinfo{person}{Joyce~C Ho}, {and} \bibinfo{person}{Carl Yang}.} \bibinfo{year}{2022}\natexlab{b}.
\newblock \showarticletitle{Counterfactual and factual reasoning over hypergraphs for interpretable clinical predictions on ehr}. In \bibinfo{booktitle}{\emph{Machine {Learning} for {Health}}}. PMLR, \bibinfo{pages}{259--278}.
\newblock


\bibitem[Yao et~al\mbox{.}(2021a)]%
        {yao2021a}
\bibfield{author}{\bibinfo{person}{Liuyi Yao}, \bibinfo{person}{Zhixuan Chu}, \bibinfo{person}{Sheng Li}, \bibinfo{person}{Yaliang Li}, \bibinfo{person}{Jing Gao}, {and} \bibinfo{person}{Aidong Zhang}.} \bibinfo{year}{2021}\natexlab{a}.
\newblock \showarticletitle{A survey on causal inference}.
\newblock \bibinfo{journal}{\emph{ACM Transactions on Knowledge Discovery from Data (TKDD)}} \bibinfo{volume}{15}, \bibinfo{number}{5} (\bibinfo{year}{2021}), \bibinfo{pages}{1--46}.
\newblock


\bibitem[Yao et~al\mbox{.}(2021b)]%
        {yao2021survey}
\bibfield{author}{\bibinfo{person}{Liuyi Yao}, \bibinfo{person}{Zhixuan Chu}, \bibinfo{person}{Sheng Li}, \bibinfo{person}{Yaliang Li}, \bibinfo{person}{Jing Gao}, {and} \bibinfo{person}{Aidong Zhang}.} \bibinfo{year}{2021}\natexlab{b}.
\newblock \showarticletitle{A survey on causal inference}.
\newblock \bibinfo{journal}{\emph{ACM Transactions on Knowledge Discovery from Data (TKDD)}} \bibinfo{volume}{15}, \bibinfo{number}{5} (\bibinfo{year}{2021}), \bibinfo{pages}{1--46}.
\newblock


\bibitem[Yeh and Lien(2009)]%
        {yeh2009the}
\bibfield{author}{\bibinfo{person}{I-Cheng Yeh} {and} \bibinfo{person}{Che-hui Lien}.} \bibinfo{year}{2009}\natexlab{}.
\newblock \showarticletitle{The comparisons of data mining techniques for the predictive accuracy of probability of default of credit card clients}.
\newblock \bibinfo{journal}{\emph{Expert systems with applications}} \bibinfo{volume}{36}, \bibinfo{number}{2} (\bibinfo{year}{2009}), \bibinfo{pages}{2473--2480}.
\newblock


\bibitem[Yin et~al\mbox{.}(2017)]%
        {yin2017spatialaware}
\bibfield{author}{\bibinfo{person}{Hongzhi Yin}, \bibinfo{person}{Weiqing Wang}, \bibinfo{person}{Hao Wang}, \bibinfo{person}{Ling Chen}, {and} \bibinfo{person}{Xiaofang Zhou}.} \bibinfo{year}{2017}\natexlab{}.
\newblock \showarticletitle{Spatial-aware hierarchical collaborative deep learning for {POI} recommendation}.
\newblock \bibinfo{journal}{\emph{IEEE Transactions on Knowledge and Data Engineering}} \bibinfo{volume}{29}, \bibinfo{number}{11} (\bibinfo{year}{2017}), \bibinfo{pages}{2537--2551}.
\newblock


\bibitem[Ying et~al\mbox{.}(2018)]%
        {ying2018graph}
\bibfield{author}{\bibinfo{person}{Rex Ying}, \bibinfo{person}{Ruining He}, \bibinfo{person}{Kaifeng Chen}, \bibinfo{person}{Pong Eksombatchai}, \bibinfo{person}{William~L Hamilton}, {and} \bibinfo{person}{Jure Leskovec}.} \bibinfo{year}{2018}\natexlab{}.
\newblock \showarticletitle{Graph convolutional neural networks for web-scale recommender systems}. In \bibinfo{booktitle}{\emph{Proceedings of the 24th ACM SIGKDD international conference on knowledge discovery \& data mining}}. \bibinfo{pages}{974--983}.
\newblock


\bibitem[Ying et~al\mbox{.}(2019)]%
        {ying2019gnnexplainer}
\bibfield{author}{\bibinfo{person}{Zhitao Ying}, \bibinfo{person}{Dylan Bourgeois}, \bibinfo{person}{Jiaxuan You}, \bibinfo{person}{Marinka Zitnik}, {and} \bibinfo{person}{Jure Leskovec}.} \bibinfo{year}{2019}\natexlab{}.
\newblock \showarticletitle{GNNExplainer: Generating {Explanations} for {Graph} {Neural} {Networks}.}. In \bibinfo{booktitle}{\emph{Conference on {Neural} {Information} {Processing} {Systems} ({NeurIPS})}}. \bibinfo{pages}{9240--9251}.
\newblock


\bibitem[You et~al\mbox{.}(2020)]%
        {you2020graph}
\bibfield{author}{\bibinfo{person}{Yuning You}, \bibinfo{person}{Tianlong Chen}, \bibinfo{person}{Yongduo Sui}, \bibinfo{person}{Ting Chen}, \bibinfo{person}{Zhangyang Wang}, {and} \bibinfo{person}{Yang Shen}.} \bibinfo{year}{2020}\natexlab{}.
\newblock \showarticletitle{Graph {Contrastive} {Learning} with {Augmentations}.}. In \bibinfo{booktitle}{\emph{Conference on {Neural} {Information} {Processing} {Systems} ({NeurIPS})}}, Vol.~\bibinfo{volume}{33}. \bibinfo{pages}{5812--5823}.
\newblock


\bibitem[Yuan et~al\mbox{.}(2020)]%
        {yuan2020xgnn}
\bibfield{author}{\bibinfo{person}{Hao Yuan}, \bibinfo{person}{Jiliang Tang}, \bibinfo{person}{Xia Hu}, {and} \bibinfo{person}{Shuiwang Ji}.} \bibinfo{year}{2020}\natexlab{}.
\newblock \showarticletitle{XGNN: Towards {Model}-{Level} {Explanations} of {Graph} {Neural} {Networks}.}. In \bibinfo{booktitle}{\emph{ACM {SIGKDD} {Conference} on {Knowledge} {Discovery} and {Data} {Mining} ({KDD})}}. ACM, \bibinfo{pages}{430--438}.
\newblock


\bibitem[Yuan et~al\mbox{.}(2022)]%
        {yuan2022explainability}
\bibfield{author}{\bibinfo{person}{Hao Yuan}, \bibinfo{person}{Haiyang Yu}, \bibinfo{person}{Shurui Gui}, {and} \bibinfo{person}{Shuiwang Ji}.} \bibinfo{year}{2022}\natexlab{}.
\newblock \showarticletitle{Explainability in graph neural networks: A taxonomic survey}.
\newblock \bibinfo{journal}{\emph{IEEE Transactions on Pattern Analysis and Machine Intelligence}} (\bibinfo{year}{2022}).
\newblock


\bibitem[Zhang et~al\mbox{.}(2022a)]%
        {zhang2022counterfactual}
\bibfield{author}{\bibinfo{person}{Baoliang Zhang}, \bibinfo{person}{Xiaoxin Guo}, \bibinfo{person}{Qifeng Lin}, \bibinfo{person}{Haoren Wang}, {and} \bibinfo{person}{Songbai Xu}.} \bibinfo{year}{2022}\natexlab{a}.
\newblock \showarticletitle{Counterfactual inference graph network for disease prediction}.
\newblock \bibinfo{journal}{\emph{Knowledge-Based Systems}}  \bibinfo{volume}{255} (\bibinfo{year}{2022}), \bibinfo{pages}{109722}.
\newblock


\bibitem[Zhang et~al\mbox{.}(2022b)]%
        {zhang2022trustworthy}
\bibfield{author}{\bibinfo{person}{He Zhang}, \bibinfo{person}{Bang Wu}, \bibinfo{person}{Xingliang Yuan}, \bibinfo{person}{Shirui Pan}, \bibinfo{person}{Hanghang Tong}, {and} \bibinfo{person}{Jian Pei}.} \bibinfo{year}{2022}\natexlab{b}.
\newblock \showarticletitle{Trustworthy Graph Neural Networks: Aspects, Methods and Trends}.
\newblock \bibinfo{journal}{\emph{arXiv preprint arXiv:2205.07424}} (\bibinfo{year}{2022}).
\newblock


\bibitem[Zhang et~al\mbox{.}(2021c)]%
        {zhang2021from}
\bibfield{author}{\bibinfo{person}{Hengrui Zhang}, \bibinfo{person}{Qitian Wu}, \bibinfo{person}{Junchi Yan}, \bibinfo{person}{David Wipf}, {and} \bibinfo{person}{Philip~S. Yu}.} \bibinfo{year}{2021}\natexlab{c}.
\newblock \showarticletitle{From {Canonical} {Correlation} {Analysis} to {Self}-supervised {Graph} {Neural} {Networks}.}. In \bibinfo{booktitle}{\emph{Conference on {Neural} {Information} {Processing} {Systems} ({NeurIPS})}}, Vol.~\bibinfo{volume}{34}. \bibinfo{pages}{76--89}.
\newblock


\bibitem[Zhang and Chen(2018)]%
        {zhang2018link}
\bibfield{author}{\bibinfo{person}{Muhan Zhang} {and} \bibinfo{person}{Yixin Chen}.} \bibinfo{year}{2018}\natexlab{}.
\newblock \showarticletitle{Link {Prediction} {Based} on {Graph} {Neural} {Networks}.}. In \bibinfo{booktitle}{\emph{Conference on {Neural} {Information} {Processing} {Systems} ({NeurIPS})}}. \bibinfo{pages}{5171--5181}.
\newblock


\bibitem[Zhang et~al\mbox{.}(2018)]%
        {zhang2018an}
\bibfield{author}{\bibinfo{person}{Muhan Zhang}, \bibinfo{person}{Zhicheng Cui}, \bibinfo{person}{Marion Neumann}, {and} \bibinfo{person}{Yixin Chen}.} \bibinfo{year}{2018}\natexlab{}.
\newblock \showarticletitle{An {End}-to-{End} {Deep} {Learning} {Architecture} for {Graph} {Classification}.}. In \bibinfo{booktitle}{\emph{AAAI {Conference} on {Artificial} {Intelligence} ({AAAI})}}. \bibinfo{pages}{4438--4445}.
\newblock


\bibitem[Zhang et~al\mbox{.}(2019)]%
        {zhang2019deep}
\bibfield{author}{\bibinfo{person}{Shuai Zhang}, \bibinfo{person}{Lina Yao}, \bibinfo{person}{Aixin Sun}, {and} \bibinfo{person}{Yi Tay}.} \bibinfo{year}{2019}\natexlab{}.
\newblock \showarticletitle{Deep learning based recommender system: A survey and new perspectives}.
\newblock \bibinfo{journal}{\emph{ACM computing surveys (CSUR)}} \bibinfo{volume}{52}, \bibinfo{number}{1} (\bibinfo{year}{2019}), \bibinfo{pages}{1--38}.
\newblock


\bibitem[Zhang et~al\mbox{.}(2021b)]%
        {zhang2021deep}
\bibfield{author}{\bibinfo{person}{Weinan Zhang}, \bibinfo{person}{Jiarui Qin}, \bibinfo{person}{Wei Guo}, \bibinfo{person}{Ruiming Tang}, {and} \bibinfo{person}{Xiuqiang He}.} \bibinfo{year}{2021}\natexlab{b}.
\newblock \showarticletitle{Deep {Learning} for {Click}-{Through} {Rate} {Estimation}.}. In \bibinfo{booktitle}{\emph{International {Joint} {Conference} on {Artificial} {Intelligence} ({IJCAI})}}. \bibinfo{pages}{4695--4703}.
\newblock


\bibitem[Zhang et~al\mbox{.}(2021d)]%
        {zhang2021multi}
\bibfield{author}{\bibinfo{person}{Xu Zhang}, \bibinfo{person}{Liang Zhang}, \bibinfo{person}{Bo Jin}, {and} \bibinfo{person}{Xinjiang Lu}.} \bibinfo{year}{2021}\natexlab{d}.
\newblock \showarticletitle{A Multi-view Confidence-calibrated Framework for Fair and Stable Graph Representation Learning}. In \bibinfo{booktitle}{\emph{2021 IEEE International Conference on Data Mining (ICDM)}}. IEEE, \bibinfo{pages}{1493--1498}.
\newblock


\bibitem[Zhang et~al\mbox{.}(2021a)]%
        {zhang2021a}
\bibfield{author}{\bibinfo{person}{Yin Zhang}, \bibinfo{person}{Derek~Zhiyuan Cheng}, \bibinfo{person}{Tiansheng Yao}, \bibinfo{person}{Xinyang Yi}, \bibinfo{person}{Lichan Hong}, {and} \bibinfo{person}{Ed~H. Chi}.} \bibinfo{year}{2021}\natexlab{a}.
\newblock \showarticletitle{A {Model} of {Two} {Tales}: Dual {Transfer} {Learning} {Framework} for {Improved} {Long}-tail {Item} {Recommendation}.}. In \bibinfo{booktitle}{\emph{The {Web} {Conference} ({WWW})}}. \bibinfo{pages}{2220--2231}.
\newblock


\bibitem[Zhang et~al\mbox{.}(2020)]%
        {zhang2020relational}
\bibfield{author}{\bibinfo{person}{Zhao Zhang}, \bibinfo{person}{Fuzhen Zhuang}, \bibinfo{person}{Hengshu Zhu}, \bibinfo{person}{Zhiping Shi}, \bibinfo{person}{Hui Xiong}, {and} \bibinfo{person}{Qing He}.} \bibinfo{year}{2020}\natexlab{}.
\newblock \showarticletitle{Relational graph neural network with hierarchical attention for knowledge graph completion}. In \bibinfo{booktitle}{\emph{Proceedings of the {AAAI} {Conference} on {Artificial} {Intelligence}}}, Vol.~\bibinfo{volume}{34}. \bibinfo{pages}{9612--9619}.
\newblock


\bibitem[Zhao et~al\mbox{.}(2016)]%
        {zhao2016a}
\bibfield{author}{\bibinfo{person}{Shenglin Zhao}, \bibinfo{person}{Irwin King}, {and} \bibinfo{person}{Michael~R Lyu}.} \bibinfo{year}{2016}\natexlab{}.
\newblock \showarticletitle{A survey of point-of-interest recommendation in location-based social networks}.
\newblock \bibinfo{journal}{\emph{arXiv}} (\bibinfo{year}{2016}).
\newblock


\bibitem[Zhao et~al\mbox{.}(2022)]%
        {zhao2022learning}
\bibfield{author}{\bibinfo{person}{Tong Zhao}, \bibinfo{person}{Gang Liu}, \bibinfo{person}{Daheng Wang}, \bibinfo{person}{Wenhao Yu}, {and} \bibinfo{person}{Meng Jiang}.} \bibinfo{year}{2022}\natexlab{}.
\newblock \showarticletitle{Learning from Counterfactual Links for Link Prediction}. In \bibinfo{booktitle}{\emph{International {Conference} on {Machine} {Learning} ({ICML})}}. \bibinfo{pages}{26911--26926}.
\newblock


\bibitem[Zhao et~al\mbox{.}(2023a)]%
        {zhao2023faithful}
\bibfield{author}{\bibinfo{person}{Tianxiang Zhao}, \bibinfo{person}{Dongsheng Luo}, \bibinfo{person}{Xiang Zhang}, {and} \bibinfo{person}{Suhang Wang}.} \bibinfo{year}{2023}\natexlab{a}.
\newblock \showarticletitle{Faithful and {Consistent} {Graph} {Neural} {Network} {Explanations} with {Rationale} {Alignment}}.
\newblock \bibinfo{journal}{\emph{arXiv}} (\bibinfo{year}{2023}).
\newblock


\bibitem[Zhao et~al\mbox{.}(2023b)]%
        {zhao2023towards}
\bibfield{author}{\bibinfo{person}{Tianxiang Zhao}, \bibinfo{person}{Dongsheng Luo}, \bibinfo{person}{Xiang Zhang}, {and} \bibinfo{person}{Suhang Wang}.} \bibinfo{year}{2023}\natexlab{b}.
\newblock \showarticletitle{Towards Faithful and Consistent Explanations for Graph Neural Networks}. In \bibinfo{booktitle}{\emph{Proceedings of the 15th ACM International Conference on Web Search and Data Mining}}. ACM.
\newblock


\bibitem[Zhao et~al\mbox{.}(2021)]%
        {zhao2021graphsmote}
\bibfield{author}{\bibinfo{person}{Tianxiang Zhao}, \bibinfo{person}{Xiang Zhang}, {and} \bibinfo{person}{Suhang Wang}.} \bibinfo{year}{2021}\natexlab{}.
\newblock \showarticletitle{Graphsmote: Imbalanced node classification on graphs with graph neural networks}. In \bibinfo{booktitle}{\emph{Proceedings of the 14th ACM international conference on web search and data mining}}. \bibinfo{pages}{833--841}.
\newblock


\bibitem[Zhou et~al\mbox{.}(2018)]%
        {zhou2018deep}
\bibfield{author}{\bibinfo{person}{Guorui Zhou}, \bibinfo{person}{Xiaoqiang Zhu}, \bibinfo{person}{Chenru Song}, \bibinfo{person}{Ying Fan}, \bibinfo{person}{Han Zhu}, \bibinfo{person}{Xiao Ma}, \bibinfo{person}{Yanghui Yan}, \bibinfo{person}{Junqi Jin}, \bibinfo{person}{Han Li}, {and} \bibinfo{person}{Kun Gai}.} \bibinfo{year}{2018}\natexlab{}.
\newblock \showarticletitle{Deep interest network for click-through rate prediction}. In \bibinfo{booktitle}{\emph{Proceedings of the 24th {ACM} {SIGKDD} international conference on knowledge discovery \& data mining}}. \bibinfo{pages}{1059--1068}.
\newblock


\bibitem[Zhu and Wang(2022)]%
        {zhu2022learning}
\bibfield{author}{\bibinfo{person}{Huaisheng Zhu} {and} \bibinfo{person}{Suhang Wang}.} \bibinfo{year}{2022}\natexlab{}.
\newblock \showarticletitle{Learning {Fair} {Models} without {Sensitive} {Attributes}: A {Generative} {Approach}}.
\newblock \bibinfo{journal}{\emph{arXiv}} (\bibinfo{year}{2022}).
\newblock


\bibitem[Zhu et~al\mbox{.}(2021)]%
        {zhu2021neural}
\bibfield{author}{\bibinfo{person}{Zhaocheng Zhu}, \bibinfo{person}{Zuobai Zhang}, \bibinfo{person}{Louis-Pascal A.~C. Xhonneux}, {and} \bibinfo{person}{Jian Tang}.} \bibinfo{year}{2021}\natexlab{}.
\newblock \showarticletitle{Neural {Bellman}-{Ford} {Networks}: A {General} {Graph} {Neural} {Network} {Framework} for {Link} {Prediction}.}. In \bibinfo{booktitle}{\emph{Conference on {Neural} {Information} {Processing} {Systems} ({NeurIPS})}}. \bibinfo{pages}{29476--29490}.
\newblock


\end{thebibliography}

\end{document}